\pdfoutput=1
\documentclass[11pt]{article}

\usepackage[final]{acl}

\usepackage{times}
\usepackage{latexsym}

\usepackage[T1]{fontenc}
\usepackage[utf8]{inputenc}

\usepackage{microtype}

\usepackage{inconsolata}

\usepackage{graphicx}

\usepackage{algorithm}
\usepackage{float}
\usepackage{algorithmicx}
\usepackage{algpseudocode}
\usepackage{wrapfig}
\usepackage{amsmath}
\usepackage{amstext}
\usepackage{rotating}
\usepackage{tcolorbox}
\usepackage{xcolor}
\usepackage{graphicx}
\usepackage{mathrsfs}
\usepackage{booktabs}
\usepackage{bbm}
\usepackage{amsmath,amssymb,amsfonts}

\definecolor{demcolor}{HTML}{ff7f0f}
\definecolor{profcolor}{HTML}{2ba02b}
\definecolor{excolor}{HTML}{9467bd}
\definecolor{gtcolor}{HTML}{e377c3}
\definecolor{singledemcol}{HTML}{ffa500}
\definecolor{bothcol}{HTML}{15becf}
\definecolor{clustcol}{HTML}{65cdaa}

\title{Value Profiles for Encoding Human Variation}

\author{Taylor Sorensen$^{1\dagger}$, Pushkar Mishra$^2$, Roma Patel$^2$, Michael Henry Tessler$^2$, \\
\textbf{Michiel Bakker}$^2$, \textbf{Georgina Evans}$^2$, \textbf{Iason Gabriel}$^2$, \textbf{Noah Goodman}$^2$, \textbf{Verena Rieser}$^2$
\\ \textsuperscript{1}Department of Computer Science, University of Washington, Seattle, WA, USA
\\ \textsuperscript{2}Google DeepMind, London, UK
\\ \textsuperscript{$\dagger$}Work done during an internship at Google DeepMind.
\\ \texttt{tsor13@cs.washington.edu}, \texttt{verenarieser@deepmind.com}
}

\newcommand{\profiletext}[1]{\textit{#1}}

\begin{document}

% \linenumbers
% \fi

\maketitle

\begin{abstract}
Modelling human variation in rating tasks is crucial for personalization, pluralistic model alignment, and computational social science.
We propose representing individuals using natural language {\em value profiles} -- descriptions of underlying values compressed from in-context demonstrations -- along with a steerable decoder model that estimates individual ratings from a rater representation. To measure the predictive information in a rater representation, we introduce an information-theoretic methodology and find that demonstrations contain the most information, followed by value profiles, then demographics. However, value profiles effectively compress the useful information from demonstrations ($>$70\% information preservation) and offer advantages in terms of scrutability, interpretability, and steerability.
Furthermore, clustering value profiles to identify similarly behaving individuals better explains rater variation than the most predictive demographic groupings.
Going beyond test set performance, we show that the decoder predictions change in line with semantic profile differences, are well-calibrated, and can help explain instance-level disagreement by simulating an annotator population.
These results demonstrate that value profiles offer novel, predictive ways to describe individual variation beyond demographics or group information.
\end{abstract}

\section{Introduction}
\label{sec:introduction}
Machine learning systems are traditionally trained to approximate a single ``ground truth" label, treating annotator disagreement as noise. However, many important tasks such as chat preferences, hate speech, and toxicity detection can have legitimate disagreement \citep{Aroyo_Welty_2015, plank-2022-problem}. Modelling this heterogeneity is important for pluralistic model alignment \citep{sorensen2024roadmappluralisticalignment}, unbiased model safety, content moderation, personalization, and more. 

We characterize three approaches to variation modelling:
(1) {\textbf{Distributional Population Modelling}}, which directly models the distribution of labels for a given rater population \citep{zhang2024divergingpreferencesannotatorsdisagree, siththaranjan2024distributionalpreferencelearningunderstanding}. This approach accounts for variance and valid disagreements between annotators but requires many raters labeling the same instances and doesn't model which raters disagree or why. (2)  {\textbf{Grouping by Characteristics}} such as demographics or annotation similarity. While grouping approaches can lead to higher agreement than the broader rater population, they still do not account for intra-group disagreement \citep{hwang2023aligninglanguagemodelsuser, prabhakaran-etal-2024-grasp}, potentially leading to flattening variance or stereotyping.
% TODO could remove this line
To capture intra-group variation, distributional learning is needed \citep{meister2024benchmarkingdistributionalalignmentlarge}.
Grouping by annotation similarity also requires significant overlap in labeled instances \citep{li-etal-2024-steerability}.
(3) {\textbf{ Individual Modelling}}. At the individual level \citep{Gordon_2022, jiang2024languagemodelsreasonindividualistic},
the target is a single "correct" answer instead of a distribution,\footnote{ At least, to the degree that people are self-consistent \citep{abercrombie2023consistencykeydisentanglinglabel}.}
allowing for standard supervised methods.
Additionally, we can obtain group or population distributions through marginalization. Individual modeling also removes the requirement for raters to have any instance overlap. 
Because of these advantages, \textit{we argue for and focus on improving individual modeling} in order to better model human variation
% TODO - could cut
% (for more, see App. \ref{app:modellingvariation}, Fig. \ref{fig:modellingvariation}).
(for more, see App. \ref{app:modellingvariation}, Fig. \ref{fig:modellingvariation}).
However, this raises the question - how should we represent an individual?

\begin{figure*}[t]
\centering
\includegraphics[width=0.8\textwidth]{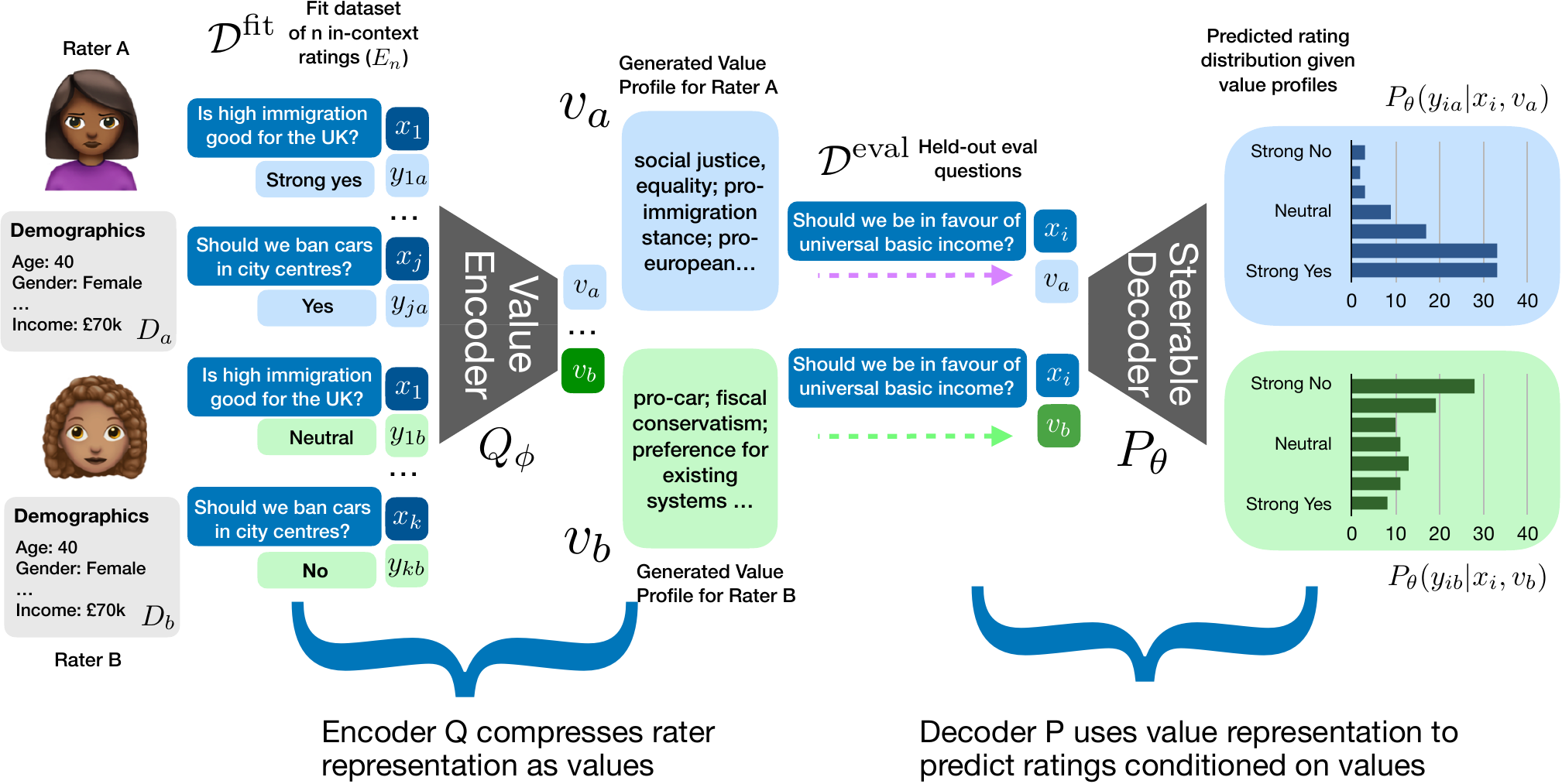}
\caption{The value profile autoencoder setup. Decoder outputs are from trained profile decoder while demographics are illustrative to preserve privacy. The encoder extracts/compresses value information from rater examples, and the decoder changes predictions on held-out questions according to the value profile.
}
% }
\label{fig:my_label}
\end{figure*}

In this work, we propose to model rater variation using individual, free-text {\em value profiles} -- interpretable natural language descriptions of human values that explain observed rating variation (\S \ref{sec:modellingvariation}).
In \S \ref{sec:informationmethodology}, we introduce a methodology to measure the information content of possible rater representations.
We carry out a series of experiments to evaluate our value profile system and other rater representations (\S \ref{sec:experimental methodology}, \ref{sec:results-ratererep}).
In \S \ref{sec:clustering}, we introduce a rater clustering algorithm that uncovers better groupings than the most predictive demographics, while loosening the typical requirement of annotators labeling overlapping instances.
In other experiments, we find that our value profile system is interpretable, well-calibrated, and helps explain rater disagreement (\S \ref{sec:extrinsicevaluation}).
We conclude by discussing related work (\S \ref{sec:relatedwork}), directions for future work (\S \ref{sec:conclusion}),
% limitations (\S \ref{sec:limitations}),
and ethical advantages (and risks) of our approach (\S \ref{sec:ethicalconsiderations}).

\section{Modelling Human Annotator Variation}
\label{sec:modellingvariation}

\subsection{Rater Representations}
Let $\mathcal{R} = \{r_1, r_2, \ldots, r_{N_R}\}$ be the raters who we wish to model, $\mathcal{X} = \{x_1, x_2, \ldots, x_{N_X}\}$ be the space of instances, and $\mathcal{Y} = \{y_1, y_2, \ldots, y_{N_Y}\}$ be the space of potential responses/ratings.
We would like to model 
$\mathcal{Y} \mid \mathcal{X}, \mathcal{R}$.
However, because we don't have sufficient information to represent (or observe) the rater $r$, we compare different potential representations for $r$:
\begin{itemize}
\item $\emptyset$: No information about $r$. In this case, 
$P(\mathcal{Y} \mid \mathcal{X}, \emptyset(r)) = \sum_{r' \in \mathcal{R}} P(\mathcal{Y} \mid \mathcal{X}, r') = P(\mathcal{Y} \mid \mathcal{X})$,
or the label distribution for the input marginalized over all raters.

\item $D$: Demographic information about $r$. $P(\mathcal{Y} \mid \mathcal{X}, D(r))$.~\footnote{In the case that many demographics are provided, this is sometimes called a ``persona" \citep{cheng-etal-2023-marked}. We also refer to this as ``demographics (all)" or ``intersectional demographics". This is in contrast to trying to model an entire demographic group at a time.
% (e.g., all raters of a particular gender).
% We compare against both settings.
}

\item $E_n$: $n$ in-context ratings as demonstrations from rater $r$. $P(\mathcal{Y} \mid \mathcal{X}, E_n(r))$.

\item $V$: A {\em value profile} natural language description of the rater's values which are relevant for the task. $P(\mathcal{Y} \mid \mathcal{X}, V(r))$.
\end{itemize}

A {\em value profile} might be elicited directly from a rater $r$ through a survey/value elicitation process.
In absence of this data, we propose to infer $V$ from observed example ratings $E_n$ through an autoencoder setup.

\begin{figure*}[t]
\centering
\includegraphics[width=\textwidth]{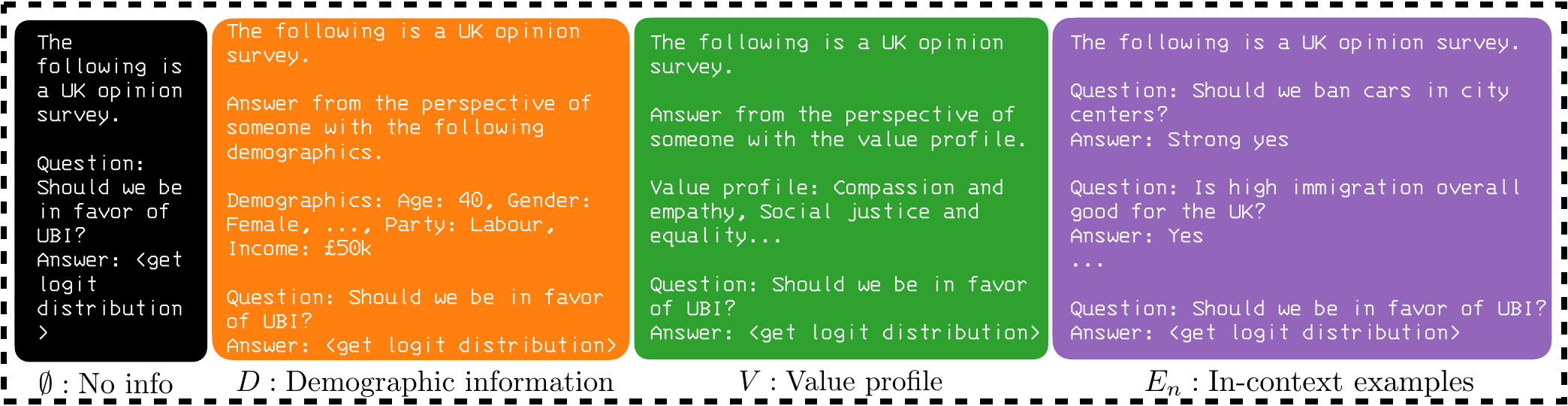}
% \includegraphics[width=0.95\textwidth]{files/raterreplarge-cropped.pdf}
% \includegraphics[width=0.9\textwidth]{files/raterreplarge-cropped.pdf}
% \small
% \vspace{-13pt}
\vspace{-5pt}
\caption{Rater representations and example corresponding decoder prompts
($\emptyset$, $D$, $V$, $E_n$).
The decoder predicts the rater's annotation given the rater representation.
}
\label{fig:raterrep}
\vspace{-2pt}
\end{figure*}

\subsection{Autoencoding Rater Values}
\label{sec:autoencoder}

Let $r_i$ be a particular rater $i$ drawn from the population of $n$ raters, $x_j$ be a particular instance $j$, and $y_{ij}$ be the rating that rater $i$ gave to instance $j$. Let $\mathcal{D}_i = \{y_{i1}, y_{i2}, \ldots, y_{iN_i}\}$ be the set of $N_i$ ratings we have for rater $i$. We can build a language model encoder $Q_\phi$ which estimates a value profile for each rater $r_i$ from a set of (fit) demonstrations drawn from $D_i$, with corresponding probability distribution $q_\phi: E_n \to V$. Similarly, a decoder $P_\theta$ can estimate the label probability distribution $P(\mathcal{Y}|\mathcal{X},V(\mathcal{R}))$ $\approx$ $p_\theta: \mathcal{X}, V \to \mathcal{Y}$. Given this, the entire autoencoder system can be evaluated by sampling a value profile from the encoder $v_i \sim q_\phi(E_n(r_i))$ and calculating the (cross-entropy) loss on unseen examples.

We randomly partition the instances into $\mathcal{D}_i^{\text{fit}}$ for fitting a value profile and $\mathcal{D}_i^{\text{eval}}$
to train the decoder to generalize to held-out ratings. The setup may be seen as a way to ``compress'' predictive information about a rater's labeling process from their examples $E_n(r_i)$ to a natural language value profile $v_i$.

In practice, we initialize the encoder and decoder parameters $\phi$ and $\theta$ as prompted language models (prompts in Figs. \ref{fig:encoderprompt}/\ref{fig:decoderprompt}). For the experiments in the paper, we freeze the encoder parameters and optimize the decoder directly with supervised finetuning. We choose to do this 1) as prompted language models performed quite well at encoding, preserving $>70$\% of usable information. (cf. Eq. \ref{eq:info_preserved}, Fig. \ref{fig:info_preserved}), 2) in order to regularize the encoder to remain human understandable/interpretable, and 3) to preserve generalizability across datasets.

We compare against the alternative rater representations of no information $\emptyset$, demographics $D$, and examples $E_n$, by similarly fitting a decoder $D_\theta$ to estimate  $p_\theta(\mathcal{Y}|\mathcal{X}, \cdot)$. All parameters are initialized with a prompted language model. To ensure comparable results, we use $\mathcal{D}^{\textrm{fit}}$ demonstrations for the in-context demonstrations $E_n$ and inferring value profiles in training and testing and the $\mathcal{D}^\textrm{eval}$ demonstrations as held-out targets for all rater representations.

\section{Estimating Usable Rater Information}
\label{sec:informationmethodology}

We wish to compare the usable information for each rater representation. To do this, we extend \citet{xu2020theoryusableinformationcomputational}'s concept of  $\mathcal{V}$-information, which was created to analogize the concept of mutual information between random variables $A,B$ to constrained computational families. We extend $\mathcal{V}$-information to the case where we have a third random variable, $C$, with computational family $\mathcal{V} \subseteq \{f: \mathscr{A} \cup \{\emptyset\}, C \to \mathcal{P}(\mathscr{B})\}$:
\begin{equation}
% \begin{align}
% \textbf{pred. cond. }\mathcal{V}\textbf{-entropy of } B|A,C \textbf{:  } \\
H_\mathcal{V}(B \mid A, C) = \inf_{f \in \mathcal{V}} \mathbb{E}_{a,b,c \sim A,B,C} [-\log f[a,c](b)]
% \end{align}
\end{equation}
\begin{equation}
% \textbf{pred. }\mathcal{V}\textbf{-information from } A \to B | C \textbf{: }
I_\mathcal{V}(A \to B \mid C) = H_\mathcal{V}(B \mid \emptyset, C) - H_\mathcal{V}(B \mid A, C)
\end{equation}

We can then measure predictive information from each rater representation ($\emptyset$, $D$, $E_n$, $V$) to ratings $\mathcal{Y}$, given instances $\mathcal{X}$. I.e., we can estimate how much more we know about how a rater will respond given particular information about them, as compared to knowing nothing about the rater.

As \citet{ethayarajh2022understandingdatasetdifficultymathcalvusable} show in a similar extension of $\mathcal{V}$-information, assuming we have an i.i.d. dataset of observations, we can get an unbiased estimate of this quantity for a computational family by training a model in each informational setting and comparing the held-out test losses to a trained model with no information. For more details, see Algorithm \ref{alg:pred-info} (inspired by \citealt{ethayarajh2022understandingdatasetdifficultymathcalvusable}).
% TODO - add back in
To contextualize the algorithm with an example loss plot, see Figure \ref{fig:example}.

\begin{algorithm}[H]
\begin{algorithmic}[0]
\caption{Computing Predictive $\mathcal{V}$-Information}
\small
\label{alg:pred-info}
% \Require \textbf{Input:}
\State \textbf{Input:}
\State $\bullet$ Training data $\mathcal{D}_\text{train} = \{(r_i, x_j, y_{ij})\}$
$= \{(r_i,x_j,y_{ij}): (x_j,y_{ij}) \in R_i^\text{test}, r_i \in \text{train raters } R^\text{train}\}$
\State $\bullet$ Test data $\mathcal{D}_\text{test}$ for held-out raters $R^\text{test}$, $R^\text{train} \cap R^\text{test} = \emptyset$
\State $\bullet$ Initialized decoder $d$, a prompted, pretrained LM
\State $\bullet$ Natural language rater representation $g: R \to \text{NL}$

\medskip
\State $d_g \gets$ finetune $d$ on $\{(g(r_i), x_j, y_{ij}) | (r_i,x_j,y_{ij}) \in \mathcal{D}_\text{train}\}$
\Comment{Train w/ rater information}
\State $d_\emptyset \gets$ finetune $d$ on $\{(\emptyset, x_j, y_{ij}) | (r_i,x_j,y_{ij}) \in \mathcal{D}_\text{train}\}$
\Comment{Train w/out rater information}
\State $H_V(\mathcal{Y}|\mathcal{X}), H_V(\mathcal{Y}|\mathcal{X},g(R)) \gets 0, 0$

\For{$(r_i,x_j,y_{ij}) \in \mathcal{D}_\text{test}$}
\Comment{Accumulate average held-out test losses}
\State $H_V(\mathcal{Y}|\mathcal{X}) \gets H_V(\mathcal{Y}|\mathcal{X}) - \frac{1}{|\mathcal{D}_\text{test}|} \log d_\emptyset(x_j, \emptyset)(y_{ij})$
\State $H_V(\mathcal{Y}|\mathcal{X},g(R)) \gets H_V(\mathcal{Y}|\mathcal{X},g(R)) - \frac{1}{|\mathcal{D}_\text{test}|} \log d_g(x_j, g(r_i))(y_{ij})$
\EndFor

\State $\hat{I}_V(g(R) \to \mathcal{Y}|\mathcal{X}) \gets H_V(\mathcal{Y}|\mathcal{X}) - H_V(\mathcal{Y}|\mathcal{X},g(R))$
\Comment{Predictive information is drop in test loss when including rater information}
\end{algorithmic}
\end{algorithm}
\vspace{-10pt}

\begin{table*}[h]
\centering
\small
\begin{tabular}{llllrrr}
\toprule
\multicolumn{1}{c}{\bf Dataset}  &\multicolumn{1}{c}{\bf Task}  &\multicolumn{1}{c}{\bf Choices}   &\multicolumn{1}{c}{\bf Dem.}   &\multicolumn{1}{c}{\bf Inst.}   &\multicolumn{1}{c}{\bf Raters}   &\multicolumn{1}{c}{\bf Ratings} \\
\hline
OpinionQA W27 (OQA) & Opinions (US) & 2-6 & 11 & 77 & 10k & 731k \\
Hatespeech-Kumar (HK) & Hate Speech & 2 & 18 & 19k & 864 & 37k \\
DICES (DIC) & Toxicity & 3 & 5 & 990 & 160 & 65k \\
ValuePrism (VP) & Moral Judgments & 3 & - & 31k & 4.5k & 199k \\
Habermas-Likert (HL)$^*$ & Opinions (UK) & 7 & 9 & 1.1k & 259 & 3.1k$^*$ \\
Prism (PR)$^*$
& Chat Preference & 2 & 20 & 8.0k & 1.4k & 8.0k$^*$ \\
\hline
\end{tabular}
% \small
% \small{
\caption{
Dataset statistics including task information, number of multiple choice options (Ch.), demographic variables (Dem), unique instances (\#I), unique raters (\#R), and total ratings (\#Rat). Datasets: OQA \citep{santurkar2023opinionslanguagemodelsreflect}, HK \citep{kumar2021designingtoxiccontentclassification}, DIC \citep{aroyo2023dicesdatasetdiversityconversational}, VP \citep{Sorensen_Jiang_Hwang_Levine_Pyatkin_West_Dziri_Lu_Rao_Bhagavatula_Sap_Tasioulas_Choi_2024}, HL \citep{habermas}, and PR \citep{kirk2024prismalignmentdatasetparticipatory}. Numbers may be smaller than in original datasets due to preprocessing/sampling (see \S\ref{app:reproducibility}).
$^*$Results are noisier for datasets with $<$10k ratings due to underfit models/small test sets.
}
\label{tab:dataset_statistics}
% }
\end{table*}

\section{Experimental Methodology}
\label{sec:experimental methodology}

\textbf{Training details }
We split raters into 50/50 train/test splits and report results for training/test runs on five random splits. We draw $|\mathcal{D}_i^{\text{fit}}| \sim \mathcal{U}(\{2,\ldots,|\mathcal{D}_i|-2\})$ to ensure that we have variable-sized fit/eval splits with at least two instances each. We train the decoder (\texttt{gemma2-9b-pt}, \citealt{gemmateam2024gemma2improvingopen}) for a single epoch (important for maintaining calibration, \citealt{ji2021earlystoppedneuralnetworksconsistent}).  For encoders, we use \texttt{gemma2-9b-it}, \texttt{gemma2-27b-it} \citep{gemmateam2024gemma2improvingopen}, and Gemini-1.5 Pro \citep{geminiteam2024gemini15unlockingmultimodal}. See App. \ref{app:reproducibility} for details.

\textbf{Datasets }
We utilize six datasets 
intended for research
(Table \ref{tab:dataset_statistics}) spanning tasks relevant to model alignment, content moderation, and computational social science. These datasets feature forced choice selection tasks and were selected to contain 1) some rater variation due to their subjective nature, 2) annotator IDs to link annotations from the same rater, and 3) ideally, some demographic information.
% (\S \ref{app:reproducibility})
Preprocessing information in \S \ref{app:reproducibility}.

\begin{figure*}[t]
\centering
\includegraphics[width=0.8\textwidth]{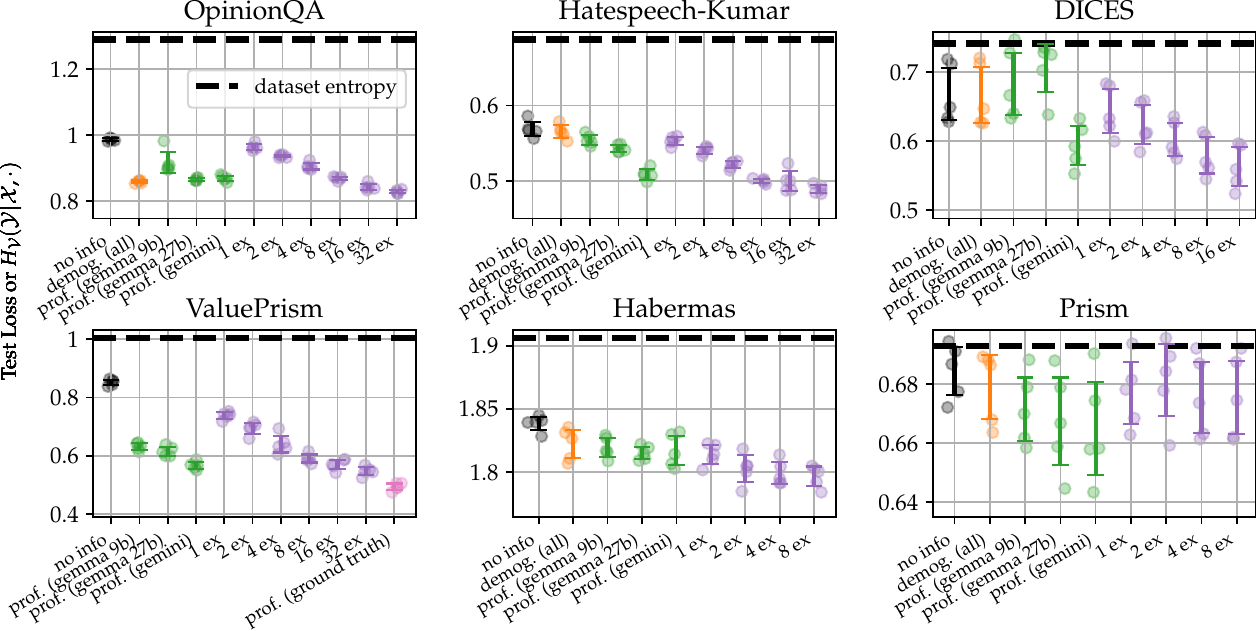}
% $ \vspace{-10pt}
% \small{
\caption{
Test losses across rater representation settings. Dashed line: label entropy $H(\mathcal{Y})$; {\bf no info}: $\emptyset$; \textcolor{profcolor}{\bf profile*}: value profiles $V$ generated by \texttt{gemma2-\{9/27\}b / Gemini-1.5-Pro}; \textcolor{demcolor}{\bf dem (all)}: $D$; \textcolor{excolor}{\bf N ex}: $E_N$, up to $N$ examples from $D_i^\textrm{fit}$. ValuePrism does not have demographics, but does have a \textcolor{gtcolor}{\bf ground truth value profile}. Each dot corresponds to a run with a differently seeded train/test split, with 95\% CI reported.
% Significance is reported with a mixed-linear effects model for a difference with "no info" (p $<$ .05/.01/.001).
Generally, in-context examples are more performant than value profiles, which are more performant than demographics.
}
% \vspace{-15pt}
\label{fig:nllresults}
% }
\end{figure*}

\section{Performance Across Rater Representation Settings}
\label{sec:results-ratererep}
Detailed results for held-out test losses across rater representations can be found in Figure \ref{fig:nllresults}. Accuracies can be found in App. \ref{fig:accresults}, but results mirror the loss results, which we will focus on.
% TODO - add back in
Detailed results for held-out test losses and accuracies across rater representations can be found in Figures \ref{fig:nllresults} and \ref{fig:accresults} respectively.

We note that error bars are much larger for 2 datasets, HL and PR. We believe that this is mainly because the datasets are smaller ($<$10k ratings), which means that 1) the trained model may be underfit and 2) there is a smaller sample size for each test split. We include results for all datasets for maximal inclusion, but focus our attention on the large datasets ($>$30k ratings: OQA, HK, DIC, VP) for which we can make higher confidence comparisons across settings.

Now, we compare decoder performance across rater representation settings (see Figs. \ref{fig:nllresults}, \ref{fig:rater_info},  \ref{fig:accresults}).
Our main findings are:

% \begin{wrapfigure}{R}{0.5\columnwidth}
\begin{figure}
\centering
\vspace{-5pt}  % Reduce extra space at the top
% \vspace{-10pt}  % Reduce extra space at the top
\includegraphics[width=0.75\columnwidth]{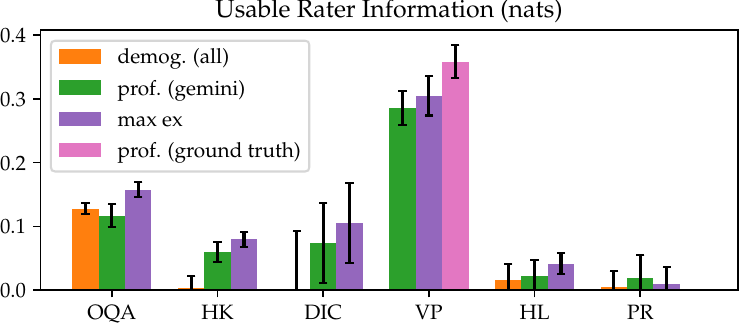}

\caption{
Usable rater information across datasets and rater representations (95\% CI).
}
\label{fig:rater_info}
% \vspace{-24pt}
% \vspace{-10pt}
\vspace{-10pt}
% \end{wrapfigure}
\end{figure}

\textbf{In-context examples improve predictions.} 
Across all four large datasets, providing the decoder with in-context examples of the rater's previous annotations significantly improved the prediction of their ratings on held-out test data in both accuracy and test loss ($p<.001$).
% , i.e.  $I_\mathcal{V}(E_n(\mathcal{R}) \to \mathcal{Y} \mid \mathcal{X}) > 0$) with $p < .001$
We observe a similar, but less significant, drop in loss/increase in accuracy on the two small datasets.
This shows that rater demonstrations offer useful information for disentangling human variation.

\textbf{Value profiles are highly predictive.} 
Value profiles generated by Gemini (version: 1.5-pro) consistently provided a significant performance boost across all four large datasets, suggesting value profiles contain useful information for modeling variation. Gemma2-9b and 27b value profiles also offered a significant boost for three of the four large datasets (VP, OQA, and HK), but not for DICES. In other words, as one might expect, value profiles improve with scale. As a result, we will focus the remainder of our experiments on the top-performing value profiles from Gemini.

\begin{figure}
\centering
% \vspace{-28pt}
% \vspace{-18pt}
\vspace{-5pt}
\includegraphics[width=0.7\columnwidth]{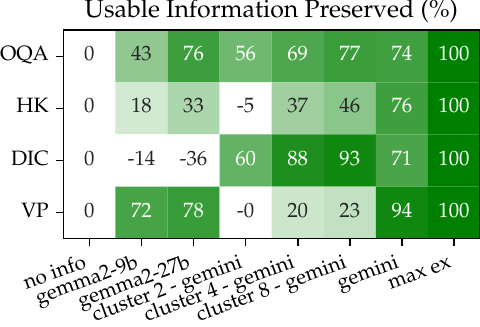}
% \vspace{-8pt}
\caption{
Info. preserved w.r.t. to using all examples. Results shown on the four large, low-variance datasets.
% Gemini value profiles preserve $>$70\% of usable information.
Gemini profiles preserve $>$70\% of usable information.
}
\label{fig:info_preserved}
% \vspace{-15pt}
\vspace{-15pt}
% \end{wrapfigure}
\end{figure}

\textbf{Value profiles effectively compress rater information.}
Since the value profiles are encoded from the same in-context examples used in the maximal example setting, we can exactly calculate the amount of decoder-usable information preserved (see Figure \ref{fig:info_preserved}):
\begin{equation}
\label{eq:info_preserved}
\frac{I_\mathcal{V}(V(E_n(\mathcal{R})) \to \mathcal{Y} \mid \mathcal{X})}{I_\mathcal{V}(E_N(\mathcal{R}) \to \mathcal{Y} \mid \mathcal{X})}
\end{equation}
% Value profiles effectively compressed the relevant information from in-context examples, preserving over 70\% of the usable information for the four large datasets. This indicates that value profiles are a concise and efficient way to represent human variation.
Value profiles effectively compressed the relevant information from in-context examples, preserving $>$70\% of the usable information for the four large datasets.
% This indicates that value profiles are a concise and efficient way to represent human variation.
This indicates that value profiles are an efficient way to represent human variation.

\textbf{Demographics have limited predictive power.}
Intersectional demographics generally offered a small and insignificant information boost, except for OpinionQA, where political affiliation was highly predictive.\footnote{It makes sense that demographic variables offer a boost for OQA as political affiliation (included in demographics) can be highly predictive for a political opinion survey.} Interestingly enough, however, the gains from demographic variables for other datasets were minimal. Additionally, value profiles contain more usable predictive information than demographics in all five datasets with demographics except OpinionQA (cf. Fig. \ref{fig:rater_info}). This suggests that demographics alone may not be sufficient to capture the full spectrum of human variation.

We also experiment with providing one demographic variable at a time (i.e., grouping by demographic) and providing value profiles and demographics together (cf. App. \ref{app:additionalexperiments}/Fig. \ref{fig:demographicresults}). As expected, single demographics provide less information than including all demographics. Also, demographics and value profiles can contain complementary information, with the best performing representation generally being demographics and value profiles together.

% \begin{wrapfigure}{R}{0.4\columnwidth}

\section{Value Profile Clustering for Grouping Raters}
\label{sec:clustering}

\begin{figure*}[ht]
\centering
\vspace{-5pt}
\includegraphics[width=0.85\textwidth]{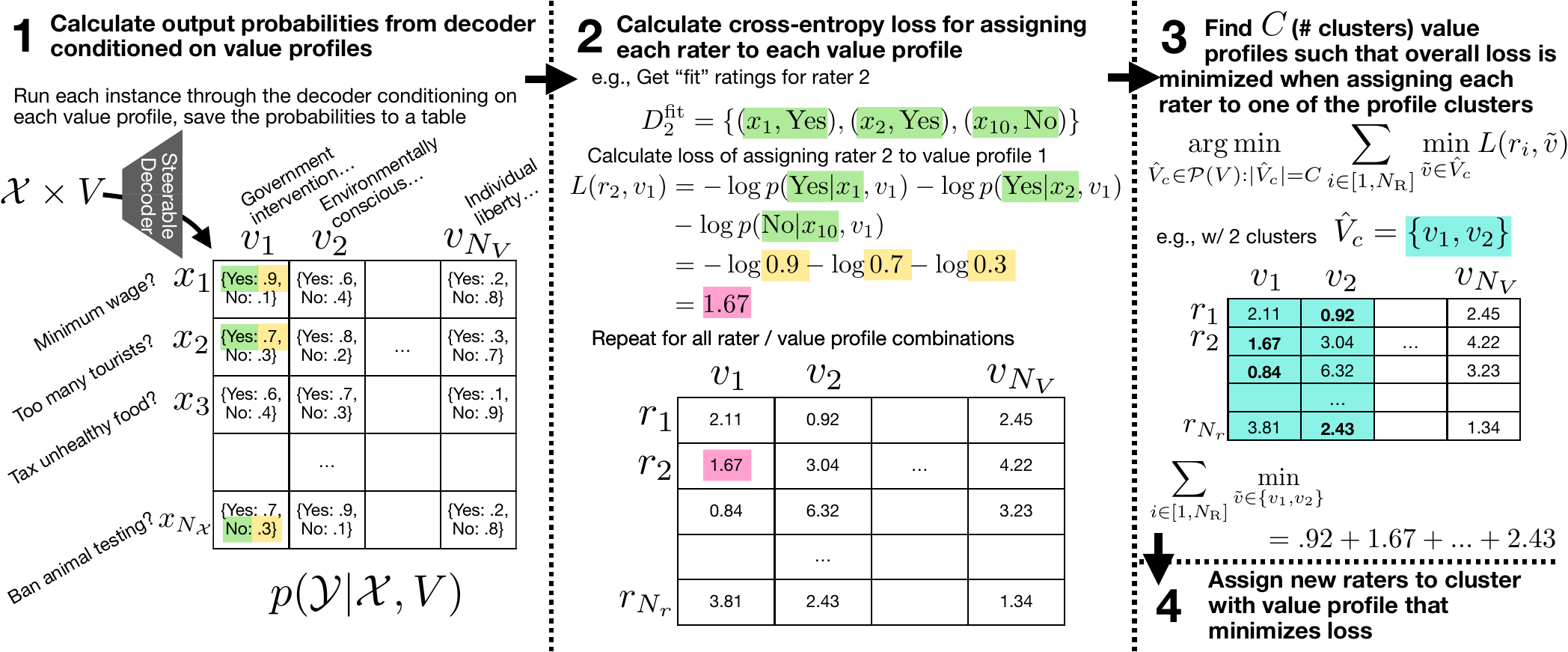}
\vspace{-5pt}
\caption{Clustering algorithm represented pictorially (also see Algorithm \ref{alg:cluster}).
1) The decoder predicts label distributions for each instance and value profile combination;
2) calculate the loss for predicting each rater's "fit" ratings with each value profile;
3) find $C$ (\# clusters) value profiles s.t. when each rater is assigned to a cluster, overall loss is minimized;
4) assign new raters to cluster with smallest loss on rater's train ratings.
}
\label{fig:clusteralgo}
\end{figure*}

To identify common modes of (dis)agreement, avoid over-personalization \citep{kirk2024benefits} and alleviate potential privacy concerns associated with inferring individual value profiles, we introduce a novel value profile-based rater clustering algorithm. Compared to traditional clustering methods, some advantages to our clustering method are that it: 1) does not require any overlap in instances seen by annotators; 2) is able to leverage semantic information between instances; 3) enables qualitative analyses through resulting cluster descriptions.

We assign the train raters to clusters using Algorithm \ref{alg:cluster} (cf. Figure \ref{fig:clusteralgo}), where each cluster corresponds to a single value profile description.\footnote{While we focus on value profiles as cluster candidates, one could also use cluster candidates such as in-context examples or preset groups. We focus on value profiles due to their interpretability.} We train a decoder to predict train rater annotations based on assigned cluster, and evaluate on held-out test raters. For all datasets, we use 100 randomly sampled value profiles as the cluster candidates. Results can be found in Figs. \ref{fig:info_preserved}/\ref{fig:clusteringresults} and the corresponding clusters can be found in Appendix \ref{app:profileclusters}.

\textbf{Clustering is effective and is suggestive of underlying modes of disagreement.}
Across all four large datasets, we observe a few common trends: 1) clustering into eight profile groups gives significant predictive improvement over no information, and 2) predictivity improves as we increase the number of clusters. Beyond this, we see some divergences.

For DIC and OQA, clustering is highly effective - using just two clusters preserves the \textit{majority} of the usable rater information (60\%/51\% respectively), and using eight clusters \textit{roughly matches} the performance of giving each rater their own profile. This suggests that perhaps most raters fall into one of very few ``modes" of agreement for these datasets, and that clustering based on value profiles is highly effective at finding these groupings.
For the other two large datasets, HK and VP, clustering preserves a significant amount of information ($\geq20\%$) but is not as predictive.
This implies either a failure to find the best clusters or that the underlying variation is inherently more difficult to categorize. Interestingly, this dataset divide coheres with our intuitions: e.g., for OpinionQA, ideology is highly explanatory and mostly centered around a few clusters, whereas ValuePrism includes a diverse set of 4k unique values which resist categorization. While it is epistemically difficult to totally disentangle a failure of our method to find correct groupings vs. a true difference in dimensionality of disagreement, we do find these results suggestive of profile clustering being able to tell us something interesting about the true underlying reasons for rater variation.

\begin{algorithm}[H]
\begin{algorithmic}
\caption{Value Profile Clustering}
\small
\label{alg:cluster}
% \Require \textbf{Input:}
\State \textbf{Input:}
\State $\bullet$
Decoder model $d : \mathcal{X},V \to \mathcal{P}(\mathcal{Y})$
\State $\bullet$
Candidate value profiles $V$%  with $N_v = |V|$, 
\State $\bullet$
Rater annotations for rater $i$: $R_i^{\text{fit}} = \{(x_j, y_{ij})\}$
\State $\bullet$
Target number of clusters $N_\text{cluster}$
\State $\bullet$
Initial clusters $C = [V_1, V_2, \ldots, V_{N_\text{cluster}}]$
\State $\bullet$
Maximum iterations $M_\text{iter}$

\medskip
\State $N_x \gets |\{x_j \text{ s.t. } \exists i, (x_j,\cdot) \in R_i^{\text{fit}}\}|$ \Comment{\# unique inst.}
\State Initialize $P \in \mathbb{R}^{N_x \times N_v \times |\mathcal{Y}|}$ \Comment{Fill in output probabilities conditioned on each value profile}
\For{$j \in [1,\ldots,N_x]$} \Comment{For each instance}
\For{$k \in [1,\ldots,N_v]$} \Comment{For each value profile}
\State $P[j,k] = d(x_j, v_k)$ \Comment{Prob. dist. over $\mathcal{Y}$ from $d$ conditioned on instance $x_j$ and profile $v_k$}
\EndFor
\EndFor

\medskip
\State $L(r_i, v_k) \gets \sum_{(x_j,y_{ij}) \in R_i^{\text{fit}}} -\log P[j,k,y_{ij}]$ \Comment{Total loss from assigning rater $r_i$ to profile $v_k$}

% TODO - add back in for camera ready
% \State {// Approximately find the $N_\textrm{cluster}$ value profiles such that loss is minimized when assigning each rater to one of the profiles}
% \State{// e.g., $\underset{\hat{C} \in \mathcal{P}(V) : |\hat{C}| = N_\textrm{cluster}}{\arg\min}
% \sum_
% % {i \in [1, N_\text{R}]}
% {r_i \in R_i^\textrm{fit}}
% \min_{\tilde{v} \in \hat{C}}
% L(r_i,\tilde{v})$
% }
% \medskip
\State converged $\gets$ False; iter $\gets$ 0; $C_\text{last} \gets C$
\While{$\text{iter} < M_\text{iter} \text{ \& not converged}$}
\For{$c \in [1,\ldots,N_\text{cluster}]$} \Comment{Fixing all profiles except $c$, greedily find best profile to replace $c$}
\State $C[c] \gets$ \Comment{New cluster that minimizes loss}
\State $\underset{\hat{v}_c \in V}{\arg\min} 
\underset{i \in [1, N_\text{R}]}{\sum}
\underset{\tilde{v} \in (C/\{\hat{v}_c\} \cup \{\tilde{v}_c\})}{\min}
L(r_i,\tilde{v})$
\EndFor
\State converged $\gets C = C_\text{last}$; $C_\text{last} \gets C$
\EndWhile
\State \textbf{Output:} Clusters C, assignments
$\underset{\tilde{v} \in C}{\arg\min}
L(r_i,\tilde{v})$

\end{algorithmic}
% TODO - add back in for camera ready
% \small{
% \textbf{Note:} The most expensive portion in the clustering algorithm is running the $N_V \times N_X$ decoder inferences - finding the clusters only requires lookups in table $P$. Thus, the biggest determiner of runtime is the number of unique instances $N_X$ and the number of value profiles considered $N_V$.
% }
\end{algorithm}
\vspace{-5pt}

\begin{figure*}[t]
\centering
\includegraphics[width=0.8\textwidth]{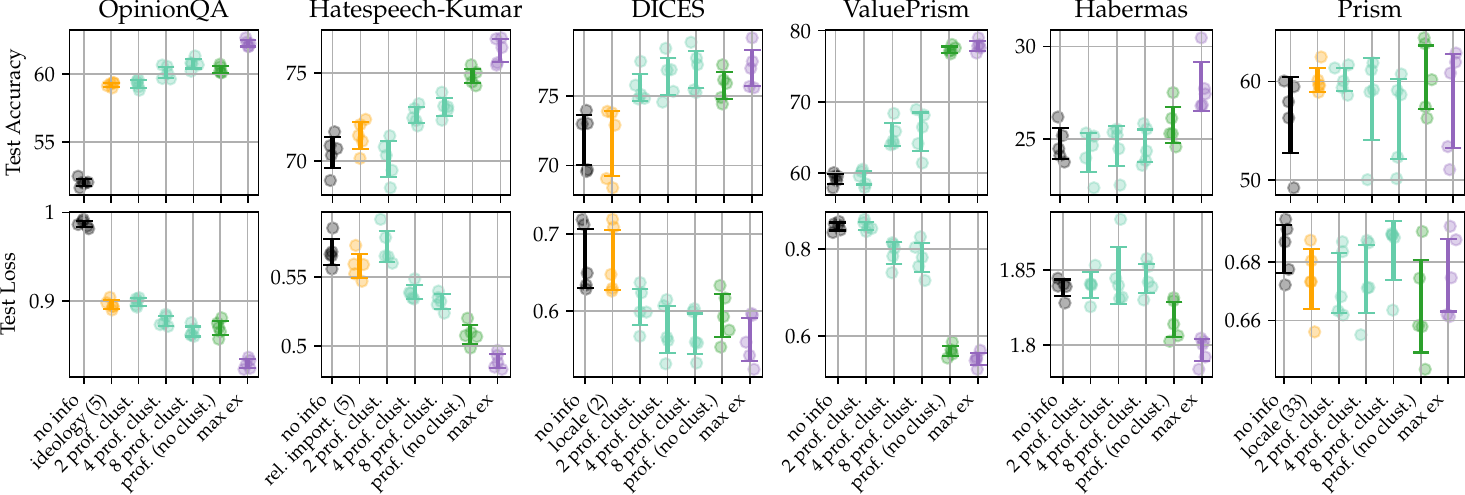}
\caption{
Performance after clustering raters into \textcolor{clustcol}{\bf 2/4/8 profile clusters} alongside the \textcolor{demcolor}{\bf best performing categorical demographic grouping}, with the \# of groups in parentheses.
Value profile clustering is highly effective, outperforming the best demographic grouping of comparable size.
}
\label{fig:clusteringresults}
\end{figure*}

\textbf{Profile clusters are more predictive than the best demographic groupings.}
Next, we compare with the best performing demographic clusters, grouping people who gave the same demographic response to a categorical demographic question (e.g., people in the same country for DIC or same political ideology for OQA). We compare specifically across the three large datasets w/ demographics: for DIC, the two profile clusters is more predictive than grouping based on country; for HK/OQA, the four profile clusters outperform grouping by religiosity/ideology respectively. In other words, clustering by value profiles is able to outperform the most performant
% unidimensional
demographic clusters when using the same or fewer number of groupings.

% \begin{wrapfigure}{R}{0.45\columnwidth}
\begin{figure}
% \vspace{-16pt}
\centering
\includegraphics[width=0.32\textwidth]{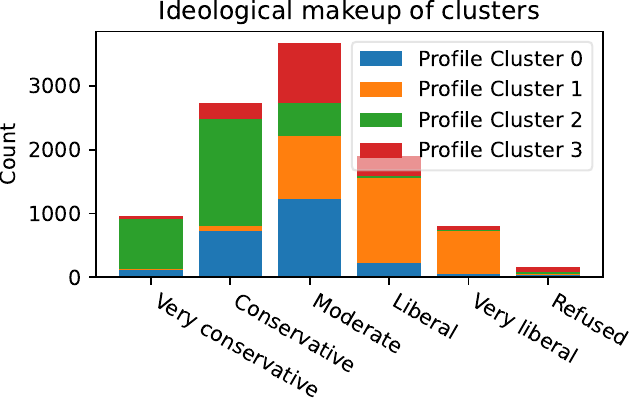}
% \vspace{-20pt}
\vspace{-5pt}
\caption{Ideological makeup of the raters sorted into each value profile cluster for OpinionQA. The clusters recover strong ideological trends.}
\label{fig:ideology}
\vspace{-6pt}
% \end{wrapfigure}
\end{figure}

\textbf{Where predictive, demographic groupings closely match profile clusters.}
Focusing in on the two datasets where clustering was most effective, OQA and DIC, we see if there are any demographic trends related with clustering. As you can see from Figure \ref{fig:ideology}, there are strong demographic trends in the OQA clusters - cluster two consists almost exclusively of self-described conservative individuals, while cluster one consists of mostly self-described liberal individuals. In other words, despite not having access to demographics, the value profiles are able to largely reconstruct the most explanatory demographic groupings. Meanwhile, for the DIC four-profile clusters, the clusters cut across almost uniformly across all demographic groupings (Figure \ref{fig:dicesdem}). This suggests that for DIC, the most important dimensions of variation are not found in the demographic groupings.

\textbf{Cluster descriptions qualitatively describe modes of disagreement.}.
The profile clustering algorithm returns not only clustering assignments, but also a single corresponding value profile for each cluster (see App. \ref{app:profileclusters}).
For DICES, even two clusters were quite predictive. The corresponding value clusters relate to overall sensitivity to toxicity (e.g., profile 1: "High tolerance for offensive language"; "Narrow definition of toxicity" vs. profile 2: "Strong reaction to overt negativity, "Sensitivity to potential harm"). Meanwhile, when going to four clusters, more nuance enters in (e.g., "Context and intent matter"). In other words, it seems that 1) overall sensitivity to toxicity is an important dimension in explaining variation, and 2) there are clusters of people who hold more nuanced views.
% For OpinionQA, descriptors that have to do with politics are often used (e.g., "Economically Conservative, but Populist on Trade"). For HK, which required more clusters to effectively predict, there are many specific phrases about what kinds of things the rater may or may not find offensive (e.g., "Profanity tolerance"; "Discomfort with stereotyping"; "Acceptance of strong opinions"; etc.). Meanwhile, for PR the clusters center around potentially conflicting chatbot preferences (e.g., "Appreciates simplicity" vs. "Appreciates nuanced and comprehensive answers").
For OpinionQA, descriptors that have to do with politics are often used (e.g., "Economically Conservative, but Populist on Trade"). For HK, which required more clusters to effectively predict, there are many specific phrases about what kinds of things the rater may or may not find offensive (e.g., "Profanity tolerance"; "Discomfort with stereotyping"; etc.). Meanwhile, for PR the clusters center around potentially conflicting chatbot preferences (e.g., "Appreciates simplicity" vs. "Appreciates nuanced and comprehensive answers").
Because value profiles are interpretable (see \S \ref{sec:interpretability}) and can recover demographic groupings where predictive, we have reason to believe that these qualitative differences map to important dimensions of disagreement for a dataset.

\section{Extrinsic Evaluation}
\label{sec:extrinsicevaluation}

In the previous sections, we have established that value profiles are predictive of individual rater annotations for a wide variety of relevant tasks, based on intrinsic performance metrics. 
We are now assessing value profiles within the context of wider real-world applications. We show that value profiles are interpretable -- which is important for enabling control by the end-user; their predictions are steerable and well-calibrated -- which enables pluralistic AI alignment; and they are reliable for extrinsic tasks in the context of computational social science, such as simulating a rater population.

\textbf{Value profiles are interpretable.}\label{sec:interpretability}
We first explore interpretability --
i.e.,
% TODO - add back in
% in other words,
do the value descriptions change the decoder outputs in a common-sense manner? Because the encoder is prompted and only the decoder is trained (cf. Section \ref{sec:autoencoder}), we believe that this serves as a strong regularization so that the value profiles correlate with held-out ratings only by the natural language
values described. To ensure that this is the case, we test the interpretability of the autoencoder as follows:
1. For each instance and 100 value profiles, we get the estimated output distribution for the decoder.
2. We select the value profiles that have the largest Jensen-Shannon divergence.
3. We create
a
% the following
binary classification task: given an instance and two value profiles, which profile corresponds to which estimated output distribution?
If the change in distribution is not correlated with common sense, we would expect 50\% accuracy, while performance would approach 100\% if an observer is always able to
match value profiles to corresponding distributions.
% tell which value profile goes with which distribution.

\begin{wraptable}{r}{0.42\columnwidth}
\vspace{-20pt}
% \begin{table}[]
% \vspace{-18pt}
\setlength{\tabcolsep}{0pt}  % Default is 6pt
\centering{
% \small
\begin{tabular}{lc}
Data & Accuracy (\%) \\
\toprule
OQA & 94.8 ($\pm 2.5$) \\
HK & 96.3 ($\pm  .6$) \\
DIC & 95.5 ($\pm .7$) \\
VP & 91.7 ($\pm .9$) \\
HL & 89.8 ($\pm .9$) \\
PR & 80.0 ($\pm .3$) \\
\hline
Chance & 50 \\
\bottomrule
\end{tabular}
}
\caption{
Profiles are semantically interpretable (95\% CI).
}
\label{tab:model_results}
% \vspace{-18pt}
\vspace{-8pt}
\end{wraptable}

As an example of the task, here is a shortened example from VP: \textit{Is it moral, immoral, or morally dependent on context to do this: ``Choosing not to get a vaccine." Profile 1: Prioritization of collective good over individual needs. Profile 2: Strong belief in individual liberty and autonomy. X probabilities: Moral: 92\%, Context-dependent: 7\%, Immoral: 2\%. Y probabilities: Immoral: 94\%, Context-dependent: 3\%, Moral: 3\%. Which profile goes with the X probabilities?} Correct answer in footnote.~\footnote{Profile two is the correct answer.}

We report accuracies for a zero-shot prompted Gemini in Table \ref{tab:model_results}. Accuracies range from 80-96\% across all datasets, demonstrating that variation in outputs from value profiles are explainable by their plain natural language
(p$<$.001).
% TODO - add back in
% (p$<$.001 for all datasets).

% \begin{wrapfigure}{r}{0.5\columnwidth}
\begin{figure}
% \vspace{-10pt}
\centering
\includegraphics[width=0.8\columnwidth]{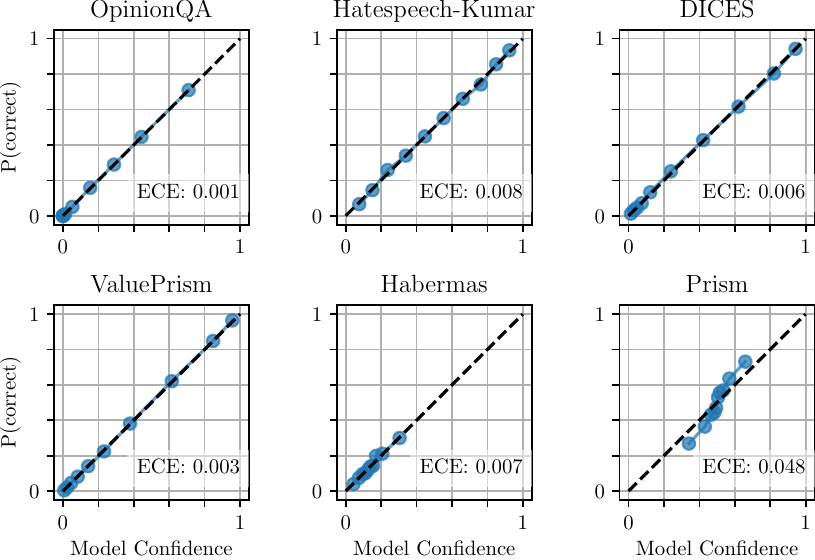}
% \vspace{-15pt}
% \small
\caption{Calibration plots for value profile decoders. (Perfect calibration $=$ dotted line).
The decoders are very well-calibrated.
}
\label{fig:calibration}
% \vspace{-10pt}
% \vspace{-5pt}
% \end{wrapfigure}
\end{figure}

\textbf{Decoders are well-calibrated.}
Decoder calibration is important for two principal reasons.  Firstly, an appropriately calibrated decoder allows us to trust the model confidence w.r.t. error rate. Secondly, even raters with shared values may have varied outputs - a well-trained decoder would model this distribution appropriately.
% TODO - add back in
% In other words, calibration is important for epistemic uncertainty of the decoder and the aleatoric uncertainty of the population who holds those values.
Calibration plots for the value profile decoders can be found in Figure \ref{fig:calibration}.
% TODO - add back in
% ~\footnote{ We observe similarly well-calibrated decoders for other rater representations.}
The trained decoders are quite well-calibrated, suggesting that we can generally trust the decoder's output confidence.
% TODO - add back in
% Proper calibration allows for exciting potential applications, such as disentangling  value-related epistemic uncertainty from aleatoric uncertainty in rater variation. (See App. \ref{app:applications} for more discussion.)

\begin{figure}
% \vspace{-8pt}
\centering
% \small
\includegraphics[width=0.8\columnwidth]{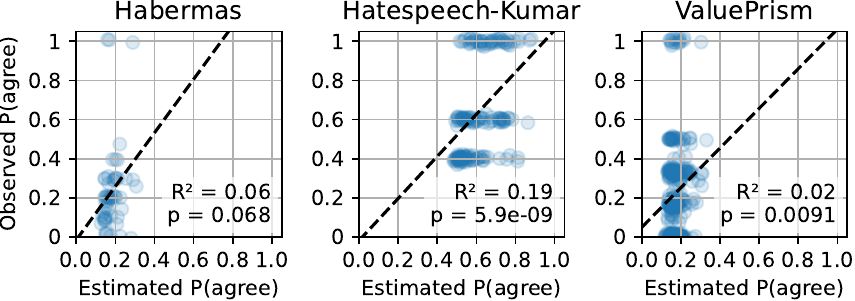}
% \vspace{-15pt}
\caption{Instance-level observed vs. estimated inter-annotator agreement (as the probability that two raters agree).
The predicted simulated agreement correlates with the observed agreement.
}
\label{fig:agreement}
\vspace{-10pt}
% \end{wrapfigure}
\end{figure}

\textbf{Simulating an annotator population with value profiles.}
Given a trained decoder and a set of value profiles, one can simulate a 
% TODO - add back in
Given a trained decoder and a set of value profiles, one can simulate a population -- or ``jury" \citep{Gordon_2022} -- of annotators on novel instances. While one can do many things with such a simulated population \citep{park2023generativeagentsinteractivesimulacra, Gordon_2022}, one experiment is to predict which instances raters would have higher or lower inter-annotator agreement (IAA).

In order to calculate out-of-distribution IAA, we first eliminate the datasets where annotators have no overlap (PR) and for which all raters annotated the same instances (DIC, OQA). For each instance, we then sample 100 value profiles that were not fit on that instance, and calculate the estimated probability of agreement between those annotators (assuming each rater annotates at random from the decoder's output distribution). We also filter to instances that were labeled by a minimum number of annotators (See \S\ref{app:reproducibility} for details). We then compare to the actual observed probability that two raters agree on the instance (see Figure \ref{fig:agreement}). For all three datasets, there is a positive correlation between the estimated and observed IAA, and this correlation is significant ($p<.001$) for HK and VP. While not much variance is explained ($R^2 < .2$), the observed P(agree) is a high variance estimate with few ($\sim 5$) raters per instance. In summary, a simulated population with value profiles provides some explanatory power at predicting inter-annotator agreement, but is not yet a high precision tool.

\section{Related Work} \label{sec:relatedwork}

\textbf{Clustering and demographics} While aligning to groups can increase agreement \citep{chen2024spicaretrievingscenariospluralistic}, it also has been shown to flatten intra-group variation \citep{orlikowski-etal-2023-ecological, Wang2025} lead to stereotyping \citep{cheng-etal-2023-marked}, or simply not be correlated with subjective NLP tasks \citet{orlikowski2025demographicsfinetuninglargelanguage}. 
% Prior work proposes clustering individuals by responses \citep{vitsakis2024voicescrowdsearchingclusters,li-etal-2024-steerability} and similarly finds that clusters cut across demographic groups.
% Prior work explores other methods for clustering individuals by responses \citep{vitsakis2024voicescrowdsearchingclusters,li-etal-2024-steerability} and similarly finds that clusters cut across demographic groups.
Prior work explores embedding-based methods for clustering individuals by responses \citep{vitsakis2024voicescrowdsearchingclusters,li-etal-2024-steerability} and similarly finds that clusters cut across demographic groups.
% Additionally, unlike these methods, our value profile clustering method does not require that annotators label the same instances.
Beyond predictivity, demographics can still be important to collect for evaluating group fairness \citep{aguirre2023selectingshotsdemographicfairness, kirk2024prismalignmentdatasetparticipatory}.

\textbf{Steering to individuals} Prompted large language models (LLMs) 
have been used
% to simulate human judgments such as:
to simulate human judgments, e.g.:
NLP task annotators \citep{Bavaresco:2024}, political survey respondents \citep{argyle_busby_fulda_gubler_rytting_wingate_2023}, fact-checking labels \citep{de2024supernotes}, or human attitudes and behavior \citep{park2024generativeagentsimulations1000}. Textual user profiles have also been proposed for personalizing chats \citep{zhang-etal-2018-personalizing}.
\citet{hu2024quantifyingpersonaeffectllm} similarly use textual demographic descriptions and find that they provide small, but statistically significant, gains in explaining human variation.
Many works focus merely on prompting an existing LLM, while our work explicitly trains an LLM to better match varied perspectives (as in \citealt{Gordon_2022, jiang2024languagemodelsreasonindividualistic}). Encoding individual information from demonstrations is also analogous to behavioral user modeling for recommender systems \citep{Radlinski:2022,ramos-etal-2024-transparent}.
\citet{poddar2024variationalpreference} also use an autoencoder to model human variation, but focus on preference data and use a vector-valued latent space instead of natural language.

\textbf{Values and alignment }
Similarly to natural language value profiles, \citet{bai2022constitutionalaiharmlessnessai}'s "constitutional AI" train models to follow textual principles, although they focus on a single set of principles. \citet{findeis2024inverseconstitutionalaicompressing} propose to learn preference principles directly from preference data (similar to our encoder setup). Values have also been normatively proposed as an alignment target \citep{gabriel2020artificial, klingefjord2024humanvaluesalignai}, and pluralistic alignment \citep{sorensen2024roadmappluralisticalignment} seeks to align AI systems to diverse values.

\section{Conclusion}
\label{sec:conclusion}
In conclusion, we proposed modeling human variation via natural language \textit{value profiles}. We also proposed a methodology to compare the usable information in various rater representations, and found that value profiles contain more information than demographics. Prompted LLMs serve as effective value encoders, retaining $\geq 70\%$ of the useful rater information from demonstrations. In addition, we introduced a profile clustering algorithm which is able to find more explanatory clusters of raters than grouping by the most predictive demographics.
Finally, we showed that value profiles are extrinsically useful for interpretability, steerability, and for simulating diverse populations, hence offering new ways to describe individual variation beyond demographics.

\newcounter{futurework}
\setcounter{futurework}{0}
\newcommand{\nextitem}{\stepcounter{futurework}\thefuturework)}

Some promising avenues for future work include:
\nextitem{} qualitative analyses extracting values from data;
\nextitem{} fairness analysis of who can (or cannot) be well-represented by value profiles;
\nextitem{} study on sensitivity of value profiles to choice of and number of ratings;
\nextitem{} extensions to additional models and datasets;
\nextitem{} how decoders handle conflicting value information in a profile;
\nextitem{} optimization of the encoder (e.g. via ELBO);
and 
\nextitem{} human evaluations to see how well represented people feel by value profiles.

\section*{Acknowledgements}
We would like to thank Lora Aroyo, Laura Weidinger and Ahmad Beirami for helpful discussions. We would also like to thank Charvi Rastogi and Jared Moore for useful draft feedback.

\newpage

\section{Limitations}
\label{sec:limitations}
We have tried to test for generalization across six tasks and datasets and more than twenty demographic distributions, However, all of the experiments use the Gemma-2 \citep{gemmateam2024gemma2improvingopen} and Gemini \citep{geminiteam2024gemini15unlockingmultimodal} families of models. This is due in part due to the TPU hardware available to us and because of the expense of the experiments (more than 650 training runs). We have no reason to think that our results are due to anything particular about these families of models though, and prior work doing similar experiments on demographics with other models has reached similar results \citep{orlikowski-etal-2023-ecological, hwang2023aligninglanguagemodelsuser}. That being said, future work could benefit from experiments on more model families.

\section{Ethical Considerations}
\label{sec:ethicalconsiderations}

We seek to improve AI systems' ability to model diverse values out of a hope that the systems can be more inclusive, better represent a range of viewpoints, and better serve a wider population. Here, we explore benefits and risks of modelling variation with value profiles.

\textbf{Profiling risks}
One of the potential ethical risks of the value profile through an autoencoder paradigm is in the name: ``profile." Value profiles are, inherently, guesses about the underlying values that people may hold that lead them to annotate in the way that they do. There are potential privacy concerns here - people may not wish to have their underlying values exposed \citep{tomasev-etal}.
It may be better if people had agency to create their own value profiles through some voluntary elicitation process \citep{park2024generativeagentsimulations1000}.
% TODO - add back in
% It may be better if people had agency to create their own value profiles through some elicitation process, such as qualitative interviews \citep{park2024generativeagentsimulations1000}, such that people can choose exactly what value-relevant information to expose.

\textbf{False generalization}
On the one hand, value profiles are an attempt to reduce the (often false) generalization inherent when grouping e.g. by demographic groups \cite{dev-etal-2022-measures}, improving on widely used current techniques. On the other hand, generalization risks remain. For example, there may be multiple possible underlying values that could support a set of rater annotations. In the absence of additional information, the value encoder may arbitrarily assign a guess to the underlying values.
Also, our experiments focus on English-language value profiles, so generalization to other languages is unknown.
% TODO - add back in
% Also, if people hold values in ways that are not commonly understood/represented by natural language, we would expect performance degradation. may lead to false generalization about a rater.
Additionally, there is always a risk of misrepresentation when using simulated human ratings in place of actual ratings at all \citep{agnew:2024}.

\textbf{Interpretability, understandability, and user agency}
However, there are also several positive attributes to value profiles. First of all, they are interpretable - and therefore, potentially more understandable to a user. While people interact with many technologies today that are trying to model their behavior and preferences, most such systems \textit{do not} break down their user representation into a format that as as easy to understand as a textual description.
% Additionally, this makes value profiles intervenable - people could \textit{change} how they choose to be represented. For example, one could imagine that after a value profile is inferred for a person, they would have the chance to change the value profile, as suggested by \cite{Radlinski2019,lazar2024moralcaseusinglanguage} for personalized recommender systems.
Additionally, this makes value profiles intervenable - people could \textit{change} how they choose to be represented \cite{Radlinski2019,lazar2024moralcaseusinglanguage}.
Relatedly, value profiles serve as a step towards explainable AI \citep{arrieta2019explainableartificialintelligencexai, koh2020conceptbottleneckmodels} for human variation.

\textbf{Enabling value reflection}
Learning values from data, while allowing users to modify the values, is loosely related to John Rawls' concept of reflective equilibrium \citep{rawls2005political, sep-reflective-equilibrium}: ratings are akin to judgments, and value profiles are an attempt to draw general ``principles" out of the judgments in a bottom-up manner. Meanwhile, a user can then edit the value profile, applying top-down reflection on whether the values/principles are ones that the person would endorse.
In this light, perhaps ``value profiles" could help a person to explore their own value system, both in the values that their decisions may imply and considering which values they would reflexively endorse.
% TODO - add back in
% In this light, perhaps ``value profiles" could help a person to explore their own value system, both in the values that their decisions seem to imply that they may have and considering which values they would reflexively endorse.

\textbf{"Chosenness" of values}
Many works modeling diversity focus on socio-demographics. However, many demographics are not a result of a person's agency, but rather a product of unchosen life factors - for example, the country in which one is born, or the economic opportunities available to them. Meanwhile, while the values that one holds can certainly be affected by unchosen factors \citep{Nguyen2024-NGUVCH}, values can also be chosen for oneself.
% Thus, in the spirit of luck egalitarianism \citep{Dworkin2002-DWOSVT, sep-justice-bad-luck} it may be more justifiable to represent someone using values that they reflexively endorse, as opposed to boxing them in to the characteristics of a group they may not have chosen.
Thus, in the spirit of luck egalitarianism \citep{Dworkin2002-DWOSVT} it may be more justifiable to represent someone using values that they reflexively endorse, as opposed to boxing them in to the characteristics of a group they may not have chosen.

\textbf{Importance of demographics for fairness}
Also, while demographics may not be the most ideal rater representation in many cases for the above reasons, it can still be important to collect demographic information for other worthwhile goals, such as fairness/evaluating group disparities, ensuring representation, etc.

\newpage

\newpage

\bibliography{custom}

\newpage

\appendix

% reset figure counter for appendix
\setcounter{figure}{0}
\renewcommand{\thefigure}{A\arabic{figure}}

% \section{appendix}
\section*{Appendix - Table of contents}
\begin{itemize}
% \item Appendix \ref{app:limitations}: Limitations
% \item Appendix \ref{app:additionalfigures}: Optional additional figures
\item Appendix \ref{app:reproducibility}: Reproducibility details
\item Appendix \ref{app:modellingvariation}: More on approaches to modelling variation.
\item Appendix \ref{app:additionalexperiments}: Additional experiments.
\item Appendix \ref{app:applications}: Discussion of potential applications and extensions
\item Appendix \ref{app:prompts}: The prompts used for the encoders and decoders.
% \item Appendix \ref{app
% \item Appendix \ref{app:ambiguity}: Example profiles for which conditioning on values gave a high or low information boost on a rater's answer.
\item Appendix \ref{app:data}: Preprocessing and demographic information for all datasets.
\item Appendix \ref{app:bigresults}: Full results for all experiments across all rater representations.
\item Appendix \ref{app:profileclusters}: The profile clusters found and used in the profile cluster experiments.
% \item Appendix \ref{app:r}: The profile clusters found and used in the profile cluster experiments.
\item Appendix \ref{app:random-profile-samples}: Ten random value profiles for each dataset and model (gemma2-9b, gemma2-27b, gemini).

\end{itemize}

% \section{Optional Additional Figures}
% \label{app:additionalfigures}
% 
% Here, we include additional, optional figures which can be used to better illustrate parts of the paper.
% 
% \begin{figure*}[t]
% \centering
% \includegraphics[width=0.9\textwidth]{files/raterreplarge-cropped.pdf}
% \small
% % \vspace{-13pt}
% \caption{Rater representations and example corresponding decoder prompts
% ($\emptyset$, $D$, $V$, $E_n$).
% The decoder predicts the rater's annotation given the rater representation.
% }
% \label{fig:raterrep}
% % \vspace{-2pt}
% \end{figure*}

% \begin{figure}
% \centering
% % \vspace{-40pt}
% \includegraphics[width=0.4\textwidth]{files/annotated_example_v2.pdf}
% \small
% 
% \caption{
% An illustrative plot on fictional data for measuring $\mathcal{V}$-info.
% % Information content is the model's decrease in test loss when given access to the information.
% }
% \label{fig:example}
% % \vspace{-20pt}
% \end{figure}

% \begin{figure*}[ht]
% \centering
% % \vspace{-5pt}
% \includegraphics[width=0.9\textwidth]{files/clusterfig.pdf}
% \small
% % \vspace{-15pt}
% \caption{Clustering algorithm represented pictorially (also see Algorithm \ref{alg:cluster}).
% 1) The decoder predicts label distributions for each instance and value profile combination;
% 2) calculate the loss for predicting each rater's "fit" ratings with each value profile;
% 3) find $C$ (\# clusters) value profiles s.t. when each rater is assigned to a cluster, overall loss is minimized;
% 4) assign new raters to cluster with smallest loss on rater's train ratings.
% }
% \label{fig:clusteralgo}
% \end{figure*}

\section{Reproducibility Details}
\label{app:reproducibility}

% % \newcommand{\clustWidth}{0.48\textwidth}
% \newcommand{\clustWidth}{0.52\textwidth}
% \begin{wrapfigure}{R}{\clustWidth}
% \vspace{-50pt}  % Reduce extra space at the top
% \begin{minipage}{\clustWidth}  % Match the width of wrapfigure
% \centering
% \vspace{-50pt}  % Reduce extra bottom space
% \end{minipage}
% \end{wrapfigure}

Here, we include additional experimental details to aid reproducibility.

\textbf{Dataset Preprocessing}
We carried out the following preprocessing steps for the datasets -
DIC: used the larger subset (990);
HL: selected raters with at least nine responses;
HK: randomly selected 5k raters and binarized annotations;
OQA: randomly selected a wave for experiments (Wave 27);
PR: select annotations from first conversation turn and compared the chosen response to the next highest rated response;
VP: Treat each value, right, or duty as a unique annotator.
Finally, for all datasets we filtered to annotators that had at least four responses.

\textbf{Decoder hyperparameters}: model: \texttt{gemma2-9b-pt} \citep{gemmateam2024gemma2improvingopen}, batch size: 4, learning rate: 1e-7, gradient clipping: 50.

\textbf{\texttt{fp32} unembedding layer}:
Gemma 2 \citep{gemmateam2024gemma2improvingopen} natively uses \texttt{bf16}. However, we found that this caused heavy quantization among high-probability logits (e.g., the valid responses). As such, we cast the embedding/unembedding parameters to \texttt{fp32} before training, which allowed for higher precision distributions, important for calibration and expressivity.

\textbf{Fit/eval partition details}: For each rater $r_i$, 
we draw $|\mathcal{D}_i^{\text{fit}}| \sim \mathcal{U}(\{2,\ldots,|\mathcal{D}_i|-2\})$ and set $|\mathcal{D}_i^\text{eval}| = |\mathcal{D}_i| - |\mathcal{D}_i^\text{fit}|$ to ensure that  we have variable-sized fit/eval splits with at least two instances each. Value profile encoders use all $D_i^\text{fit}$ instances and the decoders with in-context information $E_n$ use the first $\min(n, |D_i^\text{fit}|)$ examples from $|D_i^\text{fit}|$. This means that value profiles are fit with a variable number of ratings.

\textbf{Simulating an annotator population instance selection}: We selected the minimum number of instances per dataset as roughly the median number of annotations per instance: 3 for HL, 5 for HK, and 5 for VP. This was selected to try to ensure 1) that we had as many instances as possible and 2) that we had enough raters to have a high-precision estimate of actual rater agreement.

\section{More on Approaches to Modelling Variation}
\label{app:modellingvariation}

\begin{figure*}[ht]
\centering
\includegraphics[width=\textwidth]{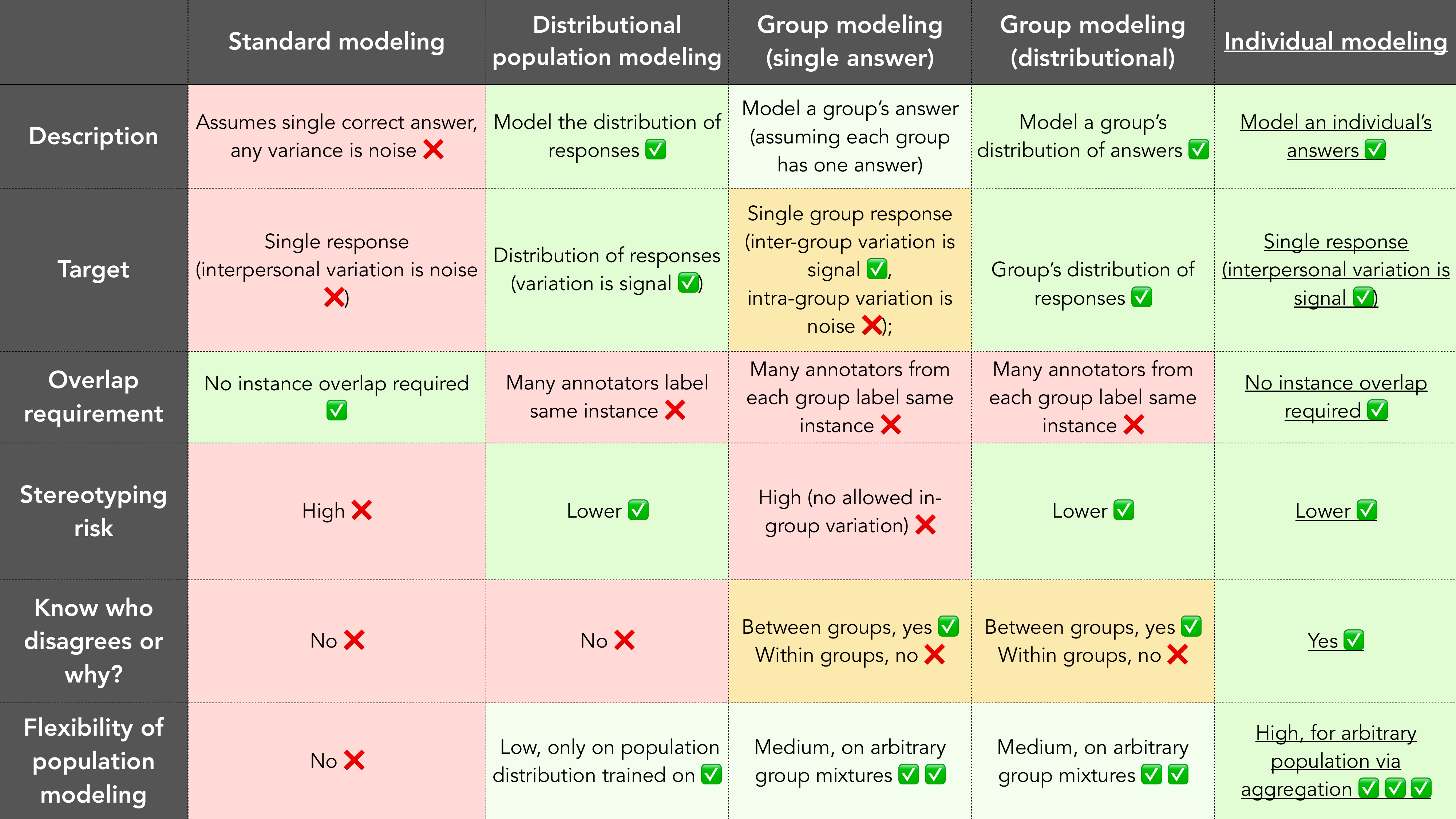}
\caption{Comparisons of various modelling approaches and their tradeoffs with respect to modelling variation}
\label{fig:modellingvariation}
\end{figure*}

In Figure \ref{fig:modellingvariation}, we flesh out more of the comparisons between various modelling approaches characterized in \S \ref{sec:introduction}.

% \begin{itemize}
%     \item Target: what is the learning target? For standard modeling
%     \item Overlap requirement:
%     \item Stereotyping risk:
%     \item Know 
% \end{itemize}

\section{Additional Experiments and Results}
\label{app:additionalexperiments}

% \subsection{Additional Figures}

% \begin{wrapfigure}{r}{0.4\columnwidth}
% \begin{figure}[0.9\columnwidth]
\begin{figure}
\centering
% \vspace{-40pt}
\includegraphics[width=0.4\textwidth]{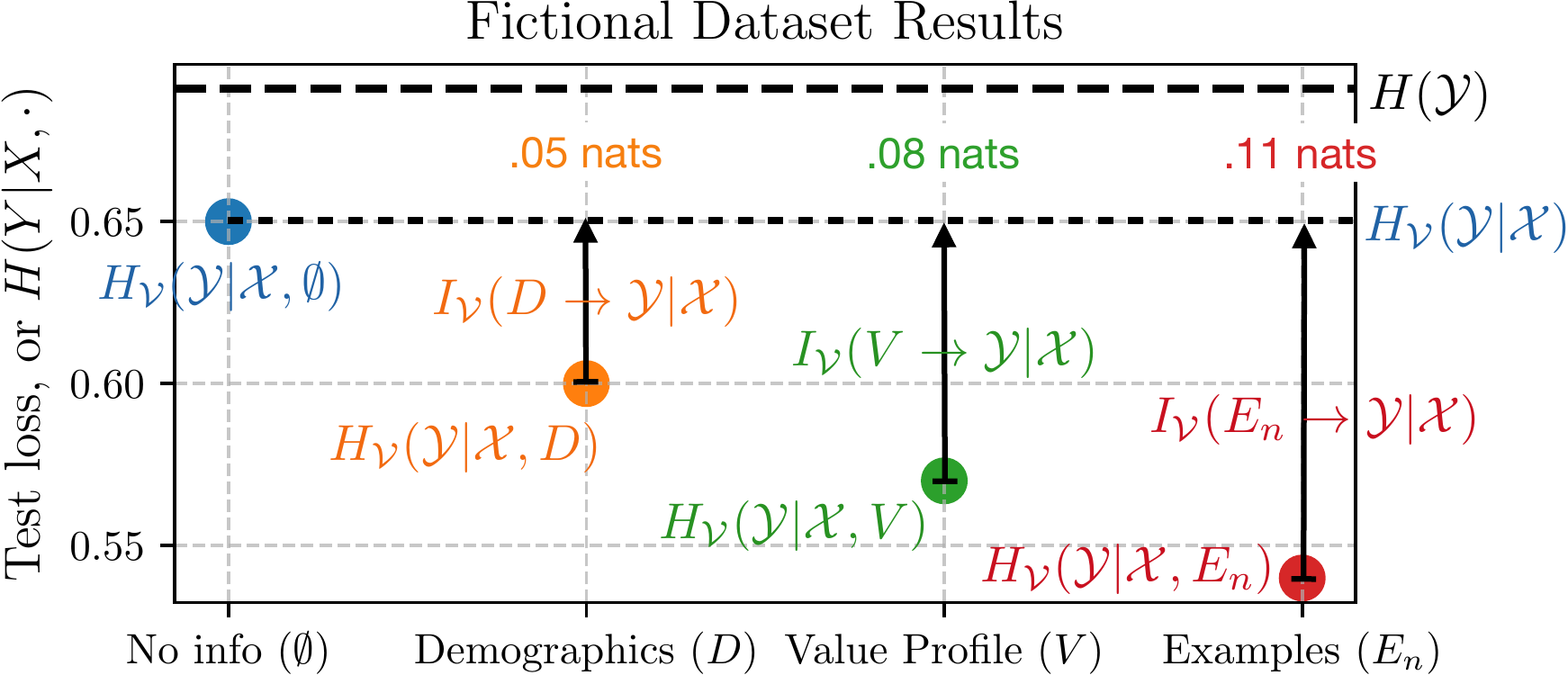}
\small
\caption{
An illustrative plot on fictional data for measuring $\mathcal{V}$-info.
% Information content is the model's decrease in test loss when given access to the information.
}
\label{fig:example}
% \vspace{-20pt}
% \end{wrapfigure}
\end{figure}

\begin{figure*}[ht]
\centering
\includegraphics[width=\textwidth]{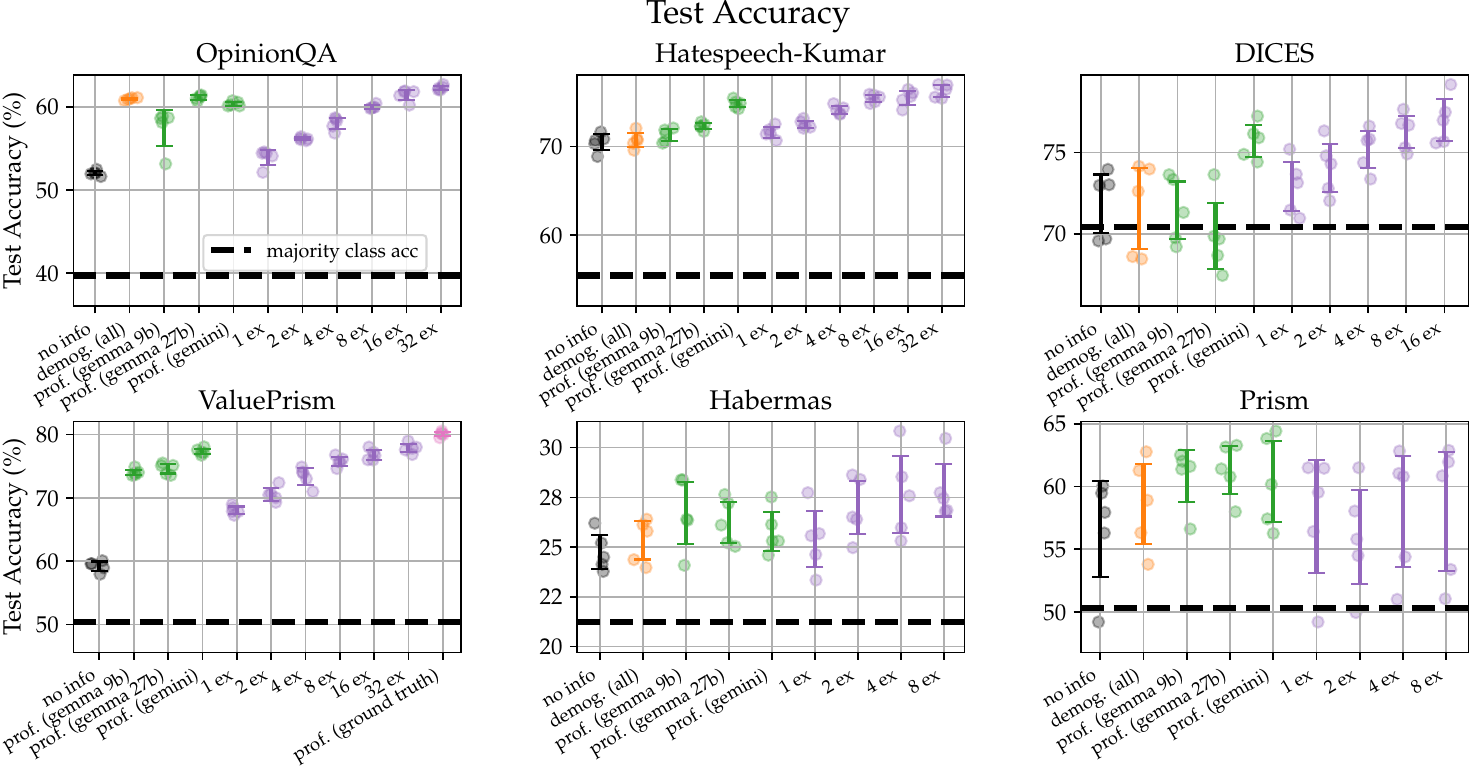}
% \caption{Each dot, here and throughout, correspond to the performance of a single train/test split, with error bars (95\% confidence) reported across random seeds. Significance levels are calculated with a mixed-linear effects model to test for significant difference in accuracies compared to the "no info" treatment (p < .05/.01/.001 for */**/***).}
\caption{Held-out accuracies across rater representations. Accuracies are reported for the same training runs as Figure \ref{fig:nllresults}, with very similar takeaways/results.} 
\label{fig:accresults}
\end{figure*}

\begin{figure*}[ht]
\centering
\includegraphics[width=0.9\textwidth]{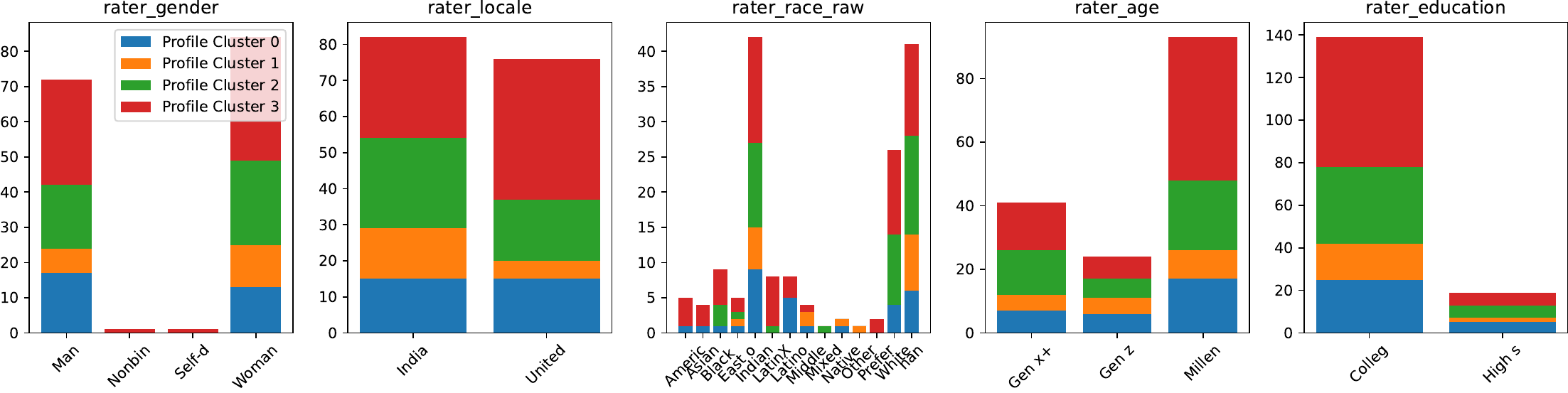}
\caption{For DICES, the four profile clusters cut across demographic groups along all dimensions.}
\label{fig:dicesdem}
\end{figure*}

\subsection{Predictive Power of Demographic Groups}

\begin{figure*}[ht]
\centering
\includegraphics[width=\textwidth]{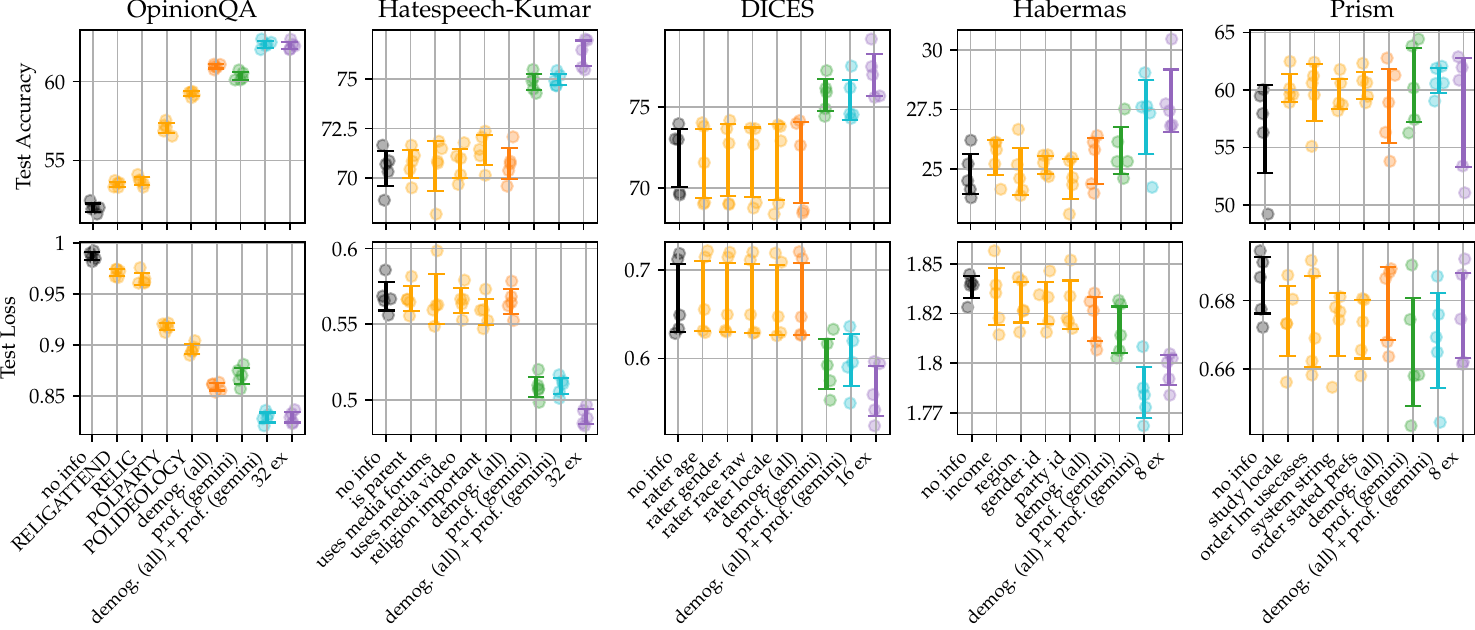}
\caption{Performance using \textcolor{singledemcol}{\bf one demographic at a time}, \textcolor{demcolor}{\bf all demographics}, \textcolor{profcolor}{\bf value profiles}, and \textcolor{bothcol}{\bf all demographics with the value profile}. \textbf{No information} and the \textcolor{excolor}{\bf max examples} settings are also reported as baselines. The four most predictive demographics (as measured by test loss) are reported for each dataset, results for the remaining demographics can be found in Appendix \ref{app:bigresults}.}
\label{fig:demographicresults}
\end{figure*}

In addition to presenting the decoder with all rater demographic variables at once (i.e., intersectional demographics), we also train a decoder for each demographic dimension individually. This allows us 1) to see the extent to which grouping individuals based on demographic dimensions is predictive, and 2) which demographic dimensions contain the most usable information for any given dataset. We also train a decoder using all demographic information plus the value profiles. See Figure \ref{fig:demographicresults} for results.
Some main findings include:

\textbf{Grouping by demographics does not add significant predictive power.} Grouping individuals based on individual demographic dimensions did not significantly improve predictive power, except for OQA, where political ideology/party and religious affiliation/attendance were most informative.

\textbf{Value profiles and demographics can be complementary.} Combining value profiles and demographics resulted in performance as good as or better than either one individually. This suggests that the decoder can leverage both types of information when relevant, e.g.\ ignore irrelevant information when it is not useful (cf.\ demographics in DIC/HK) and combine complementary information when useful (cf.\ OQA/HL).

% \subsection{Natural Language Rater Annotations}

\subsection{How does the method generalize to free-form text?}

For all experiments in the paper, rater annotations were categorical/ordinal responses to a small, finite number of options. This decision was made largely because of a lack of adequate datasets with more complex annotations. However, the question remains - how does the method generalize to free-form text outputs?

\begin{figure}[ht]
\centering
\includegraphics[width=0.4\textwidth]{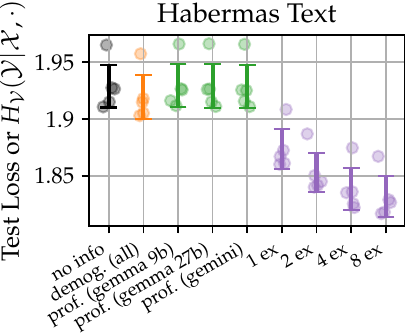}
\caption{Results when testing our method on predicting textual rater justifications.}
\label{fig:habtext}
\end{figure}

One (and only one) of our datasets, Habermas \citep{habermas}, has free-form rater outputs: the justification that people gave for why they gave the likert response that they did. These descriptions are usually a few sentences long, and contain interesting value information. To get a data point of how our method generalizes to free-form text, we also train a decoder designed to output textual justifications on this dataset (results in Figure \ref{fig:habtext}).

Similar to the categorical results, including more examples does indeed help test perplexity over the no information setting. However, demographics and profiles are not able to help significantly. We have two theories as to why this is the case. Firstly, text contains not only value information, but also stylistic and syntactic information - for example, some raters begin every justification with the same phrase, and others write short vs. long justifications. Thus, in-context examples communicate both value-relevant information and syntactic information, and it is difficult to tell which is causing the decrease in perplexity. Secondly, Habermas was our smallest dataset, making conclusions difficult to decisively draw for even the discrete likert-scale setting. Thus, it is possible that this negative result is in part due to the decoder being underfit, and that value profiles would be able to provide predictive information with additional training.
As these results are only on one (small) dataset, we believe that testing generality of the method to free-form text is a promising avenue for future work.

% \subsection{Zero-shot Decoders and Souping}
% \subsection{Generalization to }
\subsection{Zero-shot decoder performance}

\begin{figure*}[ht]
\centering
\includegraphics[width=0.9\textwidth]{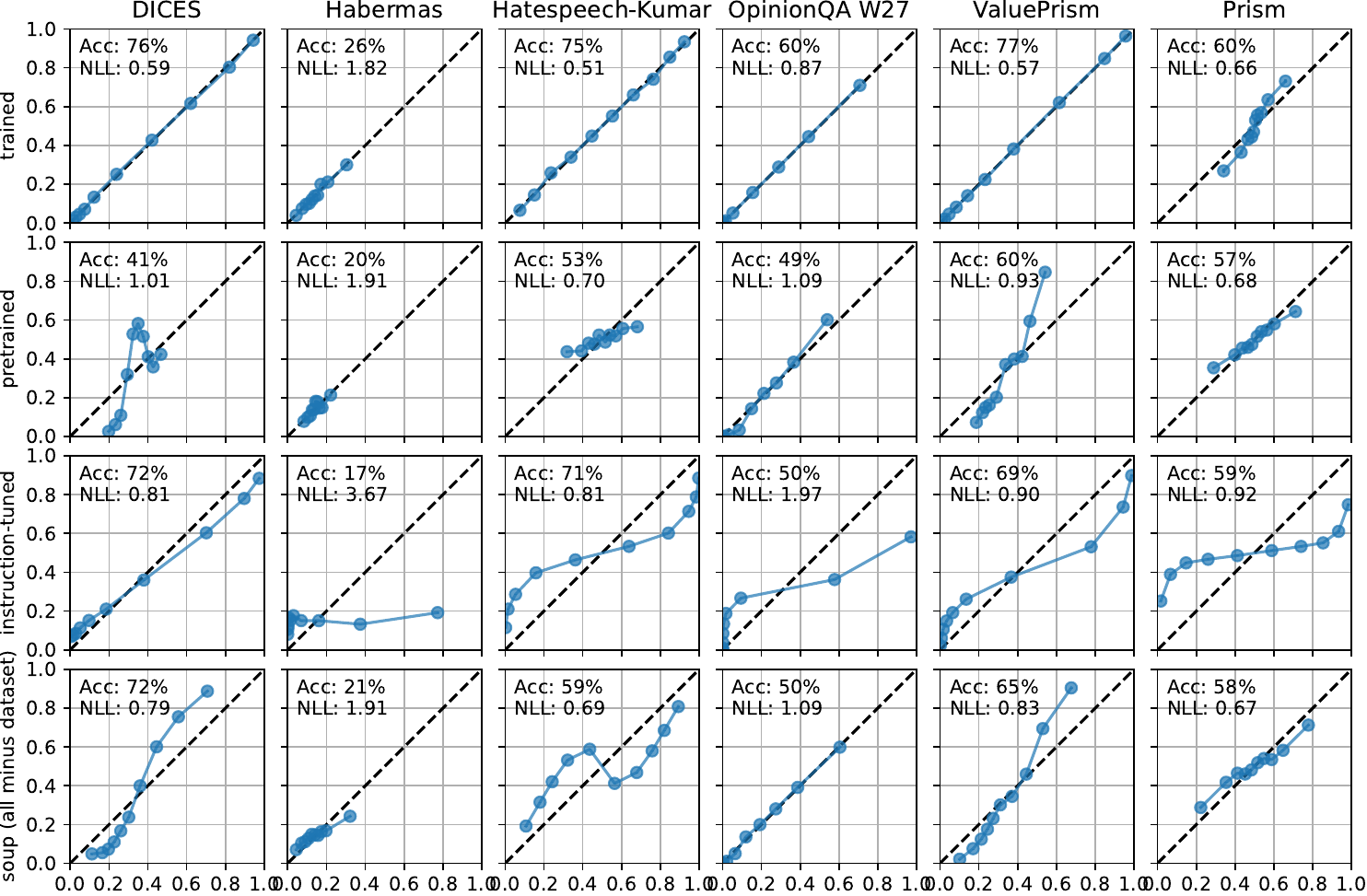}
% \caption{Calibration plot - dataset held out held out. 1. Models trained on the dataset are quite well-calibrated, 2. On 5/6 datasets, instruction-tuned get higher accuracy, but pretrained gets lower loss on 4/6 datasets (better calibration), 3. Souped zero-shot models (based on pt) get lower or same losses on all datasets, showing some ability to generalize between datasets}
\caption{Results and calibration plot for zero-shot results for pretrained/base models, instruction-tuned models, and souped models on all but the dataset to evaluate. Results are compared to the decoder trained on the dataset.}
\label{fig:bigcalibration}
\end{figure*}

For all experiments in the main paper, we train a decoder (using SFT) on a set of train raters and evaluate them on held out test raters.
While this is necessary for estimating rater information,
we are also curious to know: \textit{how well can a value profile decoder perform without dataset-specific training?}
Specifically, we evaluate on the following settings:
\begin{itemize}
\item Pretrained/base model: Prompted base model \texttt{gemma2-9b-pt}.
\item Instruction-tuned model: Prompted instruction-tuned model \texttt{gemma2-9b-it}.
\item Souped model \citep{wortsman2022modelsoupsaveragingweights}: Average the model weights from the trained decoders on all datasets \textit{except} for the evaluation dataset.
\item Trained model: For comparison, we also show results for the trained model.
\end{itemize}
Performance and calibration results are reported in Figure \ref{fig:bigcalibration}.

Some results include:
\begin{itemize}
\item As expected, the trained models both offer the best performance and calibration.
\item Instruction-tuned models generally get higher accuracy than base models (5/6 datasets), but base models generally get lower loss (4/6 datasets) due to better calibration.
\item The souped models (finetuned from pretrained model) get the same or lower loss as the base models on all datasets, showing some ability to generalize to novel datasets.
\end{itemize}

All in all, training seems important for learning calibration, and there is some demonstrated ability to generalize from one dataset to another via souping. Additional work exploring how to maintain calibration and performance on out-of-distribution dataset settings is an interesting avenue for future work.

\section{Applications and Extensions}
\label{app:applications}

Given a set of value profiles and well-calibrated, trained decoders, there are many possible exciting applications. We list a few here.

\subsection{Disentangling (Value)-Epistemic and Aleatoric Uncertainty}

In the context of modeling human variation, uncertainty can arise from two distinct sources: epistemic uncertainty (reducible through rater information) and aleatoric uncertainty (irreducible random variation).
With value profiles, we can further look at value-epistemic uncertainty, or uncertainty that can be reduced by better understanding a rater's values.

Specifically, given a set of instances, raters, and their annotations, we can measure the proportion of total uncertainty that can be attributed to value differences versus inherent randomness:

\begin{itemize}
\item \textbf{Total Uncertainty}: The entropy of ratings given just the instance, $H_V(Y|X)$
\item \textbf{Value-Epistemic Uncertainty}: The information gained by knowing value profiles, $I_V(V(R) \to Y|X) = H_V(Y|X) - H_V(Y|X,V(R))$
\item \textbf{Aleatoric Uncertainty}: The remaining uncertainty after conditioning on both instance and value profiles, $H_V(Y|X,V(R))$
\end{itemize}

The ratio $I_V(V(R) \to Y|X) / H_V(Y|X)$ represents the fraction of uncertainty that is value-epistemic (reducible by knowing values), while $H_V(Y|X,V(R)) / H_V(Y|X)$ represents the fraction that is aleatoric (irreducible even with value knowledge).

Instance-level uncertainty can similarly be measured by looking at $H_V(Y|x)$, $I_V(V(R) \to Y|x)$, and $H_V(Y|x,V(R))$. Similar definitions also exist for any other rater representation.

\begin{figure*}[ht]
\centering
\includegraphics[width=\textwidth]{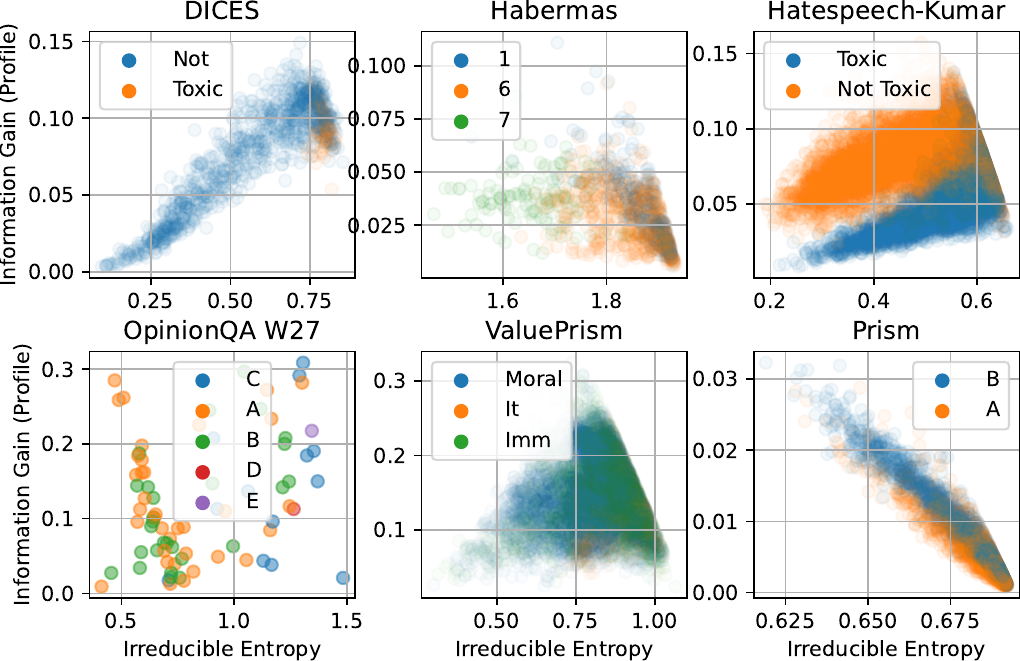}
\caption{Value-Epistemic Uncertainty (a.k.a., Information Gain from Value Profile) vs. the Irreducible Entropy (or Aleatoric Uncertainty) for each instance in each dataset, colored by label.}
\label{fig:instance uncertainty}
\end{figure*}

We plot instance-level value-epistemic vs. aleatoric uncertainty for all instances in each dataset in Figure \ref{fig:instance uncertainty}.

Such analyses and information may be useful for determining which instances have higher or lower disagreement and whether that disagreement is due to value-relevant factors or other factors.

\subsection{Identifying instance-specific value information}

Each instance may have particular values which are more or less relevant for the instance as well. Using value decoders, one can estimate the relevance of a value for an instance with $I_V(v \to Y|x)$. This could be useful in cases such as if one wants to know what values to survey raters for for a particular instance.

\subsection{Rater difficulty}

Some raters may more easily be modeled by value profiles (or profile clusters) than others.
For example, given a set of candidate value profiles (or, value profile clusters), one could measure the test loss for a rater given the optimal assignment. The lower the test loss, the more easily modeled they are by the value profile, the higher the test loss, the more they may not be easily explained by a value profile.
In this way, one could find raters that either a) are not easily modeled by a value profile in the current system or b) may be providing low-quality (or random) judgements.

\subsection{Other applications}

Other potential applications include:
\begin{itemize}
\item Designing an active learning system to select instances for a rater to annotate that are most likely to provide value-relevant information;
\item Exploring which groups are best or worst represented with value profiles;
\item Building a system to help someone explore their own values (see \S\ref{sec:ethicalconsiderations});
\end{itemize}
or more.

\section{Prompts}
\label{app:prompts}

\begin{figure}[t]
\centering
\includegraphics[width=\columnwidth]{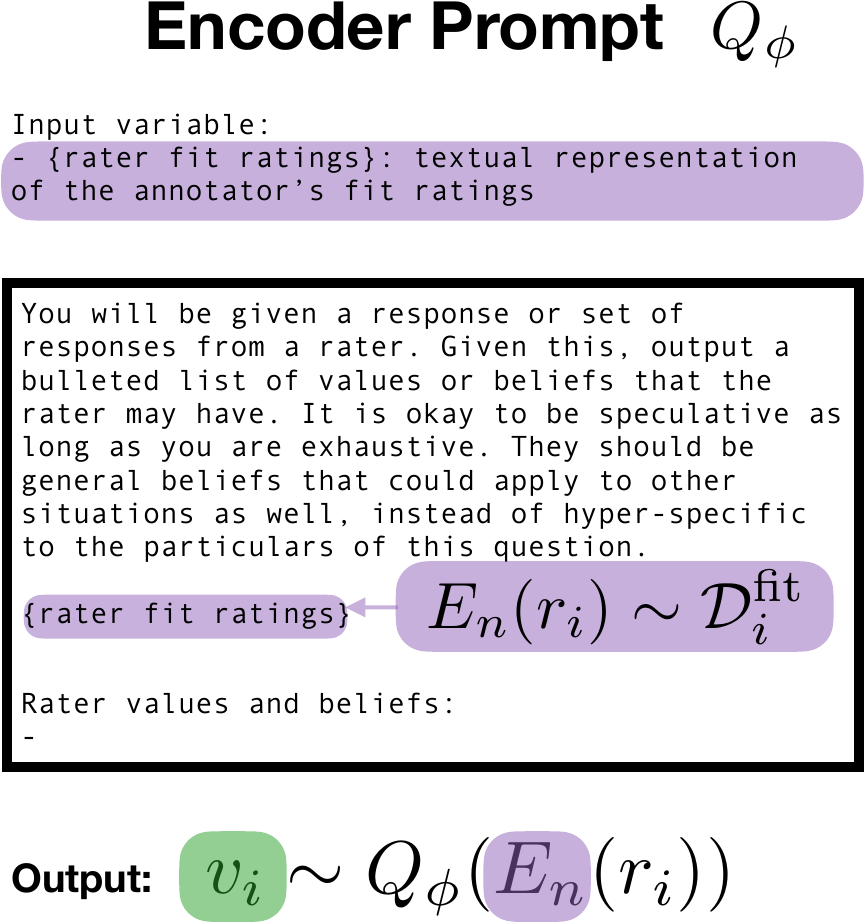}
\caption{Encoder prompt}
\label{fig:encoderprompt}
\end{figure}

\begin{figure*}[t]
\centering
\includegraphics[width=\textwidth]{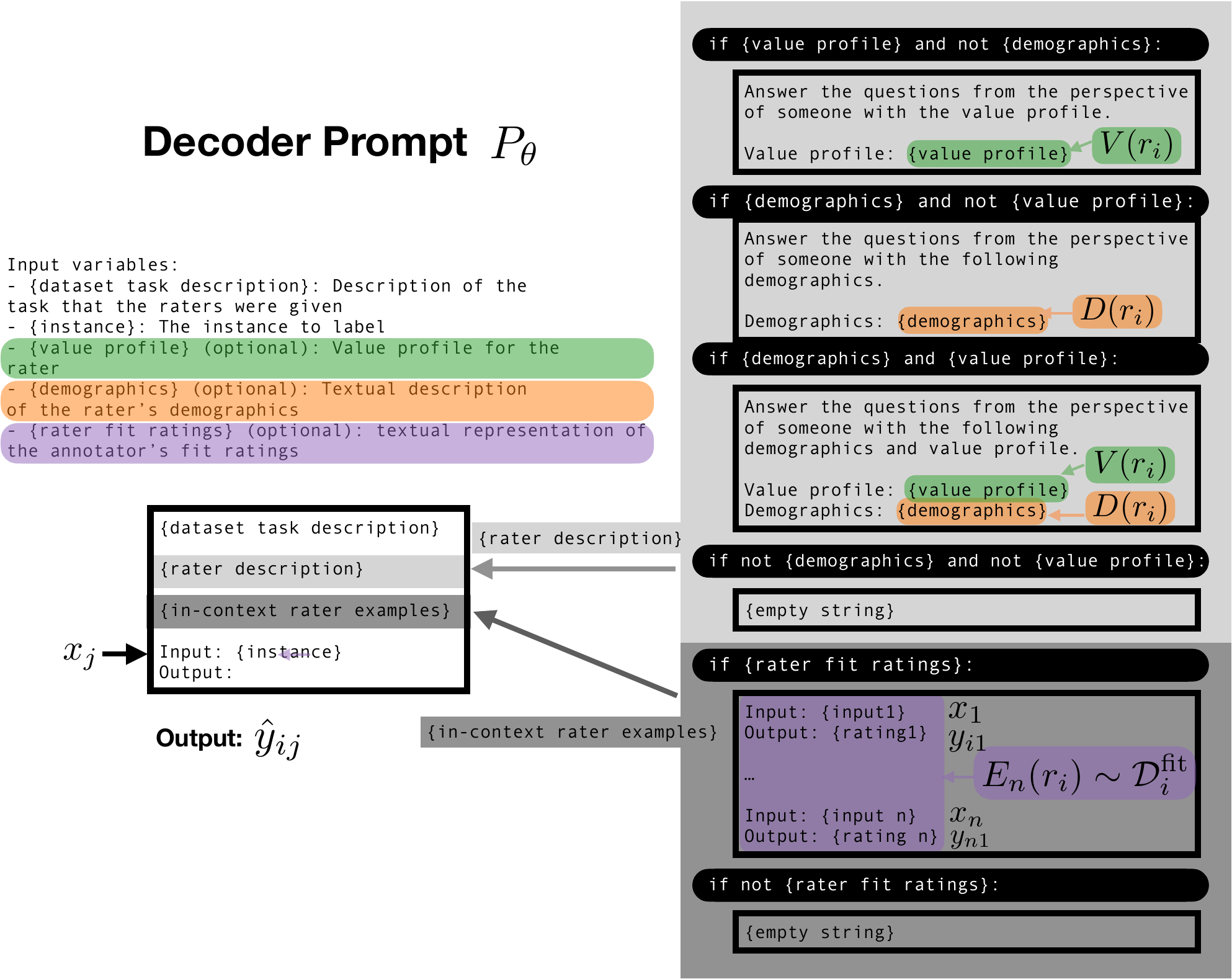}
\caption{Decoder prompt}
\label{fig:decoderprompt}
\end{figure*}

See Figure \ref{fig:encoderprompt} for the encoder prompt and Figure \ref{fig:decoderprompt} for the decoder prompt used for all experiments.

% \todo{fill in the prompts}

\begin{table*}[ht]
\centering
\small
\begin{tabular}{|p{3cm}|p{12cm}|}
\hline
Dataset & Demographics \\
\hline
OpinionQA (W27) & CREGION, SEX, EDUCATION, CITIZEN, MARITAL, INCOME, RACE, RELIG, RELIGATTEND, POLPARTY, POLIDEOLOGY \\
\hline
Habermas-Likert & party\_id, religion, age, education, ethnicity, gender\_id, immigration\_status, income, region \\
\hline
DICES & rater\_gender, rater\_locale, rater\_race\_raw, rater\_age, rater\_education \\
\hline
Hatespeech-Kumar & gender, gender\_other, race, identify\_as\_transgender, lgbtq\_status, education, age\_range, political\_affilation, is\_parent, religion\_important, technology\_impact, uses\_media\_social, uses\_media\_news, uses\_media\_video, uses\_media\_forums, personally\_seen\_toxic\_content, personally\_been\_target, toxic\_comments\_problem \\
\hline
ValuePrism & None, but has "ground truth" value profiles of the original value / right / duty. \\
\hline
Prism & lm\_familiarity, lm\_indirect\_use, lm\_direct\_use, lm\_frequency\_use, lm\_usecases, self\_description, system\_string, religion, stated\_prefs, order\_lm\_usecases, order\_stated\_prefs, age, gender, employment\_status, education, marital\_status, english\_proficiency, study\_locale, location, ethnicity \\
\hline
\end{tabular}
\caption{Dataset demographic variables}
\label{tab:demographics}
\end{table*}

\section{Data}
\label{app:data}
\subsection{Dataset Preprocessing Details}
\label{app:datasetpreprocessing}

We carried out the following preprocessing steps for the datasets -
DIC: used the larger subset (990);
HL: selected raters with at least nine responses;
HK: randomly selected 5k raters and binarized annotations;
OQA: randomly selected a wave for experiments (Wave 27);
PR: select annotations from first conversation turn and compared the chosen response to the next highest rated response;
VP: Treat each value, right, or duty as a unique annotator.
Finally, for all datasets we filtered to annotators that had at least four responses.

\subsection{All Dataset Demographic Variables}
Refer to Table \ref{tab:demographics} to see all demographic variables contained in each dataset.

\newpage
\section{Detailed Results}
\label{app:bigresults}

The full results for each dataset can be found in:
\begin{enumerate}
\item OpinionQA: Table \ref{tab:performance_opinionqa}
\item Hatespeech-Kumar: Table \ref{tab:performance_hatespeech-kumar}
\item DICES: Table \ref{tab:performance_dices}
\item ValuePrism: Table \ref{tab:performance_valueprism}
\item Habermas-Likert: Table \ref{tab:performance_habermas}
\item Prism: Table \ref{tab:performance_prism}
\end{enumerate}

\begin{table*}[h]
\centering
\small
\begin{tabular}{|l|c|c|c|r|}
\hline
Name & Test Accuracy & Test Loss & Usable Info (nats) & Info Preserved \\
\hline
no info & 52.0 (±0.14) & 0.987 (±0.002) & 0.000 & (0\%) \\
dem CITIZEN & 51.9 (±0.08) & 0.987 (±0.002) & 0.000 & - \\
dem CREGION & 51.9 (±0.12) & 0.987 (±0.002) & 0.000 & - \\
dem EDUCATION & 52.1 (±0.10) & 0.985 (±0.002) & 0.002 & - \\
dem INCOME & 52.0 (±0.13) & 0.985 (±0.002) & 0.002 & - \\
dem MARITAL & 52.4 (±0.07) & 0.983 (±0.002) & 0.004 & - \\
dem POLIDEOLOGY & 59.2 (±0.07) & 0.896 (±0.002) & 0.091 & - \\
dem POLPARTY & 57.1 (±0.17) & 0.918 (±0.002) & 0.069 & - \\
dem RACE & 52.5 (±0.15) & 0.983 (±0.002) & 0.004 & - \\
dem RELIG & 53.7 (±0.13) & 0.965 (±0.003) & 0.022 & - \\
dem RELIGATTEND & 53.5 (±0.08) & 0.971 (±0.002) & 0.016 & - \\
dem SEX & 52.1 (±0.11) & 0.985 (±0.001) & 0.002 & - \\
dem identity columns & 53.3 (±0.32) & 0.972 (±0.004) & 0.015 & - \\
dem value columns & 60.1 (±0.41) & 0.881 (±0.008) & 0.106 & - \\
dem (all) & 61.0 (±0.07) & 0.859 (±0.002) & 0.128 & - \\
profile cluster-2 & 59.3 (±0.15) & 0.899 (±0.002) & 0.088 & 56\% \\
profile cluster-4 & 60.1 (±0.22) & 0.878 (±0.003) & 0.109 & 69\% \\
profile cluster-8 & 60.8 (±0.18) & 0.866 (±0.003) & 0.121 & 77\% \\
profile 9b & 57.5 (±1.10) & 0.918 (±0.016) & 0.069 & 43\% \\
profile 27b & 61.1 (±0.14) & 0.866 (±0.002) & 0.120 & 76\% \\
profile gni & 60.3 (±0.13) & 0.870 (±0.004) & 0.117 & 74\% \\
dem+profile gni & 62.4 (±0.11) & 0.829 (±0.002) & 0.158 & - \\
1 ex & 53.9 (±0.46) & 0.964 (±0.005) & 0.023 & - \\
2 ex & 56.1 (±0.08) & 0.937 (±0.002) & 0.050 & - \\
4 ex & 58.0 (±0.34) & 0.906 (±0.005) & 0.081 & - \\
8 ex & 60.0 (±0.11) & 0.870 (±0.003) & 0.117 & - \\
16 ex & 61.4 (±0.32) & 0.843 (±0.005) & 0.143 & - \\
32 ex & 62.3 (±0.12) & 0.829 (±0.003) & 0.158 & (100\%) \\
majority class acc./dataset entropy & 39.7 (±0.00) & 1.290 (±0.000) & - & - \\
\hline
\end{tabular}
\caption{OpinionQA Performance Metrics (Model: gemma2-9b-pt) Other datasets: Appendix: \ref{app:bigresults}}
\label{tab:performance_opinionqa}
\end{table*}

\begin{table*}[h]
\centering
\small
\begin{tabular}{|l|c|c|c|r|}
\hline
Name & Test Accuracy & Test Loss & Usable Info (nats) & Info Preserved \\
\hline
no info & 70.5 (±0.46) & 0.569 (±0.005) & 0.000 & (0\%) \\
dem (all) & 70.7 (±0.40) & 0.565 (±0.004) & 0.003 & - \\
dem personally been target & 70.6 (±0.36) & 0.568 (±0.004) & 0.000 & - \\
dem personally seen toxic content & 70.6 (±0.40) & 0.567 (±0.004) & 0.001 & - \\
dem age range & 70.6 (±0.42) & 0.568 (±0.004) & 0.001 & - \\
dem uses media social & 70.6 (±0.38) & 0.568 (±0.004) & 0.001 & - \\
dem uses media news & 70.7 (±0.35) & 0.568 (±0.004) & 0.001 & - \\
dem uses media forums & 70.6 (±0.65) & 0.566 (±0.008) & 0.002 & - \\
dem toxic comments problem & 70.7 (±0.38) & 0.567 (±0.004) & 0.001 & - \\
dem technology impact & 70.6 (±0.38) & 0.569 (±0.004) & 0.000 & - \\
dem religion important & 71.4 (±0.39) & 0.558 (±0.004) & 0.010 & - \\
dem race & 70.7 (±0.37) & 0.568 (±0.004) & 0.001 & - \\
dem political affilation & 70.7 (±0.41) & 0.568 (±0.004) & 0.001 & - \\
dem lgbtq status & 70.6 (±0.36) & 0.569 (±0.004) & 0.000 & - \\
dem is parent & 70.7 (±0.36) & 0.567 (±0.004) & 0.002 & - \\
dem identity columns & 70.6 (±0.32) & 0.568 (±0.004) & 0.000 & - \\
dem identify as transgender & 70.5 (±0.35) & 0.568 (±0.004) & 0.000 & - \\
dem gender other & 70.6 (±0.40) & 0.568 (±0.004) & 0.000 & - \\
dem gender & 70.6 (±0.34) & 0.568 (±0.005) & 0.000 & - \\
dem education & 70.5 (±0.38) & 0.568 (±0.004) & 0.001 & - \\
dem uses media video & 70.7 (±0.37) & 0.566 (±0.004) & 0.003 & - \\
dem value columns & 71.0 (±0.35) & 0.563 (±0.004) & 0.005 & - \\
profile cluster-2 & 70.1 (±0.52) & 0.572 (±0.006) & -0.004 & -5\% \\
profile cluster-4 & 72.6 (±0.23) & 0.539 (±0.003) & 0.029 & 37\% \\
profile cluster-8 & 73.1 (±0.26) & 0.532 (±0.003) & 0.036 & 46\% \\
profile 9b & 71.3 (±0.35) & 0.554 (±0.003) & 0.014 & 18\% \\
profile 27b & 72.3 (±0.18) & 0.543 (±0.002) & 0.026 & 33\% \\
profile gni & 74.8 (±0.21) & 0.509 (±0.003) & 0.060 & 76\% \\
dem+profile gni & 75.0 (±0.15) & 0.509 (±0.003) & 0.059 & - \\
1 ex & 71.6 (±0.31) & 0.553 (±0.003) & 0.016 & - \\
2 ex & 72.5 (±0.21) & 0.541 (±0.002) & 0.028 & - \\
4 ex & 74.1 (±0.22) & 0.521 (±0.002) & 0.047 & - \\
8 ex & 75.5 (±0.21) & 0.500 (±0.001) & 0.069 & - \\
16 ex & 75.5 (±0.40) & 0.500 (±0.006) & 0.069 & - \\
32 ex & 76.3 (±0.33) & 0.489 (±0.003) & 0.079 & (100\%) \\
majority class acc./dataset entropy & 55.4 (±0.00) & 0.687 (±0.000) & - & - \\
\hline
\end{tabular}
\caption{Hatespeech-Kumar Performance Metrics (Model: gemma2-9b-pt) Other datasets: Appendix: \ref{app:bigresults}}
\label{tab:performance_hatespeech-kumar}
\end{table*}

\begin{table*}[h]
\centering
\small
\begin{tabular}{|l|c|c|c|r|}
\hline
Name & Test Accuracy & Test Loss & Usable Info (nats) & Info Preserved \\
\hline
no info & 71.8 (±0.92) & 0.668 (±0.020) & 0.000 & (0\%) \\
dem rater age & 71.5 (±1.09) & 0.671 (±0.020) & -0.002 & - \\
dem rater education & 71.1 (±1.21) & 0.673 (±0.020) & -0.004 & - \\
dem rater gender & 71.7 (±1.13) & 0.669 (±0.020) & -0.001 & - \\
dem rater locale & 71.6 (±1.20) & 0.666 (±0.020) & 0.002 & - \\
dem rater race raw & 71.6 (±1.10) & 0.668 (±0.020) & 0.000 & - \\
dem (all) & 71.6 (±1.27) & 0.667 (±0.021) & 0.002 & - \\
profile cluster-2 & 75.6 (±0.51) & 0.605 (±0.012) & 0.064 & 60\% \\
profile cluster-4 & 76.4 (±0.67) & 0.576 (±0.015) & 0.093 & 88\% \\
profile cluster-8 & 76.9 (±0.69) & 0.570 (±0.013) & 0.098 & 93\% \\
profile 9b & 71.5 (±0.90) & 0.683 (±0.023) & -0.015 & -14\% \\
profile 27b & 69.9 (±1.04) & 0.706 (±0.018) & -0.038 & -36\% \\
profile gni & 75.7 (±0.50) & 0.594 (±0.014) & 0.074 & 71\% \\
dem+profile gni & 75.4 (±0.62) & 0.598 (±0.015) & 0.070 & - \\
1 ex & 72.9 (±0.77) & 0.644 (±0.016) & 0.025 & - \\
2 ex & 74.1 (±0.76) & 0.625 (±0.014) & 0.044 & - \\
4 ex & 75.2 (±0.58) & 0.602 (±0.012) & 0.066 & - \\
8 ex & 76.3 (±0.51) & 0.580 (±0.013) & 0.089 & - \\
16 ex & 77.0 (±0.66) & 0.563 (±0.014) & 0.105 & (100\%) \\
majority class acc./dataset entropy & 70.4 (±0.00) & 0.742 (±0.000) & - & - \\
\hline
\end{tabular}
\caption{DICES Performance Metrics (Model: gemma2-9b-pt) Other datasets: Appendix: \ref{app:bigresults}}
\label{tab:performance_dices}
\end{table*}

\begin{table*}[h]
\centering
\small
\begin{tabular}{|l|c|c|c|r|}
\hline
Name & Test Accuracy & Test Loss & Usable Info (nats) & Info Preserved \\
\hline
no info & 59.2 (±0.37) & 0.852 (±0.005) & 0.000 & (0\%) \\
profile cluster-2 & 59.4 (±0.49) & 0.853 (±0.005) & -0.001 & -0\% \\
profile cluster-4 & 65.4 (±0.82) & 0.792 (±0.013) & 0.060 & 20\% \\
profile cluster-8 & 65.8 (±1.37) & 0.780 (±0.017) & 0.071 & 23\% \\
profile 9b & 74.0 (±0.24) & 0.632 (±0.006) & 0.220 & 72\% \\
profile 27b & 74.6 (±0.39) & 0.615 (±0.008) & 0.237 & 78\% \\
profile gni & 77.3 (±0.22) & 0.566 (±0.006) & 0.286 & 94\% \\
1 ex & 68.1 (±0.29) & 0.738 (±0.006) & 0.114 & - \\
2 ex & 70.6 (±0.51) & 0.695 (±0.010) & 0.157 & - \\
4 ex & 73.4 (±0.67) & 0.640 (±0.015) & 0.212 & - \\
8 ex & 75.8 (±0.35) & 0.591 (±0.007) & 0.261 & - \\
16 ex & 76.8 (±0.38) & 0.570 (±0.008) & 0.282 & - \\
32 ex & 77.9 (±0.35) & 0.547 (±0.007) & 0.305 & (100\%) \\
ground truth prof & 80.1 (±0.17) & 0.493 (±0.006) & 0.358 & - \\
ground truth prof+profile gni & 80.3 (±0.27) & 0.491 (±0.005) & 0.361 & - \\
majority class acc./dataset entropy & 50.4 (±0.00) & 1.004 (±0.000) & - & - \\
\hline
\end{tabular}
\caption{ValuePrism Performance Metrics (Model: gemma2-9b-pt) Other datasets: Appendix: \ref{app:bigresults}}
\label{tab:performance_valueprism}
\end{table*}

\begin{table*}[h]
\centering
\small
\begin{tabular}{|l|c|c|c|r|}
\hline
Name & Test Accuracy & Test Loss & Usable Info (nats) & Info Preserved \\
\hline
no info & 24.8 (±0.43) & 1.838 (±0.003) & 0.000 & (0\%) \\
dem demographics.age & 23.9 (±1.21) & 1.846 (±0.015) & -0.007 & - \\
dem demographics.education & 24.1 (±0.69) & 1.847 (±0.004) & -0.008 & - \\
dem demographics.ethnicity & 24.1 (±0.45) & 1.834 (±0.005) & 0.004 & - \\
dem demographics.gender id & 25.2 (±0.19) & 1.830 (±0.005) & 0.008 & - \\
dem demographics.immigration status & 23.5 (±1.13) & 1.838 (±0.009) & 0.000 & - \\
dem demographics.income & 25.5 (±0.37) & 1.834 (±0.007) & 0.005 & - \\
dem demographics.party id & 24.6 (±0.43) & 1.829 (±0.006) & 0.009 & - \\
dem demographics.region & 24.9 (±0.50) & 1.831 (±0.005) & 0.008 & - \\
dem demographics.religion & 24.7 (±0.67) & 1.843 (±0.005) & -0.004 & - \\
dem identity columns & 23.9 (±1.27) & 1.852 (±0.012) & -0.013 & - \\
dem value columns & 23.6 (±1.32) & 1.835 (±0.020) & 0.003 & - \\
dem (all) & 25.3 (±0.49) & 1.822 (±0.006) & 0.016 & - \\
profile cluster-2 & 24.3 (±0.54) & 1.840 (±0.004) & -0.002 & -4\% \\
profile cluster-4 & 24.6 (±0.56) & 1.846 (±0.010) & -0.008 & -19\% \\
profile cluster-8 & 24.6 (±0.45) & 1.844 (±0.005) & -0.006 & -14\% \\
profile 9b & 26.7 (±0.80) & 1.819 (±0.004) & 0.019 & 46\% \\
profile 27b & 26.2 (±0.52) & 1.815 (±0.003) & 0.023 & 56\% \\
profile gni & 25.8 (±0.50) & 1.817 (±0.006) & 0.022 & 52\% \\
dem+profile gni & 27.2 (±0.80) & 1.785 (±0.007) & 0.053 & - \\
1 ex & 25.4 (±0.72) & 1.814 (±0.004) & 0.025 & - \\
2 ex & 27.0 (±0.68) & 1.802 (±0.005) & 0.036 & - \\
4 ex & 27.6 (±0.98) & 1.799 (±0.004) & 0.039 & - \\
8 ex & 27.9 (±0.68) & 1.797 (±0.004) & 0.042 & (100\%) \\
majority class acc./dataset entropy & 21.2 (±0.00) & 1.906 (±0.000) & - & - \\
\hline
\end{tabular}
\caption{Habermas Performance Metrics (Model: gemma2-9b-pt) Other datasets: Appendix: \ref{app:bigresults}}
\label{tab:performance_habermas}
\end{table*}

\begin{table*}[h]
\centering
\small
\begin{tabular}{|l|c|c|c|r|}
\hline
Name & Test Accuracy & Test Loss & Usable Info (nats) & Info Preserved \\
\hline
no info & 56.6 (±1.96) & 0.684 (±0.004) & 0.000 & (0\%) \\
dem age & 58.9 (±0.92) & 0.681 (±0.006) & 0.004 & - \\
dem study locale & 60.1 (±0.61) & 0.674 (±0.005) & 0.010 & - \\
dem stated prefs & 58.4 (±0.93) & 0.680 (±0.007) & 0.004 & - \\
dem self description & 58.9 (±1.60) & 0.678 (±0.004) & 0.006 & - \\
dem religion & 55.8 (±1.41) & 0.686 (±0.002) & -0.001 & - \\
dem order stated prefs & 60.4 (±0.60) & 0.672 (±0.004) & 0.013 & - \\
dem order lm usecases & 59.8 (±1.26) & 0.674 (±0.007) & 0.011 & - \\
dem marital status & 59.3 (±1.01) & 0.676 (±0.006) & 0.009 & - \\
dem location & 58.0 (±1.87) & 0.676 (±0.007) & 0.009 & - \\
dem lm usecases & 59.3 (±1.27) & 0.675 (±0.006) & 0.009 & - \\
dem lm indirect use & 55.4 (±1.62) & 0.833 (±0.145) & -0.149 & - \\
dem lm frequency use & 59.3 (±0.88) & 0.680 (±0.006) & 0.005 & - \\
dem lm familiarity & 55.5 (±2.45) & 0.685 (±0.003) & -0.001 & - \\
dem lm direct use & 56.5 (±1.60) & 0.683 (±0.004) & 0.002 & - \\
dem identity columns & 60.8 (±0.58) & 0.671 (±0.004) & 0.013 & - \\
dem gender & 57.1 (±1.88) & 0.682 (±0.006) & 0.002 & - \\
dem ethnicity & 57.8 (±1.40) & 0.684 (±0.004) & 0.000 & - \\
dem english proficiency & 57.5 (±1.37) & 0.684 (±0.004) & 0.000 & - \\
dem employment status & 59.6 (±0.75) & 0.677 (±0.005) & 0.008 & - \\
dem education & 59.1 (±0.91) & 0.679 (±0.005) & 0.005 & - \\
dem system string & 59.6 (±0.67) & 0.673 (±0.005) & 0.011 & - \\
dem value columns & 58.6 (±1.83) & 0.676 (±0.005) & 0.008 & - \\
dem (all) & 58.6 (±1.63) & 0.679 (±0.006) & 0.005 & - \\
profile cluster-2 & 60.2 (±0.58) & 0.673 (±0.005) & 0.012 & 131\% \\
profile cluster-4 & 58.2 (±2.13) & 0.674 (±0.006) & 0.010 & 114\% \\
profile cluster-8 & 56.2 (±2.09) & 0.684 (±0.005) & 0.000 & 5\% \\
profile 9b & 60.8 (±1.07) & 0.672 (±0.006) & 0.013 & 145\% \\
profile 27b & 61.3 (±0.96) & 0.667 (±0.008) & 0.017 & 191\% \\
profile gni & 60.4 (±1.64) & 0.665 (±0.008) & 0.020 & 220\% \\
dem+profile gni & 60.8 (±0.55) & 0.668 (±0.007) & 0.016 & - \\
1 ex & 57.6 (±2.30) & 0.677 (±0.005) & 0.007 & - \\
2 ex & 56.0 (±1.91) & 0.681 (±0.006) & 0.003 & - \\
4 ex & 58.0 (±2.26) & 0.676 (±0.006) & 0.009 & - \\
8 ex & 58.0 (±2.42) & 0.676 (±0.006) & 0.009 & (100\%) \\
majority class acc./dataset entropy & 50.3 (±0.00) & 0.693 (±0.000) & - & - \\
\hline
\end{tabular}
\caption{Prism Performance Metrics (Model: gemma2-9b-pt) Other datasets: Appendix: \ref{app:bigresults}}
\label{tab:performance_prism}
\end{table*}

\newpage
\section{Profile Clusters}

\label{app:profileclusters}

\subsection{DICES}

\subsubsection{2 clusters (DICES)}

\textbf{Cluster Profile 1:} \profiletext{High tolerance for offensive language and behavior; Focus on intent rather than impact; Narrow definition of toxicity; Prioritization of conversation flow over emotional safety; Belief in personal responsibility for emotional reactions; Discomfort with addressing conflict or negativity; Lack of awareness of subtle forms of toxicity; Potential bias towards certain topics or demographics; General optimism or positivity}

\textbf{Cluster Profile 2:} \profiletext{Strong reaction to overt negativity or hate speech; Sensitivity to potential harm, even if unintentional; Prioritizes intention over impact; Tolerance for bluntness and disagreement; Focus on explicit language rather than subtle cues; Possible political bias; Inconsistency in application of criteria}

\subsubsection{4 clusters (DICES)}

\textbf{Cluster Profile 1:} \profiletext{Strong reaction to overt negativity or hate speech; Sensitivity to potential harm, even if unintentional; Prioritizes intention over impact; Tolerance for bluntness and disagreement; Focus on explicit language rather than subtle cues; Possible political bias; Inconsistency in application of criteria}

\textbf{Cluster Profile 2:} \profiletext{High tolerance for offensive language and controversial topics; Focus on intent over impact; Emphasis on personal responsibility and resilience; Broad definition of acceptable discourse; Potential desensitization to online interactions; Possible lack of understanding of subtle forms of toxicity; Potential bias towards certain viewpoints; Belief in the importance of context}

\textbf{Cluster Profile 3:} \profiletext{Direct insults and aggressive language are toxic; Simple greetings and neutral topics are not toxic; Humor and playful banter are not necessarily toxic; The presence of controversial opinions does not automatically equal toxicity; Context and intent matter, but are difficult to fully assess; Statements suggesting or promoting illegal or harmful activities are toxic; The rater may have a relatively high tolerance for differing viewpoints}

\textbf{Cluster Profile 4:} \profiletext{High tolerance for offensive language and behavior; Focus on intent rather than impact; Narrow definition of toxicity; Prioritization of conversation flow over emotional safety; Belief in personal responsibility for emotional reactions; Discomfort with addressing conflict or negativity; Lack of awareness of subtle forms of toxicity; Potential bias towards certain topics or demographics; General optimism or positivity}

\subsubsection{8 clusters (DICES)}

\textbf{Cluster Profile 1:} \profiletext{Strong reaction to overt negativity or hate speech; Sensitivity to potential harm, even if unintentional; Prioritizes intention over impact; Tolerance for bluntness and disagreement; Focus on explicit language rather than subtle cues; Possible political bias; Inconsistency in application of criteria}

\textbf{Cluster Profile 2:} \profiletext{Direct insults and aggressive language are toxic; Simple greetings and neutral topics are not toxic; Humor and playful banter are not necessarily toxic; The presence of controversial opinions does not automatically equal toxicity; Context and intent matter, but are difficult to fully assess; Statements suggesting or promoting illegal or harmful activities are toxic; The rater may have a relatively high tolerance for differing viewpoints}

\textbf{Cluster Profile 3:} \profiletext{Strong reaction to discussions of self-harm and suicide; Sensitivity to discussions about race and sexual orientation; Discomfort with overtly sexual conversations or innuendo; Low tolerance for aggressive or rude language; A broad definition of "toxic"; Uncertainty around certain topics; A belief that context matters; Prioritizes safety and well-being}

\textbf{Cluster Profile 4:} \profiletext{High tolerance for offensive language and behavior; Focus on intent rather than impact; Narrow definition of toxicity; Prioritization of conversation flow over emotional safety; Belief in personal responsibility for emotional reactions; Discomfort with addressing conflict or negativity; Lack of awareness of subtle forms of toxicity; Potential bias towards certain topics or demographics; General optimism or positivity}

\textbf{Cluster Profile 5:} \profiletext{Emphasis on intent over outcome; High tolerance for disagreement and differing opinions; Forgiveness for misunderstandings and apologies; Political neutrality or apathy; Discomfort with discussions about illegal activities; Leniency towards casual conversation and humor; Inconsistency in applying standards; Focus on last turn in the conversation}

\textbf{Cluster Profile 6:} \profiletext{High tolerance for controversial topics and strong opinions; Emphasis on intention over impact; Belief in freedom of expression; Acceptance of dark humor and sarcasm; Forgiveness for immaturity or ignorance; Discomfort with discussions directly involving their personal advice on difficult topics; May not be detecting subtle forms of toxicity; Possibly prioritizing engagement and entertainment over safety and inclusivity}

\textbf{Cluster Profile 7:} \profiletext{Discomfort with sexual topics and exploitation; Sensitivity to personal attacks and insults; Low tolerance for manipulative or misleading behavior; Dislike of aggressive or confrontational language; High tolerance for sarcasm and playful banter; Belief that repetitive or nonsensical conversations are not necessarily toxic; Uncertainty about the line between persistent questioning and harassment; Possible leniency towards conversations that are simply awkward or uncomfortable; Emphasis on intent and context; Potential bias toward focusing on the last statement}

\textbf{Cluster Profile 8:} \profiletext{High tolerance for offensive language and controversial topics; Focus on intent over impact; Emphasis on personal responsibility and resilience; Broad definition of acceptable discourse; Potential desensitization to online interactions; Possible lack of understanding of subtle forms of toxicity; Potential bias towards certain viewpoints; Belief in the importance of context}

\subsection{Habermas-Likert}

\subsubsection{2 clusters (Habermas-Likert)}

\textbf{Cluster Profile 1:} \profiletext{Values religious freedom and parental rights; Prioritizes family autonomy over state control; May be religious themselves; Pragmatic or uncertain about online medicine; Weighing competing values; Lack of knowledge; Belief in a mixed approach; Values personal responsibility; Prioritizes affordability and access to healthcare; Trusts market forces to some extent}

\textbf{Cluster Profile 2:} \profiletext{Strong disapproval of Theresa May; Public health consciousness; Environmental concern; Belief in direct democracy; Concern about overpopulation; Openness to government intervention; Possible leaning towards left-leaning or liberal politics; Pragmatism and nuanced views; UK-centric perspective}

\subsubsection{4 clusters (Habermas-Likert)}

\textbf{Cluster Profile 1:} \profiletext{Pro-worker; Value of leisure and rest; Concern for elderly well-being; Potential distrust of government or employers; Belief in social safety nets; Focus on quality of life over economic growth; Generational fairness; Compassion and empathy for those less fortunate; May hold specific political or ideological views}

\textbf{Cluster Profile 2:} \profiletext{Strong disapproval of Theresa May; Public health consciousness; Environmental concern; Belief in direct democracy; Concern about overpopulation; Openness to government intervention; Possible leaning towards left-leaning or liberal politics; Pragmatism and nuanced views; UK-centric perspective}

\textbf{Cluster Profile 3:} \profiletext{Altruism and global citizenship; Environmental concern; Collectivism and public health prioritization; Social welfare and belief in social safety nets; Potential for utilitarianism; Nuance and pragmatism; Possible support for animal welfare, but with caveats; Acceptance of minor moral flexibility; It is important to remember that these are just inferences based on a limited set of responses. The rater's true beliefs and values may be more complex and nuanced than what can be determined from this data alone.}

\textbf{Cluster Profile 4:} \profiletext{Values religious freedom and parental rights; Prioritizes family autonomy over state control; May be religious themselves; Pragmatic or uncertain about online medicine; Weighing competing values; Lack of knowledge; Belief in a mixed approach; Values personal responsibility; Prioritizes affordability and access to healthcare; Trusts market forces to some extent}

\subsubsection{8 clusters (Habermas-Likert)}

\textbf{Cluster Profile 1:} \profiletext{Slightly prefers free market principles; Concerned about affordability and access; Cautious about government overreach; Open to social responsibility and regulation where appropriate; Values personal autonomy; Pragmatic and moderate; Indecisive or uninformed on some topics; Potentially influenced by personal experience; Open to persuasion}

\textbf{Cluster Profile 2:} \profiletext{Pro-worker; Value of leisure and rest; Concern for elderly well-being; Potential distrust of government or employers; Belief in social safety nets; Focus on quality of life over economic growth; Generational fairness; Compassion and empathy for those less fortunate; May hold specific political or ideological views}

\textbf{Cluster Profile 3:} \profiletext{Altruism and global citizenship; Environmental concern; Collectivism and public health prioritization; Social welfare and belief in social safety nets; Potential for utilitarianism; Nuance and pragmatism; Possible support for animal welfare, but with caveats; Acceptance of minor moral flexibility; It is important to remember that these are just inferences based on a limited set of responses. The rater's true beliefs and values may be more complex and nuanced than what can be determined from this data alone.}

\textbf{Cluster Profile 4:} \profiletext{Supports government intervention in the economy; Progressive social views; Prioritizes social welfare; Believes in public infrastructure investment; Values education; Potentially skeptical of inherited power/privilege; May believe in reducing inequality; Possibly environmentally conscious; Optimistic about government's ability to improve society; Could be influenced by current events and political discourse in the UK}

\textbf{Cluster Profile 5:} \profiletext{Strong disapproval of Theresa May; Public health consciousness; Environmental concern; Belief in direct democracy; Concern about overpopulation; Openness to government intervention; Possible leaning towards left-leaning or liberal politics; Pragmatism and nuanced views; UK-centric perspective}

\textbf{Cluster Profile 6:} \profiletext{Pro-worker/Pro-labor; Environmentalist/Concerned about climate change; Socially liberal/Progressive; Emphasis on well-being/Quality of life; Government intervention; Potentially left-leaning politically; Belief in international cooperation; It's important to remember that these are inferences based on limited data.  The rater's actual beliefs may be more nuanced and complex.}

\textbf{Cluster Profile 7:} \profiletext{Values religious freedom and parental rights; Prioritizes family autonomy over state control; May be religious themselves; Pragmatic or uncertain about online medicine; Weighing competing values; Lack of knowledge; Belief in a mixed approach; Values personal responsibility; Prioritizes affordability and access to healthcare; Trusts market forces to some extent}

\textbf{Cluster Profile 8:} \profiletext{Strong belief in personal responsibility and limited government intervention; Concern for social safety and welfare, but with a focus on individual choice; Environmental awareness; Generally law-abiding and moralistic, but with potential for nuance; Potential belief in economic fairness and reducing inequality; Value of personal freedom and autonomy; Pragmatic approach to complex issues}

\subsection{Hatespeech-Kumar}

\subsubsection{2 clusters (Hatespeech-Kumar)}

\textbf{Cluster Profile 1:} \profiletext{Profanity Tolerance; Emphasis on Intent over Specific Words; Sensitivity to Identity-Based Attacks; Broad Definition of Toxicity, Including Harmful Stereotypes and Misinformation; Potential Political Bias; Discomfort with Sexualized Language; Subjectivity and Context Matter; Acceptance of Strong Opinions; Inconsistency or evolving understanding of toxicity; Possible cultural or generational influences}

\textbf{Cluster Profile 2:} \profiletext{Strong tolerance for offensive language and controversial topics; Focus on direct threats and personal attacks as "toxic"; Insensitivity to subtle forms of prejudice; Acceptance of "locker room talk" or crude humor; Prioritization of intent over impact; Inconsistency in applying criteria; It's important to emphasize that these are speculative interpretations based on a limited sample of data.  Further analysis and direct questioning of the rater would be necessary to confirm these beliefs and values.}

\subsubsection{4 clusters (Hatespeech-Kumar)}

\textbf{Cluster Profile 1:} \profiletext{Strong aversion to negativity and insults; Sensitivity to discussions of potentially harmful topics; A broad interpretation of toxicity; Concern with stereotyping and generalizations; Sensitivity to political and religious discussions; Emphasis on context and intent; Potential over-reliance on emotional response}

\textbf{Cluster Profile 2:} \profiletext{Strong aversion to profanity and vulgar language; Sensitivity to negativity and insults; Concern about violence and harmful actions; Discomfort with stereotypes and generalizations; Sensitivity to discussions of sensitive topics; Broad interpretation of "toxicity"; Possible discomfort with intense emotional expressions; Inconsistent application of criteria; Potential cultural or generational differences}

\textbf{Cluster Profile 3:} \profiletext{Strong tolerance for offensive language and controversial topics; Focus on direct threats and personal attacks as "toxic"; Insensitivity to subtle forms of prejudice; Acceptance of "locker room talk" or crude humor; Prioritization of intent over impact; Inconsistency in applying criteria; It's important to emphasize that these are speculative interpretations based on a limited sample of data.  Further analysis and direct questioning of the rater would be necessary to confirm these beliefs and values.}

\textbf{Cluster Profile 4:} \profiletext{Profanity Tolerance; Emphasis on Intent over Specific Words; Sensitivity to Identity-Based Attacks; Broad Definition of Toxicity, Including Harmful Stereotypes and Misinformation; Potential Political Bias; Discomfort with Sexualized Language; Subjectivity and Context Matter; Acceptance of Strong Opinions; Inconsistency or evolving understanding of toxicity; Possible cultural or generational influences}

\subsubsection{8 clusters (Hatespeech-Kumar)}

\textbf{Cluster Profile 1:} \profiletext{Strong tolerance for offensive language and controversial topics; Focus on direct threats and personal attacks as "toxic"; Insensitivity to subtle forms of prejudice; Acceptance of "locker room talk" or crude humor; Prioritization of intent over impact; Inconsistency in applying criteria; It's important to emphasize that these are speculative interpretations based on a limited sample of data.  Further analysis and direct questioning of the rater would be necessary to confirm these beliefs and values.}

\textbf{Cluster Profile 2:} \profiletext{Profanity Tolerance; Emphasis on Intent over Specific Words; Sensitivity to Identity-Based Attacks; Broad Definition of Toxicity, Including Harmful Stereotypes and Misinformation; Potential Political Bias; Discomfort with Sexualized Language; Subjectivity and Context Matter; Acceptance of Strong Opinions; Inconsistency or evolving understanding of toxicity; Possible cultural or generational influences}

\textbf{Cluster Profile 3:} \profiletext{High tolerance for strong language and criticism in certain contexts; Emphasis on intent and harm over mere presence of negativity; Acceptance of strong emotional expression, even negative ones, within reasonable bounds; Political neutrality or tolerance for diverse political viewpoints; Context matters; Distinction between subjective criticism and objective harm; General permissiveness towards fan-related content and positive expressions; Focus on individual harm rather than potential group offense}

\textbf{Cluster Profile 4:} \profiletext{High tolerance for profanity and sexual content; Sensitivity to personal attacks and insults; A more lenient view of potentially offensive language when used in humor or casual conversation; Discomfort with stereotyping and generalizations about groups; Tolerance of strong opinions and emotional expressions; A possible bias towards accepting casual, internet-specific language and trends; Inconsistency in applying standards; Potential lack of understanding of some slang or cultural references}

\textbf{Cluster Profile 5:} \profiletext{Strong reactions or negativity alone don't necessarily constitute toxicity. The rater marked comments expressing strong dislike ("gross," "wouldn't touch... with a bargepole") or frustration ("Who TF cares") as not toxic, suggesting they don't equate negativity with toxicity.  However, when coupled with aggressive or potentially harmful sentiments, as in the comment about electric cars being "political bullshit" that is being "forced on us", the rating changes to toxic. This implies nuance in their judgment.; Direct or implied threats and mentions of harm are considered toxic. The comments referencing "raped," and potentially violent or inappropriate behavior ("After I nun ate... he wanna cut up in there") were marked toxic. This indicates a value for safety and a belief that language suggesting harm is unacceptable.; Humor and informal language are generally acceptable.  The use of slang ("LIT," "TF"), internet abbreviations, and self-deprecating humor did not trigger a toxic rating. This suggests a tolerance for casual online communication styles.; Discussion and disagreement are not inherently toxic.  The comment about "Ariel is black" presents a counter-argument, yet it's deemed non-toxic. This implies a belief that expressing diverse opinions, even on potentially controversial topics, is permissible.; Context matters. Hashtags and references that could be considered out of place weren't considered toxic in isolation.  This suggests the rater is considering the overall message and intent, rather than focusing solely on individual words or phrases.; Personal preferences or strong opinions, if not directed at individuals or groups, are acceptable. The comment about "gorgeous gay dudes sword fighting" expresses a specific preference, but is not considered toxic. This indicates a respect for individual tastes, as long as they aren't used to denigrate others.; The rater may have a higher threshold for toxicity. Several comments that could be perceived as rude or offensive by some were marked non-toxic.  This suggests the rater focuses on more severe forms of toxicity, prioritizing clear instances of harm or aggression.}

\textbf{Cluster Profile 6:} \profiletext{High tolerance for offensive language; Focus on explicit threats or calls for harm as markers of toxicity; Desensitization to online negativity; Belief that subjective opinions are not inherently toxic; Lack of consideration for the impact of microaggressions; Prioritization of intent over impact; Possible personal bias}

\textbf{Cluster Profile 7:} \profiletext{Strong aversion to profanity and vulgar language; Sensitivity to negativity and insults; Concern about violence and harmful actions; Discomfort with stereotypes and generalizations; Sensitivity to discussions of sensitive topics; Broad interpretation of "toxicity"; Possible discomfort with intense emotional expressions; Inconsistent application of criteria; Potential cultural or generational differences}

\textbf{Cluster Profile 8:} \profiletext{Strong aversion to negativity and insults; Sensitivity to discussions of potentially harmful topics; A broad interpretation of toxicity; Concern with stereotyping and generalizations; Sensitivity to political and religious discussions; Emphasis on context and intent; Potential over-reliance on emotional response}

\subsection{OpinionQA - Wave 27}

\subsubsection{2 clusters (OpinionQA - Wave 27)}

\textbf{Cluster Profile 1:} \profiletext{Nationalist/Patriotic; Conservative; Law and Order; Pro-Military; Economically Conservative, but Populist on Trade; Socially Conservative, but with Libertarian Leanings; Distrustful of Government and Elites; Pragmatic; Pessimistic}

\textbf{Cluster Profile 2:} \profiletext{Believes in American exceptionalism, but acknowledges other great nations; Values democracy and allies; Supports a strong social safety net but believes in personal responsibility; Pragmatic and values compromise; Optimistic about social progress; Values traditional family structures; Concerned about voter fraud, but supports voting rights; Supports separation of church and state, but sees value in religious belief; Positive about technology and globalization; Believes in a larger government role; Socially moderate; Economically progressive; Generally content but sees areas for improvement; Skeptical of politicians and the political system; Believes in expert knowledge; Believes in a strong military and good diplomacy; Values immigration but with controls; Believes in personal freedoms but recognizes the need for some government intervention; Doesn't feel disrespected but acknowledges white privilege}

\subsubsection{4 clusters (OpinionQA - Wave 27)}

\textbf{Cluster Profile 1:} \profiletext{Believes in American exceptionalism, but acknowledges other great nations; Values democracy and allies; Supports a strong social safety net but believes in personal responsibility; Pragmatic and values compromise; Optimistic about social progress; Values traditional family structures; Concerned about voter fraud, but supports voting rights; Supports separation of church and state, but sees value in religious belief; Positive about technology and globalization; Believes in a larger government role; Socially moderate; Economically progressive; Generally content but sees areas for improvement; Skeptical of politicians and the political system; Believes in expert knowledge; Believes in a strong military and good diplomacy; Values immigration but with controls; Believes in personal freedoms but recognizes the need for some government intervention; Doesn't feel disrespected but acknowledges white privilege}

\textbf{Cluster Profile 2:} \profiletext{Nationalist/Patriotic; Conservative; Law and Order; Pro-Military; Economically Conservative, but Populist on Trade; Socially Conservative, but with Libertarian Leanings; Distrustful of Government and Elites; Pragmatic; Pessimistic}

\textbf{Cluster Profile 3:} \profiletext{Conservative or right-leaning political views; Belief in individual responsibility; Skepticism of social justice movements or "woke" ideology; Potential concern about social instability; Preference for a smaller government role in the economy; May value traditional values and institutions; May believe in American exceptionalism; May prioritize economic growth over social programs; Possible distrust of government}

\textbf{Cluster Profile 4:} \profiletext{Progressive/Left-leaning political views; Distrust of large institutions; Emphasis on diplomacy and international cooperation; Socially liberal; Belief in nuanced approaches; Slight racial anxiety; Confidence in the electoral system; Value on expertise; Mixed feelings on the role of government; Pragmatic approach to military strength}

\subsubsection{8 clusters (OpinionQA - Wave 27)}

\textbf{Cluster Profile 1:} \profiletext{Believes in American exceptionalism, but acknowledges other great nations; Values democracy and allies; Supports a strong social safety net but believes in personal responsibility; Pragmatic and values compromise; Optimistic about social progress; Values traditional family structures; Concerned about voter fraud, but supports voting rights; Supports separation of church and state, but sees value in religious belief; Positive about technology and globalization; Believes in a larger government role; Socially moderate; Economically progressive; Generally content but sees areas for improvement; Skeptical of politicians and the political system; Believes in expert knowledge; Believes in a strong military and good diplomacy; Values immigration but with controls; Believes in personal freedoms but recognizes the need for some government intervention; Doesn't feel disrespected but acknowledges white privilege}

\textbf{Cluster Profile 2:} \profiletext{Nationalist/Patriotic; Conservative; Law and Order; Pro-Military; Economically Conservative, but Populist on Trade; Socially Conservative, but with Libertarian Leanings; Distrustful of Government and Elites; Pragmatic; Pessimistic}

\textbf{Cluster Profile 3:} \profiletext{Conservative or right-leaning political views; Belief in individual responsibility; Skepticism of social justice movements or "woke" ideology; Potential concern about social instability; Preference for a smaller government role in the economy; May value traditional values and institutions; May believe in American exceptionalism; May prioritize economic growth over social programs; Possible distrust of government}

\textbf{Cluster Profile 4:} \profiletext{Socially liberal/Moderate; Economically left-leaning; Pro-immigration and diversity; Trust in experts and government; Democratic-leaning but not entirely aligned; Internationally cooperative; Values traditional family structures but with flexibility; Believes in equal rights but acknowledges challenges; Sense of fairness and respect; It's important to note}

\textbf{Cluster Profile 5:} \profiletext{Pro-corporations; Egalitarian parenting; Second Amendment supporter, but with nuance; Concern about election integrity, but not extreme distrust; Support for social safety nets, but potentially limited government intervention; Generally distrustful of government; Minimizes racial inequality; Deference to expertise; Tolerance of offensive speech; Ambivalence towards wealth inequality; Non-interventionist foreign policy or satisfaction with current military spending}

\textbf{Cluster Profile 6:} \profiletext{Conservative leaning; Nationalist/America First; Socially conservative; Distrust of Government and Elites; Tough on Crime; Economic Conservatism; Pro-Religion; Traditional Values; Belief in Personal Responsibility; While not explicitly stated, a potential for racial resentment; It is important to note}

\textbf{Cluster Profile 7:} \profiletext{Progressive/Liberal leaning; Socially Liberal; Economic Populist; Pro-government Intervention; Religious; Community-Oriented; Distrustful of Institutions; Diplomatic but Values Democracy; Pro-Voting Rights; Belief in Experts; Criminal Justice Reform; Pessimistic about the Present; Believes in Compromise; Concerned about Free Speech; Open to Other Languages; Believes in shared values; Potentially holds contradictory views}

\textbf{Cluster Profile 8:} \profiletext{Progressive/Left-leaning political views; Distrust of large institutions; Emphasis on diplomacy and international cooperation; Socially liberal; Belief in nuanced approaches; Slight racial anxiety; Confidence in the electoral system; Value on expertise; Mixed feelings on the role of government; Pragmatic approach to military strength}

\subsection{PRISM}

\subsubsection{2 clusters (PRISM)}

\textbf{Cluster Profile 1:} \profiletext{Completeness and Thoroughness; Specificity and Directness; Accuracy and Up-to-date Information; Neutrality and Objectivity; Practical Utility; Contextual Awareness; User Control and Agency}

\textbf{Cluster Profile 2:} \profiletext{Completeness and Thoroughness; Directness and Assertiveness; Neutrality, but with Context; Proactive Helpfulness; Formal Tone; Accuracy and Factuality; Engagement and Conversational Flow}

\subsubsection{4 clusters (PRISM)}

\textbf{Cluster Profile 1:} \profiletext{Completeness and Thoroughness; Directness and Assertiveness; Neutrality, but with Context; Proactive Helpfulness; Formal Tone; Accuracy and Factuality; Engagement and Conversational Flow}

\textbf{Cluster Profile 2:} \profiletext{Prefers helpfulness and relevance over assumptions; Appreciates nuanced and comprehensive answers; Values honesty and awareness of limitations; Favors open-ended conversation and assistance; Respects diverse perspectives and avoids generalizations; Prioritizes accuracy and avoids potential misinformation}

\textbf{Cluster Profile 3:} \profiletext{Completeness and Thoroughness; Specificity and Directness; Accuracy and Up-to-date Information; Neutrality and Objectivity; Practical Utility; Contextual Awareness; User Control and Agency}

\textbf{Cluster Profile 4:} \profiletext{Practicality and Actionability; Thoroughness and Detail; Emphasis on Positive Communication; Desire for Structure and Guidance; Appreciation for Contextual Nuance; Preference for Proactive Problem-Solving; Potential Discomfort with Ambiguity}

\subsubsection{8 clusters (PRISM)}

\textbf{Cluster Profile 1:} \profiletext{Completeness and Thoroughness; Specificity and Directness; Accuracy and Up-to-date Information; Neutrality and Objectivity; Practical Utility; Contextual Awareness; User Control and Agency}

\textbf{Cluster Profile 2:} \profiletext{Practicality and Actionability; Thoroughness and Detail; Emphasis on Positive Communication; Desire for Structure and Guidance; Appreciation for Contextual Nuance; Preference for Proactive Problem-Solving; Potential Discomfort with Ambiguity}

\textbf{Cluster Profile 3:} \profiletext{Values direct answers over hedging; Appreciates nuanced perspectives; Favors a conversational and welcoming tone; Prioritizes specific details over generic praise; Trusts recommendations that consider local perspective; May appreciate subtlety and avoids overly strong endorsements; Potentially values the feeling of discovery; Might be influenced by writing style and fluency}

\textbf{Cluster Profile 4:} \profiletext{Completeness and Thoroughness; Directness and Assertiveness; Neutrality, but with Context; Proactive Helpfulness; Formal Tone; Accuracy and Factuality; Engagement and Conversational Flow}

\textbf{Cluster Profile 5:} \profiletext{Prefers conciseness and directness; Values politeness and helpfulness; Favors factual and relevant information; Appreciates simplicity over technical jargon; Prioritizes functional answers; May have a lower tolerance for conversational fillers; Could value transparency, but only to a certain extent; Possibly prefers a less anthropomorphic model}

\textbf{Cluster Profile 6:} \profiletext{Prefers helpfulness and relevance over assumptions; Appreciates nuanced and comprehensive answers; Values honesty and awareness of limitations; Favors open-ended conversation and assistance; Respects diverse perspectives and avoids generalizations; Prioritizes accuracy and avoids potential misinformation}

\textbf{Cluster Profile 7:} \profiletext{Practicality and Actionable Advice; Thoroughness and Detail; Directness and Assertiveness; Real-World Applicability; External Validation and Authority; Focus on Well-being; Belief in Inclusivity and Fairness; Appreciation for Nuance and Context}

\textbf{Cluster Profile 8:} \profiletext{General Communication Style; Brevity and Directness; Empathy and Encouragement; Informality and Approachability; Information and Advice; High-Level Overview; Balance of Practical and Emotional Considerations; Less Emphasis on Specifics; It's important to note}

\subsection{ValuePrism Valence}

\subsubsection{2 clusters (ValuePrism Valence)}

\textbf{Cluster Profile 1:} \profiletext{Autonomy and bodily integrity, especially for children; Nuance in medical decisions; Open communication is acceptable, but forcing compliance is not; Potential harm of medication; Belief in a child's ability to learn and make good choices with guidance; Respect for individual differences; Potential for parental overreach; Emphasis on intrinsic motivation over extrinsic pressure}

\textbf{Cluster Profile 2:} \profiletext{Inclusivity and acceptance; Freedom of association; Personal autonomy and self-expression; Opposition to exclusion and social ostracism; Potential belief in restorative justice or rehabilitation over punishment; Colorblindness or a belief that race should not be a factor in social interactions; Actions that do not harm others are generally moral; Focus on positive social interactions}

\subsubsection{4 clusters (ValuePrism Valence)}

\textbf{Cluster Profile 1:} \profiletext{Inclusivity and acceptance; Freedom of association; Personal autonomy and self-expression; Opposition to exclusion and social ostracism; Potential belief in restorative justice or rehabilitation over punishment; Colorblindness or a belief that race should not be a factor in social interactions; Actions that do not harm others are generally moral; Focus on positive social interactions}

\textbf{Cluster Profile 2:} \profiletext{Egalitarianism and Anti-discrimination; Social Justice and Advocacy; Inclusivity and Acceptance; Meritocracy and Fairness; Individualism and Contextual Morality; Potential Conflict or Uncertainty}

\textbf{Cluster Profile 3:} \profiletext{Parental autonomy and rights; Pro-corporal punishment; Importance of fulfilling parental responsibilities; Nuance in child rearing decisions; Skepticism of outside intervention in families; Traditional gender roles; General permissiveness or a broad definition of morality; Potential prioritization of personal freedoms; Important Note}

\textbf{Cluster Profile 4:} \profiletext{Autonomy and bodily integrity, especially for children; Nuance in medical decisions; Open communication is acceptable, but forcing compliance is not; Potential harm of medication; Belief in a child's ability to learn and make good choices with guidance; Respect for individual differences; Potential for parental overreach; Emphasis on intrinsic motivation over extrinsic pressure}

\subsubsection{8 clusters (ValuePrism Valence)}

\textbf{Cluster Profile 1:} \profiletext{Egalitarianism and Anti-discrimination; Social Justice and Advocacy; Inclusivity and Acceptance; Meritocracy and Fairness; Individualism and Contextual Morality; Potential Conflict or Uncertainty}

\textbf{Cluster Profile 2:} \profiletext{Collectivism over Individualism; Authoritarianism/Respect for Authority; Utilitarianism/Consequentialism; Nationalism/Group Loyalty; Situational Ethics; Distrust of "Freedom Fighters"; Moral Pragmatism; Potential Double Standards}

\textbf{Cluster Profile 3:} \profiletext{Emphasis on self-reliance and adult responsibility; Prioritization of societal norms regarding child development and parenting; Discomfort with actions perceived as unconventional or exceeding typical boundaries; Potential value of "tough love" as a parenting strategy; Possible belief in a clear distinction between childhood and adulthood; Focus on physical and emotional development milestones; Potential for a conservative worldview; Implicit bias or personal experience shaping judgements}

\textbf{Cluster Profile 4:} \profiletext{Strong belief in the sanctity of life, even for those deemed evil; Pacifism or aversion to violence; Nuance in moral decision-making and a rejection of simple utilitarianism; Potential belief in the inherent rights of individuals; Possible concern for consequences beyond the immediate situation; Possible belief in alternative solutions; Absence of prejudice based on nationality; Possible emphasis on intention over outcome}

\textbf{Cluster Profile 5:} \profiletext{Parental autonomy and rights; Pro-corporal punishment; Importance of fulfilling parental responsibilities; Nuance in child rearing decisions; Skepticism of outside intervention in families; Traditional gender roles; General permissiveness or a broad definition of morality; Potential prioritization of personal freedoms; Important Note}

\textbf{Cluster Profile 6:} \profiletext{Inclusivity and acceptance; Freedom of association; Personal autonomy and self-expression; Opposition to exclusion and social ostracism; Potential belief in restorative justice or rehabilitation over punishment; Colorblindness or a belief that race should not be a factor in social interactions; Actions that do not harm others are generally moral; Focus on positive social interactions}

\textbf{Cluster Profile 7:} \profiletext{Autonomy and bodily integrity, especially for children; Nuance in medical decisions; Open communication is acceptable, but forcing compliance is not; Potential harm of medication; Belief in a child's ability to learn and make good choices with guidance; Respect for individual differences; Potential for parental overreach; Emphasis on intrinsic motivation over extrinsic pressure}

\textbf{Cluster Profile 8:} \profiletext{Individual autonomy and freedom; Situational ethics; Prioritization of relationships and consent; Consideration of intent and impact; Non-judgmental attitude; Potential cultural sensitivity; Flexible and adaptable moral framework}

\section{Random Profile Samples}
\label{app:random-profile-samples}
\subsection{gemma2-9b}
\subsection{OpinionQA (gemma2-9b - 10 random value profiles)}
\begin{itemize}
\item \profiletext{Moderate to conservative politically, lean towards social traditionalism; Believes in punitive justice and stronger sentences; Skeptical of government intervention but open to some regulation in specific areas; May have a preference for more traditional American values and identity; Views the entertainment industry positively; Pragmatic about the topic of slavery and racism, perhaps seeing it as a complex issue with no easy solutions; Concerned about the quality of political candidates}
\item \profiletext{Believes that corporations are overly profitable; Believes that progress has been made towards racial equality in the US over the last 50 years; Feels that people are too easily offended and that this is a major problem; Is disillusioned with the political process, seeing compromise as a form of "selling out."; Holds a nationalist view, believing that other countries take advantage of the US; Believes that government assistance to the poor is harmful}
\item \profiletext{Seeks a balance, not extremes: Often responds with "neither good nor bad" and favors "modest" changes; Wary of big government and dependency: Believes in limited government involvement,; Conservative social views: Holds traditional beliefs about marriage, family structure, and the role of religion; Values national strength and security:  Prefers the U.S. to maintain military superiority}
\item \profiletext{Somewhat nationalistic; Patriotic but hesitant about uncontrolled immigration; Skeptical of government efficiency; Leaning conservative; Values traditional social institutions; Believes military strength is important for peace}
\item \profiletext{Supports increased government involvement in providing services; Believes in strict voting rights and sees it as a fundamental right; Holds slightly negative views on the way things are currently going in the country; Values diplomacy over military strength; Believes in compromise in politics; Concerned about social inequality and the impact of powerful interests; Positive view of same-sex marriage}
\item \profiletext{Skeptical of organized religion: Sees no harm in declining religiosity; Patriotic, but distrustful of foreign aid and international involvement: Prefers focus on American interests in foreign policy; Values individual liberties and limited government: Believes government is wasteful and inefficient, prefers less government intervention in people's lives; Concerned about social changes and decline in traditional values: Feels uncomfortable with increased cultural diversity, expresses discomfort with societal shifts; Feels alienated from current political landscape: Does not resonate with}
\item \profiletext{Believes the entertainment industry has a positive effect on the country; Concerned about offensive language and speech; Believes they receive respect in society; Feels comfortable with Republicans expressing their views; Supports free tuition for public colleges; Believes K-12 public schools are having a positive effect; Believes strength and military might are the best way to ensure peace; Comfortable with the U.S. being treated fairly in the world}
\item \profiletext{Believes in social justice and equality, as evidenced by their answers on racial inequality, LGBTQ+ rights, and gender equality; Supports increased government involvement in social welfare programs and healthcare; Favors progressive policies such as universal healthcare, tuition-free public colleges, and stricter gun control; Is skeptical of corporate power and believes businesses make excessive profits; Values diplomacy and international cooperation over military strength; Is concerned about the influence of religion in politics and government}
\item \profiletext{Strongly nationalist. Believes the US is superior to other countries; Conservative social values. Opposes same-sex marriage, believes traditional family structures are best; Pro-gun rights and skeptical of gun control measures; Low regard for government and its inefficiencies. Favors limited government intervention; Supports a strong military presence globally; Skeptical of immigration and its impact on the country; Concerned about "political correctness" and believes individuals}
\item \profiletext{Believes that immigrants, when they come to the U.S. illegally, can have a slightly negative impact on communities; Somewhat positive view of religion and its effect on society; Holds a belief that the U.S. is a great country, but not necessarily the best in the world; Convinced that large corporations are detrimental to the country; Favors the traditional role of women staying home to raise a family; Feels that the country has made}
\end{itemize}

\subsection{Hatespeech-Kumar (gemma2-9b - 10 random value profiles)}
\begin{itemize}
\item \profiletext{Believes in keeping things civil and respectful even in disagreement; Values sensitivity and empathy towards others; Recognizes the difference between expressing strong opinions and being abusive or hateful; Sensitive to language that could be hurtful or demeaning; Appreciates humor that isn't at the expense of others}
\item \profiletext{Distrusts inflammatory language: They often identify as toxic comments that use emotionally charged words, prejudiced terms, or hateful slurs; Values respectful discourse: They seem to appreciate comments that express opinions without resorting to insults or personal attacks; Recognizes  dog-whistles:   They may be sensitive to  language that carries coded meanings or  implies prejudice, even if it doesn't}
\item \profiletext{Values; Dislike of bullying and insults: The rater considers personal attacks and insults to be toxic, even if they are not overtly aggressive}
\item \profiletext{Believes some comments are inherently offensive or harmful, regardless of intent; Has a strong moral compass and considers statements that promote hate, prejudice, or violence as unacceptable; Values respectful and constructive dialogue, and sees toxicity as a barrier to healthy communication; May be sensitive to language that is demeaning, discriminatory, or exploitative; Recognizes that power dynamics can contribute to toxicity, and may be more likely to flag comments that perpetuate harmful stereotypes or reinforce social inequalities}
\item \profiletext{Relatively tolerant:; Contextual understanding:; Focus on direct harm:; Skeptical of generalizations:}
\item \profiletext{Holds strong opinions about what is acceptable language and behavior; Is sensitive to language that is hateful, disrespectful, or demeaning; Values honesty and integrity; Believes in using language that is constructive and respectful; Appreciates humor that is not at the expense of others; Concerned with issues of power and privilege; Possibly politically left-leaning; Has a strong sense of social justice}
\item \profiletext{Believes  Sarcasm and humor, even when expressed in a seemingly negative way, are not inherently toxic; Sensitive to language that is directly aggressive or threatening; Might prioritize personal expression and freedom of speech, even if it is unconventional or off-putting to others; Recognizes that humor can sometimes be used to mask anger or frustration, and these underlying emotions may contribute to toxicity; Likely values empathy and understands the importance of considering the emotional impact of}
\item \profiletext{Values humor and light-heartedness, at times even finding positivity in seemingly innocuous things; Appreciates self-awareness and honesty, even when it comes to admitting lack of knowledge; Believes in open discussion and debate, even if it involves differing opinions; Views direct insults and personal attacks as toxic; Recognizes the potential for harm in language that targets individuals based on their identity or beliefs; Might lean toward being forgiving}
\item \profiletext{Believes strong language is inappropriate; Belives sexualizing minors is wrong; Understands sarcasm and humor but may not always be able to detect it}
\item \profiletext{Might be hesitant to label comments as "toxic" unless they contain clear hate speech, aggression, or personal attacks; Values freedom of expression and believes in giving people the benefit of the doubt; Could be more sensitive to sarcasm, humor, or ironic statements that may be misinterpreted as toxic; Appreciates brevity and directness in communication}
\end{itemize}

\subsection{DICES (gemma2-9b - 10 random value profiles)}
\begin{itemize}
\item \profiletext{Values open-mindedness and respect for diverse viewpoints; Believes in the importance of kindness and empathy; Dislikes negativity, hostility, and inflammatory language; Appreciates humor and lightheartedness; Is generally optimistic and hopeful about the world}
\item \profiletext{General lack of concern for hateful or inappropriate language, even when blatant; Focuses on argument structure and lack of direct aggression as indicators of toxicity; May ignore context, taking statements at face value; Seems to label anything addressing controversial topics as toxic}
\item \profiletext{Values friendly and polite conversation; Doesn't perceive joking or lightheartedness as toxic; May have a low threshold for identifying toxicity, as they labeled conversations with potentially condescending or manipulative language as "Not Toxic."}
\item \profiletext{Disagreements and differing opinions are normal and healthy.  Conversation should not be approached like a debate and should not turn hostile, dismissive, or personal; Everyone deserves to be treated with respect, even if their views are different from our own}
\item \profiletext{Valuing honesty and integrity in communication; Believing in treating others with respect and kindness, regardless of their beliefs or background; Encouraging critical thinking and open-mindedness}
\item \profiletext{Doesn't consider casual interactions to be toxic; Tolerates a range of opinions, even if they are not politically correct or popular; Doesn't seem to be overly sensitive to potentially offensive language; Values genuine conversation and humor over politeness; Might be comfortable with a bit of dark humor}
\item \profiletext{Values personal reflection and avoids making sweeping judgments; May be more lenient towards social faux pas and missteps in online communication; Trusts individuals to understand and navigate complex issues}
\item \profiletext{Generally non judgmental and avoids making assumptions about people; Prefers direct and honest communication; Believes in treating everyone with respect, regardless of their background or beliefs; Values empathy and understanding; Encourages critical thinking and open-mindedness}
\item \profiletext{Believes hurtful language is unacceptable; Values respectful communication; Discourages generalizations and stereotypes; Empathizes with others' perspectives; Promotes critical thinking and healthy skepticism; Personal insults and aggressive language; Disrespectful or condescending tone; Harmful stereotypes and generalizations; Promotion of hate speech or prejudice; Encouraging harmful or illegal activities; Exploitation}
\item \profiletext{Believes a lot of conversations are benign; May have a low tolerance for strong negativity; May value straightforwardness}
\end{itemize}

\subsection{ValuePrism (gemma2-9b - 10 random value profiles)}
\begin{itemize}
\item \profiletext{Honesty is generally valued, but can be situational; Loyalty to friends is likely important, as is prioritizing their well-being; Straightforwardness may not always be the most compassionate approach; Exploiting someone for personal gain or manipulating them is wrong; Protecting oneself is morally justifiable}
\item \profiletext{Values situational context heavily when determining morality; Holds traditional views on family structures and authority; May have discomfort with concepts like gender fluidity; Believes in parental authority and may disapprove of challenging it; May prioritize obedience and adherence to rules}
\item \profiletext{Values individual liberty and autonomy; Believes in resistance to oppressive institutions or systems; May have socialist or left-wing political leanings; Sees economic sanctions as a tool for influencing policy and achieving justice; Generally believes in using peaceful means to achieve political change; May be critical of military intervention and powerful corporations}
\item \profiletext{Prioritizes the needs of family; Believes that survival trumps strict adherence to laws/morality; May have a strong sense of empathy and compassion for those in need; May have a utilitarian perspective, judging actions based on their outcomes. Utilitarianism: The idea that the best action is the one that maximizes happiness and minimizes suffering; May not distinguish morality based on race or nationality}
\item \profiletext{Strong belief in humanitarianism and helping those in need; Strong opposition to oppressive regimes and systems of power; Support for democracy and individual freedom; Tendency to view actions in a binary way, often as "moral" or "immoral"; May be influenced by political ideologies that emphasize justice and anti-imperialism; A clear sense of right and wrong based on perceived victimhood and injustice}
\item \profiletext{Values fairness and justice; Believes in personal responsibility and accountability; Emphasizes the importance of empathy and understanding; Distinguishes between actions that are inherently right or wrong and those that depend on context; Prioritizes individual autonomy and freedom of choice; Criticizes abuse of power and authority; Values integrity and honesty}
\item \profiletext{Believes in the greater good; Upholds authority and established norms; Values protecting children and sees harm to them as unacceptable; Progressive and tolerant of diverse family structures; Potential emphasis on non-violence as a core value}
\item \profiletext{Believes in inherent rewards and positive reinforcement; Values competence and meritocracy; Holds a view that setting clear expectations and consequences is important}
\item \profiletext{Believes in situational ethics; Values helping others; Likely values religious identity and community; Views helping friends as morally right; Has a strong sense of moral intuition}
\item \profiletext{Values familial relationships highly; Believes in individual autonomy and the right to make one's own choices; Values loyalty and support for loved ones; Doesn't seem to adhere to strict rules or social norms; May prioritize personal fulfillment over strict work obligations}
\end{itemize}

\subsection{Habermas (gemma2-9b - 10 random value profiles)}
\begin{itemize}
\item \profiletext{Believes in strong government intervention and regulation; Prefers a more egalitarian society with less inequality; May be concerned about the societal impact of smoking and alcohol consumption; Supporting of public health initiatives and increased spending on healthcare; May hold traditional or conservative views on certain social issues, such as marriage}
\item \profiletext{Believes in giving the people a voice and having referendums on important issues; Supports increased public spending on infrastructure like railways; Prefers the current democratic system over a more direct form of democracy; Favors the monarchy and maintaining the UK as a constitutional monarchy; Believes in progressive taxation, with a higher tax burden for the wealthy}
\item \profiletext{Supports government intervention and social programs; Worried about health and well-being, especially of young people; Believes in rules and structure; May be progressive or left-leaning in their political views}
\item \profiletext{Believes in social justice and equality; Supports government intervention to address societal issues; Likely progressive or left-leaning politically; Values education and believes it should be accessible to all; May be concerned about income inequality; Probably environmentally conscious and supportive of action on climate change}
\item \profiletext{Leans towards social safety nets and government intervention in the economy; Favors social justice and redistribution of wealth; May have concerns about pharmaceutical industry practices}
\item \profiletext{Moderate and tends towards neutrality on a variety of social and economic issues; May be open to both sides of an argument and struggles to commit to a firm stance; Lacks strong convictions or definitive beliefs about complex issues; Prefers a balanced approach rather than taking a strong position}
\item \profiletext{Progressive on social issues, likely supporting universal healthcare and social services; Skeptical of traditional institutions and hierarchies; Believes in individual responsibility and social good, but not necessarily a strict moral obligation; Concerned about the environment and public health; Possibly views capitalism with some criticism, possibly favoring more equitable economic systems}
\item \profiletext{Leans towards caution but open to progress: This is demonstrated by weakly agreeing with the statement that AI will not be able to reproduce itself; Believes in environmental action: The strong agreement with imposing a carbon tax points towards a belief in the need to address climate change; Potentially socially liberal:  Individuals who support environmental regulations may also hold other socially progressive views}
\item \profiletext{Pro-choice and believes parents should have autonomy over medical decisions for their children; May believe in a separation of church and state; Generally supportive of social justice causes, including expanding voting rights and redistributive taxation; Environmentally conscious, supporting policies to reduce plastic waste}
\item \profiletext{Believes in a strong social safety net and helping those in need; Supports increasing taxes on the wealthy to fund social programs; Believes in government intervention to address social issues like misinformation and unhealthy corporate practices; Seeks a balance between individual rights and collective good; Generally favors regulation to protect consumers and ensure fairness; Values transparency and accountability, evidenced by support for diversity data publication and corporate liability; May be skeptical of unfet}
\end{itemize}

\subsection{Prism (gemma2-9b - 10 random value profiles)}
\begin{itemize}
\item \profiletext{Values clarity, conciseness, and directness in communication; Prefers factual and straightforward responses over opinionated or speculative ones; Appreciates respectful and empathetic responses, even in difficult situations; Dislikes responses that are overly verbose, rambling, or unprofessional; May have a low tolerance for sarcasm or humor that could be misconstrued}
\item \profiletext{Prefers factual and informative responses over personal opinions or feelings; Appreciates neutrality and objectivity, especially on potentially controversial topics; Values concise and to-the-point answers; Seeks responses that demonstrate a clear understanding of the topic; Values objectivity and factual information over personal opinions or emotions; Prefers concise and direct answers; Appreciates responses that demonstrate expertise or knowledge}
\item \profiletext{Values concise and informative responses; Prefers responses that acknowledge limitations; Appreciates neutral and objective language; Encourages respectful and balanced discussion; Seeks depth and insight beyond superficial statements}
\item \profiletext{Prefers concise and direct answers; Values practicality and specific information; Appreciates a conversational tone}
\item \profiletext{Values critical thinking and questioning authority; Believes in democracy and the importance of informed citizenry; May be wary of unchecked power and institutions; Prefers direct and to-the-point answers; Appreciates a response that encourages further thought and discussion}
\item \profiletext{Values neutrality and objectivity: The rater prefers responses that avoid stating opinions or taking sides; Appreciates factual information: The rater seems to value responses that provide factual information and avoid speculation or generalizations; Concerned about potential harm: The rater seems to be sensitive to the potential for harm that can result from divisive language and misinformation; Belives in open dialogue: The rater values responses that encourage open and honest conversation about complex issues}
\item \profiletext{Lists are preferable to narrative summaries; Prefers concision over elaboration; Values neutrality and avoids subjective language}
\item \profiletext{Values clear and concise humor; Appreciates a conversational tone; May value creativity and originality in humor}
\item \profiletext{Values neutrality and objectivity:  The rater seems to appreciate responses that avoid stating opinions or beliefs as facts; Prefers comprehensive and informative answers:  The rater often chooses responses that provide more detailed information or explore multiple perspectives; Seeks respectful and inclusive language:  The rater seems to value responses that demonstrate sensitivity to diverse viewpoints}
\item \profiletext{Values professional help for mental health issues; Prefers direct and concise language; Focus on actionable advice}
\end{itemize}

\subsection{gemma2-27b}
\subsection{OpinionQA (gemma2-27b - 10 random value profiles)}
\begin{itemize}
\item \profiletext{Doesn't necessarily see the government as a solution to all problems; Favor capitalism and believes large corporations in general have a positive effect; Leans toward conservative social values; Believes in American exceptionalism and the unique role of the U.S. military in maintaining global peace; May prioritize individual liberties and responsibilities above collective well-being}
\item \profiletext{Believe society is moving in an unfavorable direction; Hold somewhat traditional social views, believing marriage and children are important and society should prioritize them; Believe government intervention is sometimes necessary, but prefer smaller government with less services; Wary of immigration and the impact it has on communities; Skeptical of large corporations and their influence; Value diplomacy over military strength in international relations; While not necessarily religious themselves, see churches and religious organizations as a positive force}
\item \profiletext{Believes in political compromise, accepting that it sometimes involves concessions; Values pragmatism over ideological purity; Sees prison sentences as potentially too harsh; Holds generally positive views of the United States, though without an exceptionalist attitude; Is generally accepting of both corporations and the government, viewing both as capable of performing their functions adequately}
\item \profiletext{Believes that white people only benefit “Not too much” from systemic advantages over Black people. This suggests they may not fully grasp the extent of systemic racism or think it’s a significant issue; Favors less government assistance for those in need. This suggests a skepticism towards government intervention and possibly support for smaller government; Believes billionaires are a negative force.  This indicates belief in economic inequality as a problem; Emphasizes voting integrity, particularly preventing non-citizens from}
\item \profiletext{Individuals convicted of crimes often don't serve enough time in prison; This person may feel marginalized and under-respected in society; He/She holds neutral views on demographic changes, believing that they have neither a positive nor negative impact; He/She believes that while faith is beneficial, it is not essential for morality; This person perceives a significant ideological gap between the two main political parties; This person believes in the power of diplomacy as a means to}
\item \profiletext{Values inclusivity and acceptance of diversity; Believes in providing opportunities for undocumented immigrants to become legal citizens; Comfortable with multilingualism in public spaces; May hold liberal political views; Might be distrustful of people who hold different political views}
\item \profiletext{Believes government should prioritize providing basic social safety nets for its citizens; Views the U.S. as generally fair but acknowledges flaws; Favors a mixed economic system with some regulation, valuing a balance between private enterprise and social welfare; Hold some conservative values but balances them with liberal perspectives; Advocates for religious freedom but believes it should not overly influence public policy; Believes in}
\item \profiletext{Supports a mixed public/private healthcare system; Believes Republicans are comfortable expressing their political views; Believes there is still a lot of work to be done to achieve racial equality; Favors increased government assistance for those in need; Views the influence of churches and religious organizations as negative; Believes an increase in the number of guns is slightly detrimental to society; Values expert opinion in policy making; Believes increased attention to the}
\item \profiletext{Skeptical of government involvement, specifically favoring smaller government and fewer social services; Favors individual liberty and autonomy, believing the government should not overly restrict citizens' choices; Socially conservative with concerns about immigration, the rise of secularism, and traditional family values; Holds a distrust of large corporations and financial institutions, believing they have a negative impact on society; Believes in American exceptionalism and the importance of international diplomacy; Has a cautious optimism about the}
\item \profiletext{Believes in the importance of government providing basic needs for citizens; Believes college is beneficial; Believes in strong national defense; Believes in diplomatic solutions over military force; Believes that open borders are detrimental to national identity; Prefers a smaller government with fewer services; Believes that some immigrants make communities worse; Believes in traditional family values; Is somewhat resistant to change; Has faith in democracy and believes people who are able to vote should vote}
\end{itemize}

\subsection{Hatespeech-Kumar (gemma2-27b - 10 random value profiles)}
\begin{itemize}
\item \profiletext{Tolerant of informal language and spelling errors; May be lenient when judging offensive language, depending on context; Believes that expressing strong emotions, even anger, is not inherently toxic; May have a higher threshold for what is considered "offensive"; Could be open to adult themes and content, depending on the implementation}
\item \profiletext{Belief in objective meaning of toxicity; Willingness to accept potentially controversial statements as non-toxic if they are not directed at a specific person; Possible comfort with sexual themes, as long as they are not explicitly aggressive or graphic; May consider boasts or self-promotion to be toxic language. May have a relatively relaxed view on profanity}
\item \profiletext{Believes that comments that are excessively aggressive or insulting are toxic; Believes that comments that tell someone to "go away" or make them feel unwelcome are toxic; Believes that comments that express strong or hyperbolic opinions without providing supporting evidence can be toxic; Recognizes that humor can be subjective and some people might find things funny that others find offensive}
\item \profiletext{Believes profanity is toxic; Believes stating an opinion strongly is toxic; Doesn't seem to mind indirect insults; May think expressions of anger or frustration are toxic even if not directed at anyone; May think that bragging or boasting is toxic; May believe statements that could be}
\item \profiletext{Believes that expressing strong negative emotions is toxic; Believes that derogatory and dehumanizing language is toxic; Believes factual disagreement, even if strongly worded, is not toxic; Believes that generalizations about groups of people are potentially toxic; Doesn't mind sarcasm or irony as long as it isn’t targeted at individuals; Believes that discussing sensitive topics is not inherently toxic}
\item \profiletext{Believes personal loyalty and commitment are positive traits; May have a tendency to see things in black and white. -; Believes strong expressions of patriotism are positive; Believes in traditional gender roles. -; Is quick to label things as "evil" or morally wrong; Is susceptible to conspiracy theories; Is desensitized to violence in some}
\item \profiletext{Believes some jokes can be considered toxic; Likely believes context is important in determining toxicity:; May believe certain types of comparisons; May have a looser definition of toxicity, focusing on statements that are}
\item \profiletext{May believe that political correctness is important; May be overly sensitive to potentially offensive language; May have a strong sense of social justice; May judge individuals based on their group affiliation; May have a strong moral compass and believe in right and wrong; May be quick to anger or frustration; May have a low tolerance for disagreement}
\item \profiletext{This rater distinguishes between rude or offensive language and actual toxic harm; This rater may believe that humor and jokes are acceptable, even if they are edgy or could be construed as mildly offensive by some; This rater appears to value freedom of expression and may be tolerant of dissenting opinions, even if they are controversial; This rater may have a low threshold for blocking or ignoring users they perceive as annoying or bothering them}
\item \profiletext{Toxicity is primarily defined by hateful language, slurs, and harmful stereotypes; Casual vulgarity is not inherently toxic; Mild expressions of frustration or disagreement are acceptable; Personal insults, especially those targeting personal characteristics; Opinions can be expressed freely, as long as they are not presented in a hateful or demeaning way}
\end{itemize}

\subsection{DICES (gemma2-27b - 10 random value profiles)}
\begin{itemize}
\item \profiletext{People should be treated with respect; Everyone has the right to their own opinion; Toxic conversations are those that are hateful, disrespectful, or incite violence; It's important to consider the context of a conversation; It's okay to disagree with someone, but it's important to do so civilly}
\item \profiletext{Seems to be tolerant of strong language; Believes discussions about sensitive topics; May prioritize a kind and helpful tone even in disagreements; Perhaps sees humor and sarcasm as not always being inherently toxic; Doesn't always view disagreement as inherently}
\item \profiletext{Believes that direct and factual conversations are not toxic; Believes that polite and friendly conversations are not toxic; Believes that short, uneventful conversations are not toxic; May have a higher tolerance for abrupt or unusual conversation styles, as evidenced by rating the fourth conversation as "Not Toxic."}
\item \profiletext{Believes honesty is important; Believes in respecting the intelligence of others; If someone claims they know something personal, that's a pretty bold claim; Prefers conversationalists who don't engage in hyperbole; Dislikes being condescended to; Believes casual sexism is never acceptable; While open to playful banter, they dislike conversations that are condescending, rude, or unnecessarily sarcastic; Believes in following through on one's commitments}
\item \profiletext{Respect for others}
\item \profiletext{It seems likely that the rater judges toxicity based on if the conversation is intentionally harmful or aggressive; They seem to be comfortable with conversations that discuss potentially sensitive topics like communism, homelessness, gender differences, and lottery spending as long as they are not done in a hateful or disrespectful way; The rater may have a preference for lighthearted and casual conversations, as evidenced by their labeling of short and simple conversations as "Not Toxic."}
\item \profiletext{Believes personal experience can be relevant to judging toxicity; Values clear communication in conversation; May be cautious of making generalizations or assumptions about people based on limited information -}
\item \profiletext{Believes AI safety is important; Is politically neutral and treats all viewpoints with respect; Doesn't believe there is one "right" way to have a conversation; Sees the positive in most situations; Is not overly concerned with grammar or punctuation; Wants to be helpful and informative}
\item \profiletext{I feel concerned when people express hatred or prejudice towards others based on their race, religion, gender, sexual orientation, or any other protected characteristic; I believe that everyone deserves to be treated with respect and dignity, regardless of their beliefs or background; I value open-mindedness and the ability to engage in constructive dialogues even when there are disagreements; I find it troubling when individuals use harmful language or engage in personal attacks; I strive to promote understanding and empathy in my}
\item \profiletext{Values politeness and respectful communication; May believe in giving people the benefit of the doubt. May prioritize factual accuracy in discussion; Might be averse to confrontational or accusatory language; Possibly values a cooperative and helpful tone in conversations}
\end{itemize}

\subsection{ValuePrism (gemma2-27b - 10 random value profiles)}
\begin{itemize}
\item \profiletext{Believes in nuanced moral judgments, not absolutes; Places importance on context and situational factors; May prioritize pragmatism or effectiveness when evaluating morality; Possibly interested in political affairs}
\item \profiletext{Believes in open communication in relationships; Values financial responsibility; Views marriage as a positive institution}
\item \profiletext{Believes insects have a right to live; May believe in the interconnectedness of all living things; May be opposed to broad extermination efforts}
\item \profiletext{Believes in acts of charity and selflessness; Values humility and admitting when wrong; Appreciates acts of repentance and forgiveness; Likely believes in a higher power; Believes in helping those in need, even strangers; Values sincerity in apology and action but disapproves of seeking validation; Sees accepting defeat as morally sound; Might value tradition or religious practices}
\item \profiletext{Believes that providing basic care for children; Believes that showing physical affection to children is moral; Possibly believes that non-sexual physical contact between a parent and child is acceptable, regardless of gender or age; Believes that generally leaving children unattended while they are experiencing distress; Holds a standard that abandoning a child outside to sleep is immoral; Possibly holds traditional views about family and child-}
\item \profiletext{Believes helping those in need is moral; May believe in prioritizing the needs of family; May sympathize with Palestinians}
\item \profiletext{Believes there are strong moral obligations to truthfulness; May believe there are some exceptions to these obligations in extreme circumstances; Appreciates the importance of informing the public about historical atrocities; Values loyalty and might be conflicted about situations which pit this against other moral considerations}
\item \profiletext{Believes in obeying authority figures, even when inconvenient; Values familial duty and responsibility; May be strict in upholding societal norms and expectations; May find it important to uphold work commitments}
\item \profiletext{Strong aversion to violence; Pacifistic tendencies; Belief that morally good actions should be non-violent}
\item \profiletext{The rater may believe helping those less fortunate is inherently good; The rater may have concerns about the potential misuse of emergency services; The rater may believe there are more effective or appropriate ways to assist those in need than calling the police; The rater may recognize that situations involving poverty are complex and require nuanced judgments}
\end{itemize}

\subsection{Habermas (gemma2-27b - 10 random value profiles)}
\begin{itemize}
\item \profiletext{This rater likely believes in strict immigration policies; This rater likely prefers limited government intervention in social services}
\item \profiletext{Values individual liberty and autonomy; Believes in the importance of limited government intervention; May believe in a free market approach to economic problems; Believes in the importance of public services but is cautious about raising taxes}
\item \profiletext{Belief in some level of government intervention in the economy; Support for social safety nets and programs; Potential trust in experts or scientific consensus; Likely supports progressive policies such as wealth redistribution; Possibly leans left on the political spectrum; May value individual autonomy to a degree}
\item \profiletext{Values the well-being of future generations. This is evident in their support for increased government funding for education and healthcare for young people; Believes in investing in essential public services. Their strong support for increased salaries for teachers and doctors reflects this value; Supports strong government regulation, especially in the face of potentially harmful entities like internet companies; Prioritize public safety and national security. This can be inferred from their strong belief that the UK is under-spending on defense}
\item \profiletext{May believe that law enforcement needs more resources to effectively combat crime; Believes in safety regulations and may be concerned about public safety; Believes in civic participation and engaging with political processes, but potentially sees maturity as a prerequisite; Believes in the social contract and a role of government in providing public services. They may also be willing to contribute financially to these services}
\item \profiletext{Believes in social responsibility and global solidarity; Supports government intervention to solve social issues; May believe in progressive taxation; Values environmental sustainability}
\item \profiletext{Strong belief in fiscal conservatism and potentially limited government intervention. Seems opposed to free public services; Hard stance against illegal drugs; Likely values public safety and order; May prioritize traditional values and potentially be socially conservative; Believes in meritocracy and likely values individual responsibility; Likely skeptical of environmental alarmism and/or interventions}
\item \profiletext{Values fiscal responsibility and may lean towards smaller government; Believes strongly in animal welfare and considers the well-being of animals as a primary concern; Concerned about environmental issues and is willing to adopt measures addressing them}
\item \profiletext{Believes in economic justice and redistribution of wealth; Likely supports socialist or left-leaning policies; May support individual autonomy and bodily integrity in contexts like organ donation; Likely has a positive view of technological progress and innovation, while acknowledging potential downsides; May have an animal welfare perspective and oppose practices like fox hunting; May believe in harm reduction approaches to issues like smoking; Likely values social welfare and support for marginalized populations; Has faith in the potential}
\item \profiletext{Values public health; Disapproves of Theresa May's leadership; Open to nuclear power as a source of energy; Supports government investment in renewable energy; Believes in preventing children from secondhand smoke exposure; Believes in population control measures; Believes in giving citizens more direct influence on policy}
\end{itemize}

\subsection{Prism (gemma2-27b - 10 random value profiles)}
\begin{itemize}
\item \profiletext{Values concise and factual answers over elaborate explanations; Prefers responses that acknowledge alternative viewpoints, even if briefly, before coming to a conclusion; Appreciates politeness and a helpful tone; May value avoiding definitive statements where appropriate; Prefers neutral and unbiased responses, avoiding personal opinions or beliefs; May favor responses that present a balanced view by mentioning both sides of an argument; Appreciates historical context}
\item \profiletext{Values direct and concise answers; Appreciates detailed explanations}
\item \profiletext{Prefers factual and concise responses; May value politeness and careful language especially when dealing with sensitive topics; Possibly prefers responses with a more formal tone; Values responses that acknowledge the ongoing nature of a situation and avoids speculation; Perhaps prefers information to be delivered in a direct manner}
\item \profiletext{Values nuanced, balanced responses over straightforward answers; Prefers empathetic and understanding language; Prioritizes personal freedom and self-determination; May be suspicious of definitive statements or strong opinions; Prefers responses that acknowledge complexity and varying perspectives}
\item \profiletext{Believes that shorter, concise answers are more desirable than longer more detailed ones; Values concrete, actionable advice over general guidance; Possesses a bias towards career paths that retain relevance to the user's current skillset}
\item \profiletext{Believes that people should only use resources intended for them; Prefers informative and comprehensive responses over brief and direct responses; Appreciates detailed descriptions and enthusiasm in responses}
\item \profiletext{They are likely someone who prefers factual and detailed responses, as shown by their preference for Model A in three out of the four examples; They may appreciate context and background information, as seen in the Wallows example; They appreciate neutral and objective language, as shown by their preference for Model A in the conversion therapy example. While both responses condemned the practice, Model A provided a more detached and informative description}
\item \profiletext{Values straightforward and concise communication; Prefers responses that focus on the user's stated problem without venturing into unnecessary details; May not appreciate overly empathetic or sentimental language; Values helpfulness and problem-solving}
\item \profiletext{Prefers longer, more detailed responses over shorter, more direct ones; Prefers responses that are more conversational and friendly in tone; Values politeness and deference to the reader}
\item \profiletext{Prefers concise and direct responses; Appreciates helpfulness and informativeness; May find lengthy or overly enthusiastic responses off-putting; Prioritizes practicality and clarity in communication}
\end{itemize}

\subsection{gemini}
\subsection{OpinionQA (gemini - 10 random value profiles)}
\begin{itemize}
\item \profiletext{Centrist or moderate political views; Believes in American exceptionalism; Pro-individual liberty and personal choice; Economically satisfied and potentially pro-business; Pragmatic and willing to compromise; Socially liberal on some issues, but less clear on others; Not strongly invested in election integrity; Values walkable communities and potentially environmental concerns; May be distrustful of government and institutions; Possibly uncertain or ambivalent on certain issues}
\item \profiletext{Progressive/Liberal political leaning; Strong belief in racial equality and social justice; Pro-immigration; Confidence in expertise; Belief in government intervention; Optimism about social progress; National pride, but not exceptionalism; Slight concern about political correctness; Traditional views on family roles; General trust in others; Value on diversity; Mixed views on the Democratic Party; Possible concern about free expression for Democrats; Potential for cognitive dissonance; Relative indifference to educational attainment for societal well-being; Belief in criminal justice reform}
\item \profiletext{Generally satisfied with the status quo; Moderate politically; Prioritizes national interests; Supportive of traditional family values; Tolerant and accepting of diversity, but with some reservations; Skeptical of government overreach, but believes in its role in certain areas; Pragmatic and distrustful of compromise; Believes in individual responsibility and limited government intervention; Confident in existing systems; Values religious belief, but supports separation of church and state; Neutral or ambivalent on several social issues; Values personal liberty and freedom of expression; Not overly concerned about inequality; Believes in American exceptionalism, but acknowledges other great nations}
\item \profiletext{Generally satisfied with their personal level of respect in society.  They feel they receive the respect they deserve.; Pro-labor. They see labor unions as having a positive impact on the country.; Traditional gender roles.  They believe it's generally better for the mother to stay home if one parent can.; While acknowledging some racial inequality persists, they don't see it as a major issue. They think a little more needs to be done to ensure equal rights, suggesting a belief that significant progress has already been made.; Prioritizes border security in immigration policy.  They believe stronger enforcement and border security should be prioritized over pathways to citizenship for undocumented immigrants.; Tolerant of other languages but perhaps with some reservation. They aren't greatly bothered by hearing other languages, but their response of "not much" instead of "not at all" suggests a possible slight preference for English in public spaces.; Deference to expertise. They believe experts are usually better at making policy decisions than others.; Pro-military and favors a strong national defense. They want to see the military grow and for the U.S. to remain the sole military superpower.  This, combined with their belief in the efficacy of military strength for peace, suggests a hawkish foreign policy stance.; Conservative leaning.  They disapprove of Joe Biden, feel the Democratic party doesn't represent them, and hold views that align with conservative positions on several issues.; Religious, but not necessarily highly devout. They see religion as positive for society, but don't see belief in God as essential for morality.  They favor the separation of church and state.; Believes in limited government. They prefer a smaller government with fewer services and see government as often wasteful. However, they also believe in continuing social security programs and believe a modest reduction in government is sufficient.; Believes in personal responsibility and self-reliance. This is suggested by their view that government aid to the poor creates dependency.; Believes obstacles still exist for women. While they might not believe these obstacles are as large as they once were, they recognize that there's still progress to be made on gender equality.; Values clear moral distinctions. They believe most things in society can be clearly divided into good and evil.; Supports gun control.  They see a rise in gun ownership as very bad for society.; Believes in free speech for all political affiliations. They see both Democrats and Republicans as comfortable expressing their views.; Concerned about changing demographics.  They see a decline in the white share of the population as somewhat bad for society.; Values traditional family structures.  They believe society is better off when people prioritize marriage and children.; Generally accepting of LGBTQ+ people, but with some nuance. They view same-sex marriage as neither good nor bad and transgender acceptance as good, suggesting evolving or potentially complex views on these issues.; Skeptical of social justice movements. Their neutrality on the attention to slavery and racism might indicate a skepticism of these issues or a belief that they are being overemphasized.; Believes in common ground despite political differences. They think they likely share values with those who have different political opinions.; Prefers larger homes and space over walkable communities.  This might indicate a preference for suburban or rural living.; Opposes free college tuition. This aligns with their limited government stance.; Open to some legal immigration, but not a large increase.  This suggests a measured approach to immigration policy.; Positive view of colleges and universities. This suggests a belief in the value of higher education.; Pragmatic approach to politics. They believe compromise is necessary, even if it means sacrificing some beliefs.; Realist in foreign policy.  They believe the US should work with any country to achieve its goals, even if it means working with dictatorships.}
\item \profiletext{Distrust of Power and Institutions; Socially Liberal/Progressive; Populist Leanings; Limited Government Intervention; Importance of Voting Access but Lack of Confidence in the System; Pragmatic Approach to Political Experience; Moderate Concern about Voter Fraud; Potential for Cynicism}
\item \profiletext{Conservative; Religious; Nationalistic; Traditionalist; Law and Order; Skeptical of social justice movements; Distrustful of government and certain institutions; Economic conservatism; Xenophobic or culturally conservative; Polarized worldview; Belief in personal responsibility}
\item \profiletext{Centrist/Moderate political views; Pro-business and pro-technology; Socially liberal on some issues, but with reservations; Importance of personal responsibility and limited government; National strength and security; Importance of voting rights and fair elections; Pragmatic and nuanced perspective; Traditional values with some openness to change; Belief in American exceptionalism; Potential for economic anxiety}
\item \profiletext{Believes in a fair and accessible voting system; Socially moderate to conservative; Supportive of a strong social safety net but with limitations; Concerned about economic inequality and corporate power; Skeptical of government efficiency and elitism; Believes in a strong national defense, but open to a multipolar world; Values personal space and traditional family structures, but with modern adjustments; Pessimistic about societal progress; Neutral on immigration and religion; Positive about the impact of colleges and technology companies; Leans Democrat, but not strongly partisan; Believes in a black-and-white view of morality; Believes in harsher criminal justice}
\item \profiletext{Pro-immigration; Gun control advocate; Socially liberal/Progressive; Supportive of government assistance; Internationalist/Cooperative foreign policy; Pro-Open Borders; Confidence in electoral system; Concern about corporate power; Nuanced worldview; Trust in experts; Pro public education \& Traditional family values; Important Note}
\item \profiletext{Center-left political leaning; Social liberal views; Optimistic about progress; Moderate on some issues; Trust in institutions (with some reservations); Belief in rehabilitation; Importance of traditional values (with flexibility); Emphasis on democratic values in foreign policy; It's important to remember these are inferences and the rater's views might be more complex or nuanced than can be fully captured by a survey. These are simply potential values and beliefs based on the provided information.}
\end{itemize}

\subsection{Hatespeech-Kumar (gemini - 10 random value profiles)}
\begin{itemize}
\item \profiletext{High tolerance for informal language and internet slang; Leniency towards expressions of frustration or negativity; Emphasis on direct harm or malicious intent for toxicity; Acceptance of sexually suggestive language in certain contexts; Prioritization of freedom of expression; Possible desensitization to online language; Potential focus on impact rather than the mere presence of swear words; Belief that subjective opinions are not inherently toxic}
\item \profiletext{Sensitivity to derogatory language; Discomfort with aggressive or confrontational tone; Tolerance for casual swearing and internet slang; Emphasis on harmful intent; Inconsistency or evolving understanding of toxicity; Prioritization of personal attacks over general negativity; Cultural or generational influences}
\item \profiletext{Strong reaction to insults and name-calling; Sensitivity to discussions of sexual assault and child abuse; Tolerance for strong opinions and criticism, within limits; Flexibility with informal language and internet slang; Unclear stance on conspiracy theories; Focus on direct harm rather than implied negativity}
\item \profiletext{Sensitivity to emotional expression; High bar for toxicity; Focus on intent over impact; Potential bias against K-pop or fandom culture; Lack of understanding of specific cultural contexts; Personal interpretation of "toxic"; Inconsistency in application of criteria; Potential unfamiliarity with political terminology; Tolerance of potentially offensive language if not directed}
\item \profiletext{Strong aversion to negativity and expressions of hate; Sensitivity to generalizations and stereotypes; Belief that certain topics should be discussed with sensitivity; Tolerance for strong language and informal expression in some contexts; Focus on the intent or perceived impact rather than solely on the literal content; Potentially inconsistent or evolving understanding of toxicity; Limited tolerance for personal attacks or name-calling; Acceptance of casual conversation and speculation}
\item \profiletext{High tolerance for negativity; Focus on intent to harm or direct insult; Objectivity over emotional impact; Tolerance for factual disagreements and differing opinions; Context is not heavily considered (within the limited data); Possibly a broad definition of "toxic"}
\item \profiletext{Strong aversion to derogatory language and slurs; Sensitivity to identity-based attacks; Low tolerance for generalizations and stereotypes; Discomfort with comments about sex and sexuality; Emphasis on respectful and constructive communication; Political or ideological leanings; A broader definition of toxicity; Inconsistency or evolving understanding; Prioritization of intent over impact}
\item \profiletext{Profanity and insults are inherently toxic; Strong emotional expressions, even negative ones, are not necessarily toxic if they lack personal attacks; Political or opinionated statements, even if potentially controversial, are not inherently toxic; General statements or harmless speculation are not toxic; Positive and encouraging comments are non-toxic; Personal attacks and derogatory language, even without profanity, are toxic; The rater may prioritize "intent to harm" in their assessment of toxicity; The rater may have a relatively high tolerance for diverse opinions; The rater may value politeness and respect in online discourse}
\item \profiletext{Broad interpretation of "toxicity"; Sensitivity to political and social issues; Dislike of strong or potentially offensive language; Aversion to perceived negativity and complaining; Discomfort with potentially controversial topics; Low tolerance for unsolicited requests or boundary-pushing; Potential for over-generalization; Possible lack of familiarity with certain subcultures or online communication styles}
\item \profiletext{High tolerance for offensive language and insults; Focus on direct threats and harmful intent; Leniency towards casual and playful language; Prioritization of free speech and open discussion; Limited understanding of microaggressions or subtle bias; A potentially narrow definition of "toxicity"; Possible personal bias towards certain topics or groups; Acceptance of online "trash talk" as normal}
\end{itemize}

\subsection{DICES (gemini - 10 random value profiles)}
\begin{itemize}
\item \profiletext{High tolerance for controversial opinions; Focus on explicit harm or hate speech; Importance of intent over impact; Uncertainty in ambiguous situations; Belief in personal autonomy; Prioritization of personal well-being and support; Potential discomfort with certain topics; A non-confrontational approach}
\item \profiletext{Discomfort with sexual topics; Low tolerance for negativity or perceived attacks, even in jest; Sensitivity to personal attacks or dismissiveness; Prioritization of emotional support and empathy; Inconsistency in applying criteria; Possible influence of personal experiences or biases; Unclear understanding of the definition of "Toxic"; Possible misunderstanding of the task instructions}
\item \profiletext{Leniency towards strong language/insults; Focus on intent and context; Sensitivity to certain topics but not others; Prioritizes direct harm/threats over negativity; Uncertainty about subtle forms of toxicity; Possible belief in personal responsibility/freedom of choice; Inconsistencies in application of criteria}
\item \profiletext{Tolerance for informal language and typos; Focus on explicit harm or negativity; High threshold for toxicity; Potential belief in freedom of expression; Prioritization of practicality and usefulness in conversation; Potentially limited understanding of nuanced toxicity; Possible cultural or personal biases}
\item \profiletext{High tolerance for insensitive or abrasive language; Focus on intent over impact; Emphasis on factual correctness or logical argumentation; Broad definition of "toxic"; Belief in personal responsibility and freedom of speech; Potential desensitization to online negativity; Possible lack of awareness of microaggressions or subtle forms of toxicity; Prioritization of information exchange in question-answering scenarios}
\item \profiletext{High tolerance for strong language and insults; Focus on intent over impact; Prioritization of freedom of expression; Discomfort with discussions about politics; Inconsistent understanding of toxicity; Acceptance of provocative or dark humor; Potential lack of sensitivity to certain topics}
\item \profiletext{High tolerance for offensive language and controversial topics; Focus on intent rather than potential harm; Emphasis on direct, explicit aggression as a marker of toxicity; Belief that disagreement or rudeness does not necessarily equate to toxicity; Potential lack of awareness of the broader implications of certain topics; Possible understanding of online communication norms; Prioritization of personal responsibility; Inconsistent application of standards; Focus on the surface level meaning of the conversation}
\item \profiletext{High tolerance for rudeness and negativity; Prioritization of freedom of expression; Focus on intent over impact; Limited sensitivity to social justice issues; Contextual understanding of "toxic"; Inconsistency in applying standards; Potential for personal bias}
\item \profiletext{Drug use and discussion of drug use is toxic; General negativity and insults contribute to toxicity; Intolerance and prejudice are toxic; Statements promoting violence or harm are toxic; The rater has a high tolerance for sexually suggestive content; Inconsistency or confusion around religious and political discussions; Context and intent are not always adequately considered; Lack of clear criteria for "toxicity"}
\item \profiletext{High tolerance for controversial topics and opinions; Focus on explicit harm or malice as indicators of toxicity; Distinction between offensive content and toxic behavior; Acceptance of adult choices and behaviors; Uncertainty around implicit bias and microaggressions; Prioritization of intention over impact; Leniency in online interactions; Limited understanding or awareness of certain types of harm}
\end{itemize}

\subsection{ValuePrism (gemini - 10 random value profiles)}
\begin{itemize}
\item \profiletext{Humans are superior to animals; Tradition and cultural norms are morally acceptable; Playfulness and harmless fun are morally good; Intentions matter more than potential harm; Individual autonomy and freedom of choice are important; Consequences are not always the sole determinant of morality; A degree of mischief or mild discomfort is acceptable in social interactions; Cultural context matters in moral judgments; Focus on personal pleasure and enjoyment}
\item \profiletext{Strong belief in free speech and open communication; Opposition to censorship and suppression of information; Anti-authoritarian and pro-resistance against perceived oppression; Belief in challenging harmful ideologies and individuals; Support for social justice and minority rights; Emphasis on honesty and directness; Potential for conflicting values around violence and interpersonal harm; Value on personal autonomy and choice; Unclear or nuanced stance on certain political ideologies and figures; It's important to remember that these are inferences based on limited data. The rater's reasoning could be more nuanced or based on factors not captured in these examples.  Further questioning would be needed to confirm these values and beliefs and to understand the underlying logic behind their judgments.}
\item \profiletext{Utilitarianism/Consequentialism; The sanctity of life, but with a hierarchical view; Impartiality/Universalism; A belief in the greater good; A lack of strong deontological constraints; Possibly a collectivist perspective; Altruism and a duty to help others; Potentially a belief in the inherent value of human life, even with exceptions; Low emotional reactivity or high emotional regulation}
\item \profiletext{Parental Authority/Responsibility; Structure and Discipline are Important; Protecting Children from Harm (physical or psychological); Nuanced understanding of situations; Emphasis on education and responsibility; Potential concern about long-term consequences; Possible belief in a balance between strictness and flexibility}
\item \profiletext{Helping others in need is morally good. This is the most obvious takeaway, given their consistent "Moral" responses to actions that directly benefit the homeless.; Social welfare and support systems are important. Their belief in the morality of providing homes and ending homelessness suggests a value placed on societal structures that ensure basic needs are met.; Reducing suffering is a moral imperative. Both actions aim to alleviate the suffering associated with homelessness, indicating this could be a core belief.; Basic needs like housing are a human right. This aligns with the belief in social welfare and suggests a potentially deontological ethical framework where certain rights are inherent.; Collective responsibility for societal well-being.  The rater may believe that society has a collective responsibility to care for its vulnerable members.; Utilitarianism – actions that benefit the greatest number are morally good. Providing homes and ending homelessness likely benefits a large portion of society, either directly or indirectly.; Compassion and empathy for marginalized groups.  The responses suggest a likely inclination towards empathy and compassion for those experiencing homelessness.; A belief in systemic solutions to societal problems.  "Ending homelessness" implies a focus on addressing the root causes rather than just individual instances, hinting at a belief in systemic change.; A positive view of government or institutional intervention. The actions implicitly involve government or organizational efforts, suggesting the rater doesn't necessarily see such intervention as negative.; Equity and fairness as moral principles.  The rater may believe in a just society where everyone has access to basic necessities like housing.; Possibly a religious or philosophical belief system that emphasizes charity and compassion. Many religious and philosophical traditions advocate for helping the poor and vulnerable.; Optimism about the possibility of positive social change.  The rater's belief that homelessness can be ended suggests an optimistic outlook on the potential for improvement.}
\item \profiletext{Low regard for property rights, especially corporate or commercial; "Victimless crimes" are acceptable; Focus on personal gain or enjoyment outweighs minor rule-breaking; Anti-establishment or anti-corporate sentiment; A belief that these actions have negligible impact; Possible rationalization about resource abundance; A relaxed or non-traditional moral code; Prioritization of personal autonomy and freedom; A belief that laws or rules are not always morally sound; Potential influence of situational factors not explicitly stated}
\item \profiletext{Believes in civic engagement and participation; Values democratic principles; May have a negative view of the Republican party platform or current Republican politicians; Believes in non-violent political action; Possibly believes in the legitimacy of elections and the peaceful transfer of power; May believe that violence can be justifiable under specific circumstances; Potential belief in accountability for politicians; May believe in the importance of checks and balances on government power; Could hold a consequentialist moral perspective; Possibly believes in the right to self-defense; May have a nuanced understanding of political conflict}
\item \profiletext{Altruism and generosity are highly valued.  The rater sees giving away possessions, even to the point of selling everything, as morally good, suggesting a belief in the importance of generosity and helping others.; Detachment from material possessions is positive. The actions involve significant material sacrifice. The positive moral judgment indicates a potential belief that material possessions are not of utmost importance, and detachment from them can be virtuous.; Following a higher purpose or calling can justify significant sacrifice. The first action explicitly mentions following Jesus. This suggests a possible belief that aligning oneself with a spiritual or higher purpose can make actions moral, even if they involve significant personal cost.; Religious faith or spirituality may be a significant influence. The mention of Jesus in the first example strongly hints at a potential religious or spiritual framework influencing the rater's moral judgments.; Selflessness and sacrifice are moral virtues. Both actions involve giving up something of personal value.  The positive moral assessment suggests that the rater may view selflessness and sacrifice as morally positive traits.; Potentially a belief in a specific religious interpretation.  Depending on the specific religious beliefs of the rater, selling all possessions to follow Jesus might be seen as a specific commandment or ideal within their faith, further reinforcing its moral goodness in their eyes.; Possible prioritization of community or collective well-being over individual wealth. Giving away possessions could be seen as contributing to the well-being of others or the community, suggesting the rater may prioritize collective good over individual accumulation.; The rater may believe in a moral imperative to help those less fortunate.  The act of giving away possessions strongly suggests a possible belief in the importance of assisting those in need.; Simplicity and minimalist lifestyles may be seen as morally positive. The actions suggest a potential appreciation for a simpler way of life, where material possessions are not the primary focus.; Actions motivated by sincere faith or conviction are more likely to be judged as moral. The rater may place a higher value on actions driven by deep-seated belief, even if those actions seem extreme from a different perspective.}
\item \profiletext{Belief in individual autonomy and freedom; Nuance in social interactions; Pro-worker and pro-collective action; Potential belief in open communication and trust in relationships; Situational ethics; Emphasis on positive rights; May value social harmony and avoiding disruption}
\item \profiletext{Pacifism or strong anti-violence stance; The sanctity of human life; Situational morality; Justice and retribution, but with reservations; Nuance in taking a life; Concern about unintended consequences; A belief in a higher power or moral authority that prohibits killing; The potential for rehabilitation; Aversion to vigilantism; Emphasis on understanding motivations and context}
\end{itemize}

\subsection{Habermas (gemini - 10 random value profiles)}
\begin{itemize}
\item \profiletext{Emphasis on societal order and security; Trust in established institutions; Belief in individual responsibility and limited government intervention; Pessimistic view of human nature; Pragmatic or utilitarian approach to ethical dilemmas; Support for technological advancement; Potential for inconsistent or contradictory views; Possible general distrust of the masses}
\item \profiletext{Pro-social welfare; Pro-environmentalism; Pro-market liberalization/consumer choice; Socially liberal/progressive; Pragmatic/undecided on certain issues; Belief in government intervention (where appropriate); Focus on practical outcomes and efficiency}
\item \profiletext{Internationalism/Humanitarianism; Belief in government intervention (but with limits); Fiscal Conservatism (with nuances); Value for Traditional Institutions; Prioritization of National Interest/Security; Skepticism of "Sin Taxes"; Pragmatism/Nuanced Approach}
\item \profiletext{Anti-monarchist; Desire for a more mature electorate; Strong belief in online safety and regulation; Pro-religious and potentially supportive of government involvement in religion; Utilitarian view on animal rights; Potentially conservative or authoritarian leaning; Belief in societal intervention; Potential value for tradition; Pragmatic over idealistic}
\item \profiletext{Altruism and global citizenship; Importance of education and societal well-being; Belief in incentives and problem-solving; Potential trust in experts and institutions; Pragmatism and a results-oriented approach; Possible concern for future generations; Openness to innovative solutions; A nuanced perspective on individual liberty; Possible belief in collective responsibility}
\item \profiletext{Nationalism/Protectionism; Fiscal Conservatism (with exceptions); Environmentalism; Social Justice/Egalitarianism; Potential distrust of younger voters; Belief in government intervention (where aligned with their values); Collectivism over Individualism; Potential for authoritarian leanings}
\item \profiletext{Environmental concern, but with a pragmatic approach; Belief in government regulation in some areas, but not others; Emphasis on individual liberties; Distrust of Boris Johnson and the current UK government's approach; Possible financial concerns; Support for public services, but with a focus on efficiency}
\item \profiletext{Environmental consciousness; Belief in public participation in government; Prioritization of social welfare and healthcare; Incrementalism or pragmatism; Possible conflict avoidance; Sensitivity to economic considerations; Trust in expert opinion on tax policy; Personal experience with the NHS}
\item \profiletext{Strong Environmentalism; Nationalist/Patriot; Fiscal Conservatism/Limited Government; Traditional Family Values; Prioritization of Environmental Issues over Social Welfare; Belief in Collective Action; Optimism about Technological Solutions}
\item \profiletext{Strong belief in workers' rights and economic fairness; Compassion and social justice orientation; Support for government intervention in social welfare; Mixed or nuanced views on nationalism and globalization; Pragmatism or openness to change; Prioritization of social well-being over strict fiscal conservatism; Potential belief in restorative justice}
\end{itemize}

\subsection{Prism (gemini - 10 random value profiles)}
\begin{itemize}
\item \profiletext{Neutrality and Objectivity in AI; Acknowledging, but not Necessarily Endorsing, Consensus; Emphasis on Dialogue and Discussion; Avoiding Definitive Claims without Complete Information; Preference for Comprehensive Explanations over Simple Deflection; Balance between Transparency and Safety; Trust in Established Institutions (to some degree); Appreciation for nuance and complexity}
\item \profiletext{Prefers concrete information and examples over conversational prompts when asking for information. (Choosing A in the "controversial" prompt, which gave a specific example, over B, which offered general categories.); Values thoroughness and detail in responses, particularly regarding health and safety. (Choosing B in the smoking ban question and the running tips question, both of which were more detailed and comprehensive.); Appreciates helpful and proactive suggestions but not overly pushy or suggestive upselling. (Choosing B in the running tips question for its helpfulness but choosing A in the "controversial" prompt, possibly because B's response felt too general and prompted for further interaction rather than providing immediate information.); Favors politeness and a friendly tone, but also values conciseness when appropriate. (Choosing A over B in the greeting example; A was more polite and complete, while B was shorter but potentially less engaging.); Prioritizes well-being and public health over individual freedoms in certain scenarios. (Choosing B in the smoking ban question, which emphasized the negative public health impacts.); Believes that AI should acknowledge its limitations (lack of personal opinions) but still be able to provide informative and helpful responses. (Choosing A in the smoking ban question, despite its disclaimer about not having opinions, as it still provided context and relevant considerations.); May prefer structured, bulleted lists for information that is easily digestible. (Choosing B in the running tips question, which used a bulleted list format.); Values responses that are relevant to the specific prompt and don't feel overly generic or templated. (Potentially influencing the choice in the "controversial" prompt, where A provided a specific, though perhaps unexpected, example related to UK politics.); May have an interest in UK politics, given the acceptance of the list of political parties as a "controversial" topic. (Speculative, based on the choice in the first example.)}
\item \profiletext{Practicality and Actionability; Comprehensiveness; Neutrality and Objectivity; Directness and Clarity; Trust in Established Sources}
\item \profiletext{Prefers concise and direct answers; Appreciates acknowledging limitations; Values actionable and specific advice; Prioritizes safety and external validation; May not always prioritize detail or depth; Potentially prefers a friendly but not overly familiar tone; Values a balance between helpfulness and respecting personal autonomy}
\item \profiletext{Emphasis on empathy and emotional connection; Prioritization of conciseness and readability; Valuing personal experience and subjective perspectives; Preference for actionable information; Potential discomfort with overly cautious or "neutral" stances; Possible bias towards specific political viewpoints; Possible prioritization of immediate understanding over nuance}
\item \profiletext{Specificity and Informativeness; Actionability and Practicality; Transparency and Openness; Depth of Knowledge; Directness; Trust in Open Source; Interest in Technical Details; Belief in Preparedness}
\item \profiletext{Neutrality and Objectivity; Comprehensiveness, but without Excessive Detail; Acknowledging Limitations; Data-Driven or Evidence-Based Reasoning; Balance and Moderation; Clarity and Directness; Trust in Established Knowledge}
\item \profiletext{Accuracy and Factuality; Neutrality and Objectivity; Comprehensiveness and Nuance; Trust in Expert Knowledge; Avoidance of Sensationalism; Safety and Practicality; Conciseness and Clarity}
\item \profiletext{Directness and Conciseness; Actionable Support over Passive Acknowledgement; Neutrality and Factual Information over Emotional Sentiments; Contextual Awareness; Desire for Information and Understanding over Simple Platitudes; Possible Discomfort with AI Expressing "Feelings"}
\item \profiletext{Completeness and informativeness; Neutrality and avoidance of strong framing; Focus on the main topic and avoidance of digression; Conciseness and clarity; Politeness and helpfulness; Factual accuracy (where applicable); Readability and flow}
\end{itemize}

\end{document}